\documentclass[10pt,journal,compsoc]{IEEEtran}
\pdfoutput=1
\ifCLASSOPTIONcompsoc
  \usepackage[nocompress]{cite}
\else
  \usepackage{cite}
\fi

\usepackage{graphicx}
\usepackage[cmex10]{amsmath}
\usepackage[ruled,vlined,lined]{algorithm2e}
\usepackage{array}
\usepackage{subcaption}
\usepackage{epsfig}
\usepackage{mathptmx}
\usepackage{times}
\usepackage{amsmath}
\usepackage{amssymb}
\usepackage{amstext}
\usepackage{multirow}
\usepackage{mathtools}
\usepackage{tikz}
\usepackage{verbatim}
\usepackage{epstopdf}
\usetikzlibrary{arrows,shapes,chains,matrix,positioning}
\usetikzlibrary{scopes,decorations,shadows,backgrounds}

\usepackage{lmodern}
\usepackage[T1]{fontenc}
\usepackage{float}
\usepackage{units}
\usepackage{textcomp}
\usepackage{url}
\usepackage{bigstrut}
\usepackage{colortbl,color,soul}
\usepackage{todonotes}

\DeclareMathOperator*{\argmin}{arg\,min}
\DeclareMathOperator*{\argmax}{arg\,max}

\begin{document}
\title{Face Alignment Robust to Pose, Expressions and Occlusions}

\author{Vishnu~Naresh~Boddeti$^{\dagger}$, Myung-Cheol~Roh$^{\dagger}$, Jongju~Shin, Takaharu~Oguri, Takeo~Kanade%
\IEEEcompsocitemizethanks{\IEEEcompsocthanksitem $^{\dagger}$ These authors contributed equally.}%
\IEEEcompsocitemizethanks{\IEEEcompsocthanksitem Contact \protect E-mail: vishnu@msu.edu}%
\IEEEcompsocitemizethanks{\IEEEcompsocthanksitem Vishnu~Naresh~Boddeti is with the Computer Science Department at Michigan State University}%
\IEEEcompsocitemizethanks{\IEEEcompsocthanksitem Myung-Cheol~Roh, Jongju~Shin, Takaharu~Oguri and Takeo~Kanade are with the Robotics Institute at Carnegie Mellon University}%
\thanks{}}

\markboth{}%
{Shell \MakeLowercase{\textit{et al.}}: Bare Demo of IEEEtran.cls for
  Computer Society Journals}

\IEEEcompsoctitleabstractindextext{%

\begin{abstract}
We propose an \emph{Ensemble of Robust Constrained Local Models} for alignment of faces in the presence of significant occlusions and of any unknown pose and expression. To account for partial occlusions we introduce, \emph{Robust Constrained Local Models}, that comprises of a deformable shape and local landmark appearance model and reasons over binary occlusion labels. Our occlusion reasoning proceeds by a hypothesize-and-test search over occlusion labels. Hypotheses are generated by \emph{Constrained Local Model} based shape fitting over randomly sampled subsets of landmark detector responses and are evaluated by the quality of face alignment. To span the entire range of facial pose and expression variations we adopt an \emph{ensemble} of independent \emph{Robust Constrained Local Models} to search over a discretized representation of pose and expression. We perform extensive evaluation on a large number of face images, both occluded and unoccluded. We find that our face alignment system trained entirely on facial images captured ``in-the-lab" exhibits a high degree of generalization to facial images captured ``in-the-wild". Our results are accurate and stable over a wide spectrum of occlusions, pose and expression variations resulting in excellent performance on many real-world face datasets.
\end{abstract}

}

\maketitle

\IEEEdisplaynotcompsoctitleabstractindextext
\IEEEpeerreviewmaketitle

\section{Introduction \label{sec:introduction}}
\IEEEPARstart{A}{ccurately} aligning a shape, typically defined by a set of landmarks, to a given image is critical for a variety of applications like object detection, recognition~\cite{zhu2012face} and tracking and 3D scene modeling~\cite{zia2013detailed}. This problem has attracted particular attention in the context of analyzing human faces since it is an important building block for many face analysis applications, including recognition~\cite{taigman2014deepface} and expression analysis~\cite{martinez2012model}.

\begin{figure}[t]
	\captionsetup{font=small}
	\centering
	\includegraphics[scale=0.6]{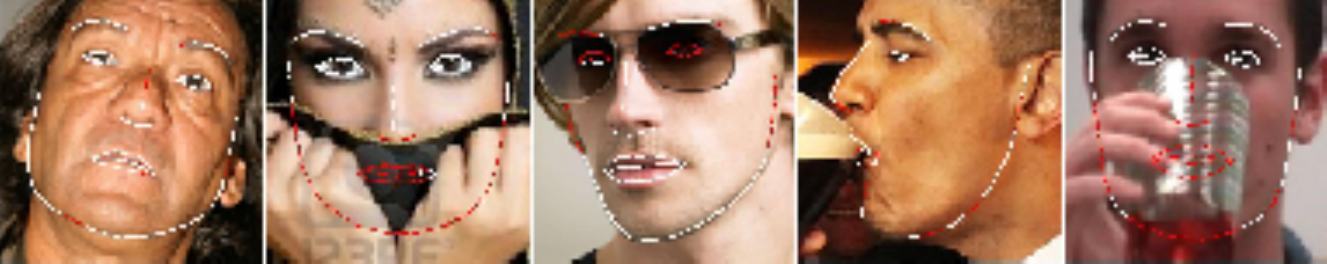}
	\caption{Face images ``in-the-wild" exhibit wide ranging pose variations and partial occlusions presenting significant challenges for face alignment. The white curves and broken red curves represent parts which are determined as visible and occluded, respectively, by ERCLM, our face alignment approach.}
	\label{fig:teaser}
\end{figure}

Robust face alignment is a very challenging task with many factors contributing to variations in facial shape and appearance. They include pose, expressions, identity, age, ethnicity, gender, medical conditions, and possibly many more. Facial images captured ``in-the-wild" often exhibit the largest variations in shape due to pose and expressions and are often, even significantly, occluded by other objects in the scene. Figure \ref{fig:teaser} shows examples of challenging images with pose variations and occlusions, such as food, hair, sunglasses, scarves, jewelery, and other faces, along with our alignment results.

Many standard face alignment pipelines resolve the pose, expression and occlusion factors independently. Shape variations are handled by learning multiple 2D models and selecting the appropriate model at test time by independently predicting pose and expression. Occlusions are typically estimated by thresholding part detector responses which is a difficult and error prone process due to the complexity involved in modeling the entire space of occluder appearance.

Fully or partially occluded faces present a two-fold challenge to this standard face alignment pipeline. First, predicting pose and expressions using global image features is prone to failure, especially for partially occluded faces. Features extracted from the occluded regions adversely affect the response of pose and expression predictors. Second, occluded facial landmarks can adversely affect the response of individual landmark detectors, resulting in spurious detections which, if not identified and excluded, severely degrade the quality of overall shape fitting. However, outlier detections can be identified only through their inability to ``explain away" a valid facial shape.

Facial pose/expression can be reliably estimated by identifying and excluding the occluded facial regions from the pose/expression estimation process. Occluded facial regions can be reliably identified by estimating the correct shape. Therefore, partial occlusions, unknown pose and unknown expressions result in a ``chicken-and-egg" problem for robust face alignment. The pose, expression and landmark occlusion labels can be estimated more reliably when the shape is known, while facial shape can be estimated more accurately if the pose, expression and occlusion labels are known.
\tikzstyle{format} = [draw, thin, fill=blue!20]
\tikzstyle{medium} = [ellipse, , draw, thin, fill=green!20, minimum height=2.0em]
\tikzstyle{decision} = [diamond, draw, fill=blue!20, text width=4.5em, text badly centered, node distance=2cm, inner sep=0pt]
\tikzstyle{block} = [rectangle, draw, fill=blue!20, text width=4.0em, text centered, rounded corners, minimum height=7em]
\tikzstyle{line} = [draw, -latex']
\tikzstyle{cloud} = [draw, ellipse,fill=red!20, node distance=2cm,minimum height=1.5em]
\tikzstyle{circ} = [draw, circle, fill=red!20, node distance=2cm,minimum height=2em]

Alignment of ``in-the-wild" faces of unknown pose, unknown expressions and unknown occlusions is the main focus of this paper. We propose \emph{Ensemble of Robust Constrained Local Models (ERCLMs)} to address the ``chicken-and-egg" problem of joint and robust estimation of pose, expression, occlusion labels and facial shape by an explicit and exhaustive search over the discretized space of facial pose and expression while explicitly accounting for the possibility of partially occluded faces. More specifically ERCLM addresses these challenges thusly,
\begin{enumerate}
	\item we adopt a discretized representation of pose, expression and binary occlusion labels, that are spanned by multiple independent shape and landmark appearance models,
	\item we adopt a hypothesize-and-test approach to efficiently search for the optimal solution over our defined space of facial pose, expression and binary occlusion labels, and finally,
	\item we choose the best hypothesis that minimizes the shape alignment error and pass it through a final shape refinement stage.
\end{enumerate}
Unlike most previous face alignment approaches, ERCLM explicitly deals with occlusion and is thus \emph{occlusion-aware}; more than just being robust to occlusion, i.e., it also estimates and provides binary occlusion labels for individual landmarks in addition to their locations. This can serve as important auxiliary information and can be leveraged by applications that are dependent on face alignment, such as face recognition~\cite{prabhu2011unconstrained}, 3D head pose estimation, facial expression recognition, etc. We evaluate ERCLM on a large number of face images spanning a wide range of facial appearance, pose and expressions, both with and without occlusions. Our results demonstrate that our approach produces accurate and stable face alignment, achieving state-of-the-art alignment performance on datasets with heavy occlusions and pose variations.

A preliminary version of RCLM appeared in~\cite{roh2011face} where the general framework for alignment of frontal faces in the presence of occlusions was proposed. In this paper we present a significantly more robust version of this algorithm for handling unknown facial pose, expression and partial occlusions. This is achieved by using a more robust local landmark detector, a new hypothesis generation scheme of sampling hypotheses from non-uniform distributions and a new hypothesis filtering process using exemplar facial shape clusters. We demonstrate the generalization capability of ERCLM by training our models on data collected in a laboratory setting with no occlusions, and perform extensive experimental analysis on several datasets with face images captured ``in-the-wild".

The remainder of the paper is organized as follows. We briefly review recent face alignment literature in Section \ref{sec:related_work} and describe ERCLM, our proposed face alignment approach, in Section \ref{sec:approach}. In Section \ref{sec:experiments} we describe our experimental results as well as the datasets that we evaluate ERCLM on and perform ablation studies in Section \ref{sec:ablation}. Finally we discuss some features of ERCLM in Section \ref{sec:discussion} and conclude in Section \ref{sec:conclusions}.

\section{Related Work \label{sec:related_work}}
Early work on face alignment was largely designed to work well under constrained settings i.e., no significant occlusions, near frontal faces or known facial pose. These approaches~\cite{wang2008enforcing,zhou2003bayesian,cootes1995active,gu2008generative,cristinacce2006feature,belhumeur2011localizing}, try to find the optimal fit of a regularized face shape model by iteratively maximizing the shape and appearance responses. However, such methods often suffer in the presence of gross errors, called \emph{outliers}, caused by occlusions and background clutter. There has been a tremendous surge of interest on the problem of facial alignment of late and a large number of approaches have been proposed. A full treatment of this vast literature is beyond the scope of this paper. We instead present a broad overview of the main techniques and focus on a few state-of-the-art methods against which we benchmark our proposed approach.

\noindent\textbf{Parametrized Shape Models:} Active Shape Models (ASM)~\cite{cootes1995active} and Active Appearance Models (AAM)~\cite{cootes2001active} are the earliest and most widely-used approaches for shape fitting. In ASM landmarks along profile normals of a given shape are found, the shape is updated by the landmarks, and is iterated until convergence. AAM, a generative approach, finds shape and appearance parameters which minimize appearance error between an input image and generated appearance instances via optimization. Building upon the AAM, many algorithms have been proposed~\cite{matthews2004active,xiao2004real,donner2006fast,lee2009tensor,dedeoglu2007asymmetry} to address known problems like pose variations, illumination variations and image resolution. However due to their poor generalization capability, AAMs are prone to fail when the input image is different from the training set~\cite{gross2005generic}. Furthermore, while AAM based approaches~\cite{lee2009tensor,cootes2002view} using multiple shape models to span the large range of possible facial poses have been proposed, they still require pose estimation to select the right shape model.

Constrained Local Models (CLMs)~\cite{wang2008enforcing,cristinacce2008automatic,lucey2009efficient,liang2008face,valstar2010facial,saragih2009face,zhu2012face,asthana2015pixels} are another class of approaches for face alignment that are largely focused on global spatial models built on top of local landmark detectors. Since CLMs use local appearance patches for alignment, they are more robust to pose and illumination variations compared to holistic and generative approaches like AAMs. Typical CLM based methods assume that all the landmarks are visible. However including detections from occluded landmarks in the alignment process can severely degrade performance. From a modeling perspective, our approach is conceptually a CLM, i.e., with an appearance and a shape model. However, it is explicitly designed to account for occluded facial landmarks, predicting not only the landmark locations but their binary occlusion labels as well.

\noindent\textbf{Exemplar Models:} Belhumeur et.al.\cite{belhumeur2011localizing} proposed a voting based approach to face alignment. Facial shape was represented non-parametrically via a consensus of exemplar shapes. This method demonstrated excellent performance while being also robust to small amounts of occlusions. However, their approach was limited to near frontal faces and only detected landmarks that are relatively easy to localize, ignoring the contours which are important for applications like face region detection and facial pose and expression estimation.

\noindent\textbf{Shape Regression Models:} Many discriminative shape regression~\cite{cao2012face,xiong2013supervised,burgos2013robust} based face alignment approaches have been proposed in the literature. Instead of relying on parametrized appearance and shape models, these techniques leverage large amounts of training data to learn a regressor, typically a cascaded series of them, mapping stationary image features \cite{fleuret2008stationary} to the final facial shape.

\noindent\textbf{Occlusion Methods:}
Recently, a few face alignment methods have been proposed that are robust to occlusions. Ghiasi and Fowlkes \cite{ghiasi2014occlusion} proposed a CLM based approach to account for occlusions at the learning stage by simulating facial occlusions. Burgos-Artizzu et. al.~\cite{burgos2013robust} proposed a shape regression based approach that is explicitly designed to be robust to occlusions when facial landmark occlusion labels are available at training. These approaches require occluded landmarks, real or artificially simulated, for training their models in a purely discriminative manner. Our approach, in contrast, does not require landmark occlusion labels (which are usually unavailable, especially for dense landmarking schemes used in this paper) for training. We employ a generative shape model at inference and account for outlier landmark detections caused by occlusions, without being trained on occluded faces (real or simulated).

\section{Occlusion Robust Face Alignment \label{sec:approach}}
\begin{figure}[!ht]
	\captionsetup{font=small}
	\centering
	\includegraphics[width=1\linewidth]{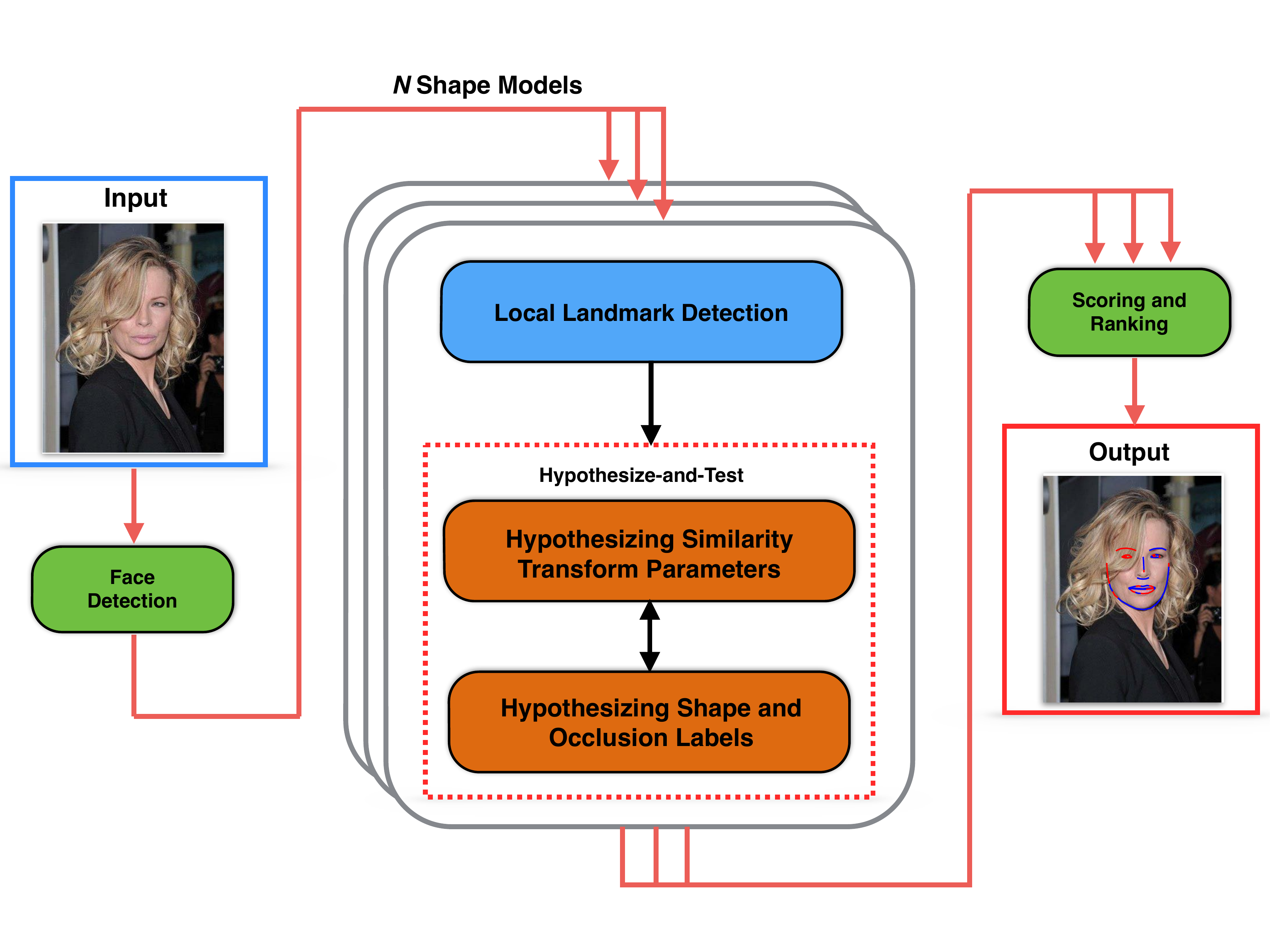}
	\caption{An overview of Robust Constrained Local Models (ERCLMs) for face alignment. Given an image, faces are detected using a face detector. For each face in the image, we fit $N$ different shape models corresponding to different facial pose and expression modes. For each mode we first get candidate landmark estimates from the local landmark appearance models. We then estimate the geometric transformation parameters and the shape parameters via a hypothesize-and-evaluate procedure. We finally select the best facial shape alignment hypothesis and refine it to get the final face alignment result.}
	\label{fig:overview}
\end{figure}

While there has been much focus on face alignment models, there has been relatively little attention paid to the robustness aspect of this task. Large gains in performance for alignment can be achieved by explicitly accounting for variations in pose, deformations and occlusions. Given a face image, in our approach, the goal of face alignment is to find the correct facial pose and expression, a combination of visible and correct landmarks, and the corresponding shape parameter. A pictorial illustration of our face alignment framework is shown in Fig. \ref{fig:overview}.
For the sake of computational efficiency we first estimate a coarse face region using a face detector (ours is based on \cite{schneiderman2004object}). Given the face region and a shape mode, the corresponding local landmark detectors are applied at multiple image scales to obtain response maps. The response maps are processed to extract candidate landmark locations which serve as initializations for the corresponding shape model. From this set of landmark initializations we seek a correct combination of the peaks, i.e., visible landmarks whose locations match well with the facial shape model. We employ a coarse-to-fine hypothesize-and-test approach, first estimating the geometric transformation parameters followed by the shape parameters. We simultaneously hypothesize the right combination of peaks, from the multiple candidate landmark estimates, as well as the occlusion labels of the selected landmarks. We repeat this procedure for each and every facial pose and expression mode and select the one that best ``explains" the observations. Finally, this face alignment result is refined using landmark detector responses re-estimated on the aligned face image. Landmarks which contribute to the final face alignment result are labeled as visible while the rest are deemed to be occluded. In the following subsections we describe the various components of ERCLM, namely, local landmark appearance model, facial shape model and our occlusion reasoning algorithm.

\subsection{Appearance Model \label{sec:appearance_model}}
The appearance model is tasked with providing candidate landmark estimates which serve as initializations for the shape model. These local landmark detectors must be robust to the high variability in the appearance of facial parts due to factors like skin color, background clutter, facial pose and expressions. We now describe the different components of our appearance model i.e., the detector model, the representation we use for the multi-modal response maps and our clustering based approach to handle the multi-modal nature of the landmark appearance due to pose and expression variations.

\subsubsection{Landmark Detector}
In the CLM framework, an independent detector is trained for each individual facial landmark. Due to background clutter and substantial variations in color and pose, capturing the local appearance can be quite challenging. Discriminative feature representations in conjunction with discriminative classifiers can help overcome these challenges. Many different feature representations can be used for our task including Haar-like features~\cite{viola2001rapid}, Local Binary Patterns (LBP)~\cite{ojala1996comparative}, Modified Census Transform (MCT)~\cite{froba2004face}, Scale-Invariant Feature Transform (SIFT)~\cite{lowe1999object} and Histogram of Oriented Gradient (HOG)~\cite{dalal2005histograms}. Our local landmark detector is based on MCT+Adaboost due to its robustness to illumination variations and good detection performance~\cite{froba2004face,liao2007learning}. The MCT features, like LBP features, are very easy to compute. Conceptually LBP and MCT features are a non-linear mapping of 3$\times$3 blocks of pixel intensities to binary edge kernels. LBP spans 256 of the 511 possible binary edge kernels in a 3$\times$3 block while MCT spans all 511 of them. MCT features, therefore, have greater representational capacity in comparison to LBP and form the basis of our local landmark detector.
\begin{figure}[!ht]
	\captionsetup{font=small}
	\centering
	\includegraphics[scale=0.7]{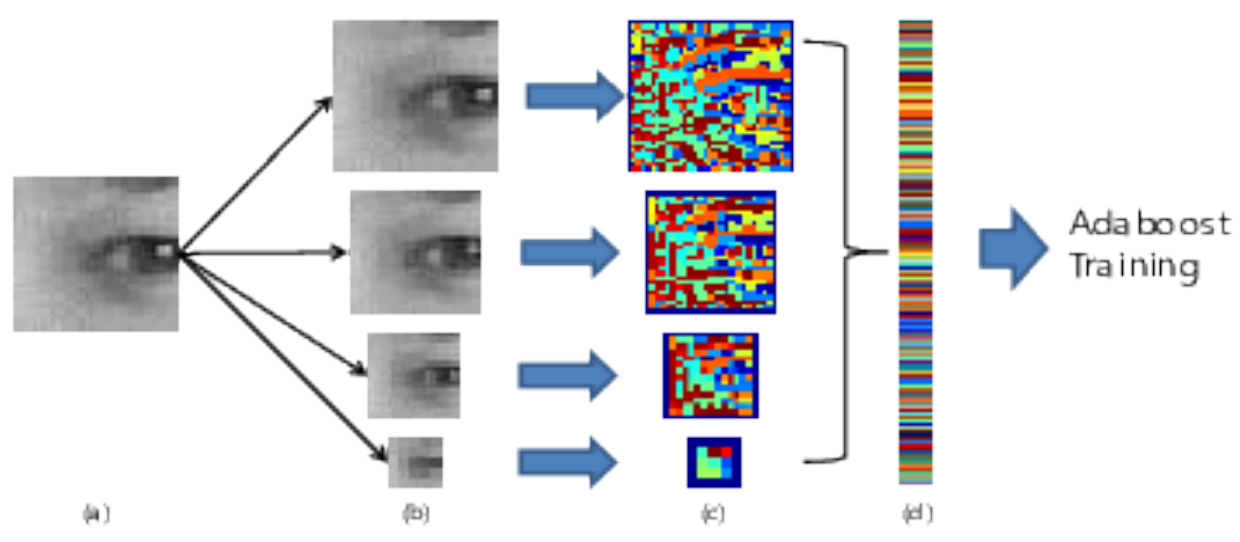}	
	\caption{Hierarchical MCT+Adaboost: (a) given an image, (b) a four level image pyramid is built, (c) MCT feature descriptor is extracted at each pyramid level and (d) MCT feature descriptors are concatenated and used to select weak classifiers by Adaboost.}
	\label{fig:hmct}
\end{figure}
The scale or resolution of each landmark determines the amount of local information that aids in detecting the corresponding landmark. Different landmarks could however be best localized using different amounts of detail. To capture information at multiple scales we propose a hierarchical MCT feature representation as our feature descriptor. Figure \ref{fig:hmct} shows our hierarchical MCT feature extraction process for an eye corner. For a given landmark, we construct a pyramid of patches at different scales and extract MCT features. The MCT features from all the patches are concatenated into a single vector. In this paper, we consider four different pyramid levels of sizes 35 $\times$ 35, 25 $\times$ 25, 15 $\times$ 15, 5 $\times$ 5 to get a 1796 dimensional feature vector while the conventional MCT method results in a vector with 1089 dimensions.
\begin{table}[t]
\centering
\caption{Comparison of Appearance Models}
\label{table:appearance}
\scalebox{0.9}{
\begin{tabular}{|c|c|c|c|c|}
	\hline
	& \multicolumn{2}{c|}{Conventional} & \multicolumn{2}{c|}{Hierarchical} \\
	\hline
	& LBP & MCT & LBP & MCT \\
	\hline
	FN & 1 & 3 & 7 & 3 \\
 	\hline
 	FP & {\footnotesize 31018(25\%)} & {\footnotesize 15746(12.7\%)} & {\footnotesize 12661(10.2\%)} & {\footnotesize 3972(3.2\%)} \\
 	\hline
\end{tabular}}
\end{table}
In Table \ref{table:appearance} we compare the discriminative performance of LBP, MCT, hierarchical LBP and hierarchical MCT features on a training set of 31,032 positive and 123,867 negative samples. We used training patches of size 35 $\times$ 35 pixels for the LBP and MCT features and patches with four different contextual extents for the hierarchical LBP and MCT. Using Adaboost we learned 100 weak classifiers and compare the number of false negatives and false positives for the different feature representations. We note that hierarchical MCT has the lowest number of false positives.
\begin{figure}[!ht]
	\captionsetup{font=small}
	\centering
	\includegraphics[trim=0 2 0 0,clip,scale=1]{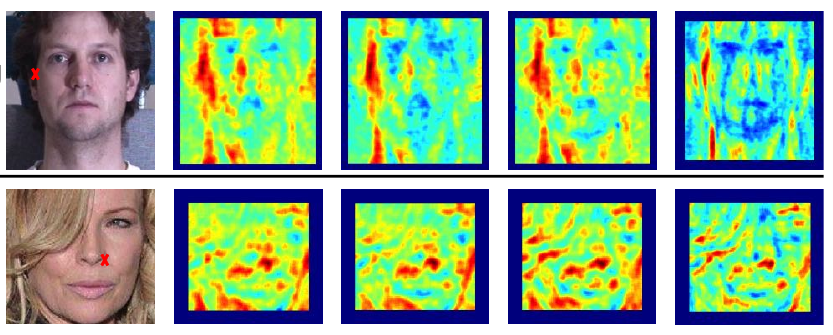}
	\caption{Response maps of landmark detectors. The input image with the {\color{red}$\times$} showing the landmark under consideration is shown along with the response maps of conventional LBP, conventional MCT, hierarchical LBP, and hierarchical MCT respectively.}
	\label{fig:appearance}
\end{figure}
Figure \ref{fig:appearance} shows the response maps for each feature descriptor computed as the sum of the responses of the weak classifiers' learned using Adaboost. The hierarchical MCT based classifier, in comparison to the other features, results in fewer false positives and better landmark localization.
\subsubsection{Representation of Multi-Modal Response Maps}
The response maps ($\mathbf{r}_i$) are discretized by first finding the modes corresponding to a detection and approximating each mode by an independent Gaussian. We represent the entire response map for a given landmark as a combination of independent Gaussians. For a given landmark, the number ($K$) of candidate landmark estimates can range from zero to many, depending on the number of detected modes.
\begin{eqnarray}
	\mathbf{r}_i = \sum_{k=1}^K\delta_{k} \mathcal{N}(i;\mu_{i;k},\Sigma_{i;k})
	\label{eq:independent_gaussians}
\end{eqnarray}
where $\mu_{i;k}$ and $\Sigma_{i;k}$ are the mean and the covariance respectively of the $k$-th Gaussian corresponding to the $i$-th landmark, and $\delta$ is the Kronecker delta function.

The modes of the response map are found by partitioning it into multiple regions using the Mean-Shift segmentation algorithm~\cite{comaniciu2002mean}. Each of these segmented regions is approximated via convex quadratic functions~\cite{wang2008enforcing}:

{\scriptsize
\begin{eqnarray}
	\label{eq:meanshift}
	\argmin_{\mathbf{A},\mathbf{b},c} && \sum_{\Delta\mathbf{x}} \|E\{I(\mathbf{x}+\Delta\mathbf{x})\} - \Delta\mathbf{x}^T\mathbf{A}\Delta\mathbf{x} + 2\mathbf{b}^{T}\Delta\mathbf{x} - c\|^2_2 \\
	s.t. && \mathbf{A} \geq \mathbf{0} \nonumber
\end{eqnarray}}
\noindent where $E\{I\}$ is the inverted match-score function obtained by applying the landmark detector to the input image $I$, $\mathbf{x}$ is the center of the landmark search region, $\Delta\mathbf{x}$ defines the search region. The parameters $\mathbf{A} \in \mathbb{R}^{2\times 2}$, and $\mathbf{b} \in \mathbb{R}^{2\times 1}$ and $c \in \mathbb{R}$ characterize the convex quadratic function (2-D Gaussian) approximating the landmark detector response in each segment.
\begin{figure}[!ht]
\captionsetup{font=small}
	\centering
	\includegraphics[scale=0.7]{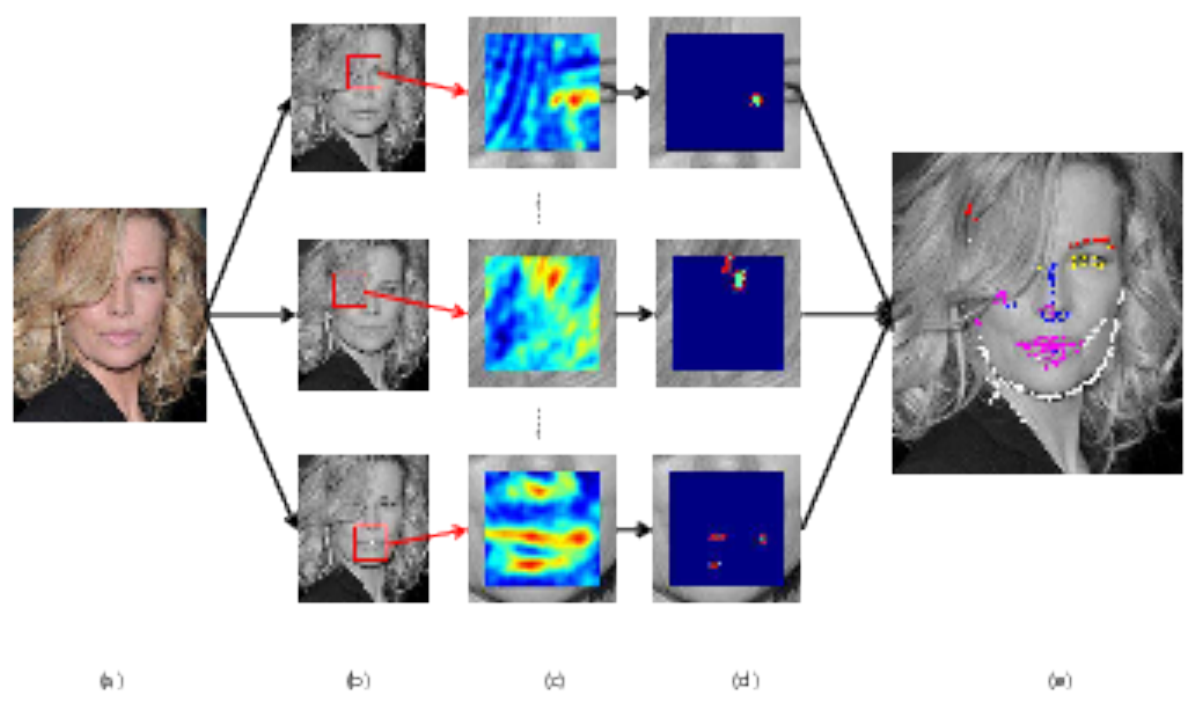}	
	\caption{Local landmark detection process. (a) input image, (b) search region for each landmark, (c) response map for landmark obtained from hierarchical MCT+Adaboost, (d) candidate landmark estimates in each response map, and (e) all candidate landmark estimates.}
	\label{fig:localdetectors}
\end{figure}
Figure \ref{fig:localdetectors} shows how an input image is processed to generate the initial landmark detections. Given an input image, for each landmark response maps from the corresponding detectors are processed to obtain the landmark detections. The circles in Fig. \ref{fig:localdetectors}(d) show the detections along with their estimated distributions. In Fig. \ref{fig:localdetectors}(c), the second row shows the response map where the landmark is occluded. Due to the hair occluding her right eye and eyebrow the corresponding landmark detections are false positives and should ideally be excluded from the alignment process. However, as described earlier, the occlusion label of the landmark detections cannot be determined unless the face alignment is known.

\subsubsection{Clustering}
\begin{figure}[h]
\captionsetup{font=small}
	\centering
	\begin{subfigure}[b]{.3\linewidth}
		\centering
		\includegraphics[scale=0.8]{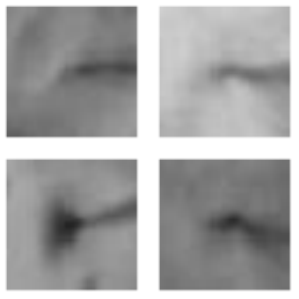}
		\caption{Neutral}
	\end{subfigure}
	\begin{subfigure}[b]{.3\linewidth}
		\centering
		\includegraphics[scale=0.8]{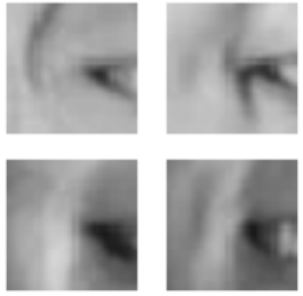}
		\caption{Smile}
	\end{subfigure}
	\begin{subfigure}[b]{.3\linewidth}
		\centering
		\includegraphics[scale=0.8]{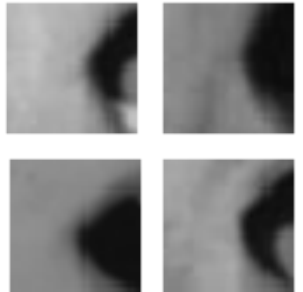}
		\caption{Surprise}
	\end{subfigure}
	\caption{The appearance of the mouth corner varies with facial expressions: (a) neutral,(b) smile, and (c) surprise. Multiple landmark detectors are used to detect the mouth corner under different expressions.}
	\label{fig:expression}
\end{figure}

Facial parts exhibit large appearance variations with pose and expressions. For example, the shape and texture of the mouth is heavily dependent on facial  expression (see Fig. \ref{fig:expression} for illustrative examples). Using a single detector to localize the landmarks associated with the mouth, over all shapes and appearances, severely degrades the detection performance. Therefore, we employ multiple detectors to effectively capture the wide range of appearance variations of the mouth. For each landmark associated with the mouth, we manually cluster the training data into multiple expressions: neutral, smile and surprise. At the test stage, for each landmark associated with the mouth region, detections from all the multiple landmark detectors are merged.

In summary, given a face region, the landmark response maps are obtained at multiple scales (for robustness to imperfect face detection) and landmark detections are obtained from each response map. These detections are then aggregated to get the final set of candidate detections for each landmark.

\subsection{Shape Model \label{sec:shape_model}}
During shape fitting the CLM framework for object alignment regularizes the initial shape, from the local landmark detectors, using a statistical distribution (prior) over the shape parameters.

\subsubsection{Point Distribution Model}
In our model the variations in the face shape are represented by a Point Distribution Model (PDM). The non-rigid shape for $N$ local landmarks, $S = \left[\mathbf{x}_1 , \mathbf{x}_2 ,\dots , \mathbf{x}_N\right]$, is represented as,
\begin{eqnarray}
	\mathbf{x}_i = \mathbf{sR}(\mathbf{\bar{x}}_i+\Phi_i\mathbf{q})+\mathbf{t}
	\label{eq:shapemodel}
\end{eqnarray}
\noindent where $\mathbf{s}$, $\mathbf{R}$, $\mathbf{t}$, $\mathbf{q}$ and $\Phi_i$ denote the global scale, rotation, translation, shape deformation parameter, and a matrix of eigenvectors associated with $\mathbf{x}_i$, respectively. Let $\mathbf{\Theta} = \{\mathbf{s},\mathbf{R},\mathbf{t},\mathbf{q}\}$ denote the PDM parameter. Assuming conditional independence, face alignment entails finding the PDM parameter $\mathbf{\Theta}$ as follows~\cite{saragih2009face}:
\begin{eqnarray}
	\argmax_{\mathbf{\Theta}} p(\{l_i=1\}^N_{i=1}|\mathbf{\Theta}) = \argmax_{\mathbf{\Theta}}\prod_{i=1}^N p(l_i=1|\mathbf{x}_i)
	\label{eq:pdm}
\end{eqnarray}
\noindent where $l_i \in \{-1,+1\}$ denotes whether the $\mathbf{x}_i$ is aligned or not.

Facial shapes have many variations depending on pose and expression and a single Gaussian distribution, assumed by a PDM model, is insufficient to account for such variations. Therefore, we use multiple independent PDM (Gaussian distribution) models. Using multiple shape models to span a range of pose and expressions is not new. Among recent work, Zhu et.al \cite{zhu2012face} and Jaiswal et.al. \cite{jaiswal2013guided} use multiple shape models with the former using manual clustering while the latter performs unsupervised clustering (on frontal faces only).

We partition the training data into $P$ clusters to capture the variations in pose and further partition each cluster into $E(k)$, $k \in \{1,\dots,P\}$ clusters to account for different expressions. We learn one PDM model for each partition. Given the pose and expression cluster assignments $n$ and $m$ respectively, the shape is represented by,
\begin{eqnarray}
	\mathbf{x}_i(n,m) = \mathbf{sR}(\mathbf{\bar{x}}_i(n,m) + \Phi_i(n,m)\mathbf{q}) + \mathbf{t}
\end{eqnarray}
From Eq. \ref{eq:pdm} and the model described above, the face alignment problem is now formulated as:
{\small
\begin{eqnarray}
	\argmax_{\mathbf{\Theta},n,m} p(\{l_i=1\}^N_{i=1}|\mathbf{\Theta},n,m) = \argmax_{\mathbf{\Theta},n,m}\prod_{i=1}^N p(l_i=1|\mathbf{x}_i(n,m))
	\label{eq:pdm_clusters}
\end{eqnarray}
}
\subsubsection{Dense Point Distribution Model}
\begin{figure}[!ht]
\captionsetup{font=small}
	\centering
	\includegraphics[scale=0.8]{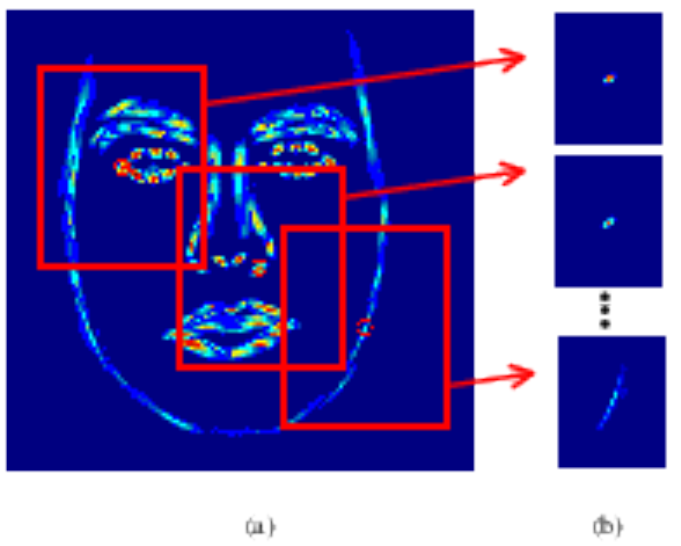}
	\caption{Distribution of landmark detector responses: (a) landmark detector response distributions of all landmarks. (b) distributions: right eye corner (top), left nostril (middle), and left jawline (bottom).}
	\label{fig:shapedistribution}
\end{figure}
\begin{figure}
\captionsetup{font=small}
	\includegraphics[scale=0.6]{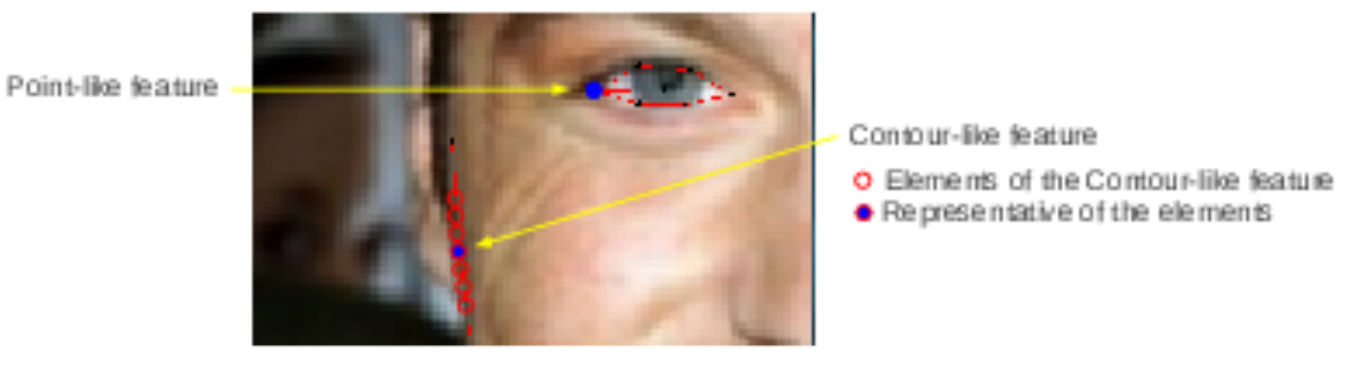}
	\caption{Examples of point-like and contour-like landmarks. Each contour-like landmark, is composed of one ``representative" and seven ``element" landmarks.}
	\label{fig:contour}
\end{figure}

Observing the distributions of detector responses of individual landmarks in Fig. \ref{fig:shapedistribution} we notice that there are two distinct types of landmarks, namely points ($\Omega$) and contours ($\Upsilon$). For example, the distributions of eye corner and nostril detectors (top and middle images in Fig. \ref{fig:shapedistribution}(b)) in the landmark response maps are shaped like points while that of the jawline region detector (bottom image in Fig. \ref{fig:shapedistribution}(b)) is shaped like a contour. While the point-like landmarks are relatively easy to localize, the contour-like landmarks are often poorly localized due to their positional uncertainty along the contour. Therefore, using the contour-like candidate landmark estimates in the shape-fitting process may result in a misalignment. To mitigate this effect we define a dense point distribution model (DPDM) for contour-like landmarks. From the PDM shape $S = [\mathbf{x}_1$,\dots,$\mathbf{x}_N]$, we define the new DPDM shape $S^D$ as:
\begin{eqnarray}
	S^{D} = \cup_{i=1}^N D_i = [\mathbf{x}^D_1,\dots,\mathbf{x}^D_{N^D}], N \leq N^D
	\label{eq:DPDM}
\end{eqnarray}
\[D_i = \left\{
  \begin{array}{lr}
    {\mathbf{x}_i} & : {\mathbf{x}_i \in \Omega}\\
    {\mathbf{x}^{'}_{j}|\mathbf{x}^{'}_{j}=C(\mathbf{x}_{i-1},\mathbf{x}_i,\mathbf{x}_{i+1},N_s)} & : {\mathbf{x}_i \in \Upsilon}\\
  \end{array}
\right.
\]
\noindent where $C(\mathbf{x}_{i-1},\mathbf{x}_i,\mathbf{x}_{i+1},N_s)$ is an interpolation function that generates $N_s$ samples on the curve between $\mathbf{x}_{i-1}$ and $\mathbf{x}_{i+1}$. Therefore, a contour-like landmark ($D_i$) is composed of one ``representative" landmark and a few ``element" (interpolated) landmarks. Figure \ref{fig:contour} shows an example where the red circles and the blue dots represent the ``elements" and ``representative" landmarks respectively. Each ``representative" landmark is explicitly allowed to move along its contour. Further, all the ``elements" associated with the same ``representative" landmark share the same landmark detector response map. Therefore the DPDM does not incur any additional computational cost over the PDM with respect to the appearance model. In the alignment process, only one of the selected ``elements" of the contour-like landmark contributes to the alignment. The alignment problem from Eq. \ref{eq:pdm_clusters} is now re-formulated as:

{\footnotesize
\begin{eqnarray}
	\argmax_{\mathbf{\Theta},n,m,\mathcal{F}} p(\{l_i=1\}^N_{i=1}|\mathbf{\Theta},n,m,\mathcal{F}) = \\
	  \argmax_{\mathbf{\Theta},n,m,\mathcal{F}}\prod_{i=1}^N p(l_i=1|\mathbf{x}^D_{\mathcal{F}(i)}(n,m)) \nonumber
	\label{eq:shape}
\end{eqnarray}}
\noindent where $\mathcal{F}(i)$ is an indicator function selecting the $i$-th ``element" among $D_i$. Through the rest of the paper, `Shape Model' refers to this dense shape model.

\subsection{Occlusion Model and Inference \label{sec:occlusion_model}}
In our framework, the problem of face alignment is to find the correct facial pose and expression ($n$ and $m$) mode, a combination of visible and correct landmarks ($\mathcal{F}$), and the PDM parameter ($\mathbf{\Theta}$). Given the landmark detections from the processed landmark response maps, shape estimation grapples with the following challenges:
\begin{enumerate}
	\item Landmarks could be occluded and this information is not known a-priori. The associated candidate landmark estimates could be at the wrong locations and hence should be eliminated from the shape fitting process.
	\item Each unoccluded landmark can have more than one potential candidate. While most of them are false positives there is one true positive which should contribute to face alignment.
\end{enumerate}

We address these challenges by first noting that the shape model lies in a space whose dimensionality is considerably less than the dimensionality of the shape $S^D$. Therefore, even a small minimal subset of ``good" (uncorrupted) landmarks is sufficient to ``jump start" the PDM parameter $\mathbf{\Theta}$ estimation process and hallucinate the full facial shape. Given the landmark detections from the appearance model, for each of the $Q$ (=$n$$\times$$m$) shape models, we perform the following operations: hypothesize visible and correct candidate landmarks, hallucinate and evaluate a shape model by its agreement with the landmark response map and find the best hypothesis. $Q$ shapes obtained from the $Q$ different shape models are evaluated by their agreements to the observed shape and the best shape is chosen and further refined. The salient features of our occlusion model are:
\begin{enumerate}
	\item Generating PDM parameter hypothesis $\mathbf{\Theta}$ using subsets from the pool of landmark detections. We sample the hypotheses from distributions derived from the landmark detector confidence scores.
	\item Using median for evaluating hypotheses based on the degree of mismatch, due to better tolerance to outliers compared to the mean. This favors a hypothesis in which a majority of the landmarks match very well while some do not (possibly occluded landmarks), instead of one in which all the landmarks match relatively well on average.
\end{enumerate}

In the following subsections we will describe our hypothesis generation and shape hallucination procedure, our shape evaluation and selection procedure and the final shape refinement process.

\subsubsection{Hypothesis Generation and Shape Hallucination}
Given the set of landmark detections, a subset of these are selected to generate a shape hypothesis, a facial shape is hallucinated and evaluated. This procedure is iterated until a given condition (find a good hypothesis) is satisfied. Since the occlusion label of each landmark is unknown along with the correct detections which fit the facial shape, two different kinds of hypotheses are taken into account: hypothesis of landmark visibility and hypothesis of correct landmark candidates i.e., visibility of landmarks is hypothesized along with the candidate landmark detection associated with that landmark.
\begin{figure*}[!ht]
\captionsetup{font=small}
	\begin{subfigure}[b]{0.58\linewidth}
	\centering
	\begin{tikzpicture}
		\node (a) {\reflectbox{\includegraphics[trim=30 100 20 30,clip,width=0.17\linewidth]{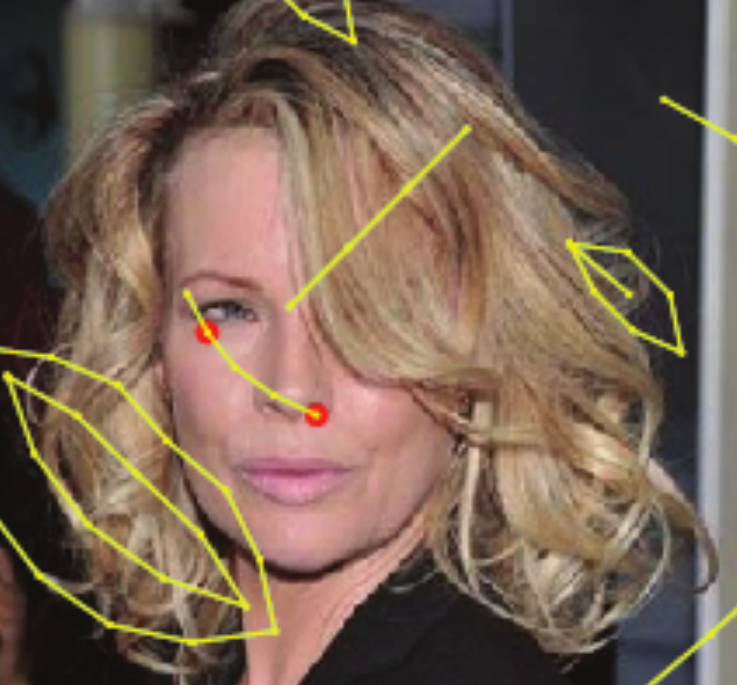}}};
		\node[right of=a, node distance=1.9cm] (b) {\reflectbox{\includegraphics[trim=30 100 20 30,clip,width=0.17\linewidth]{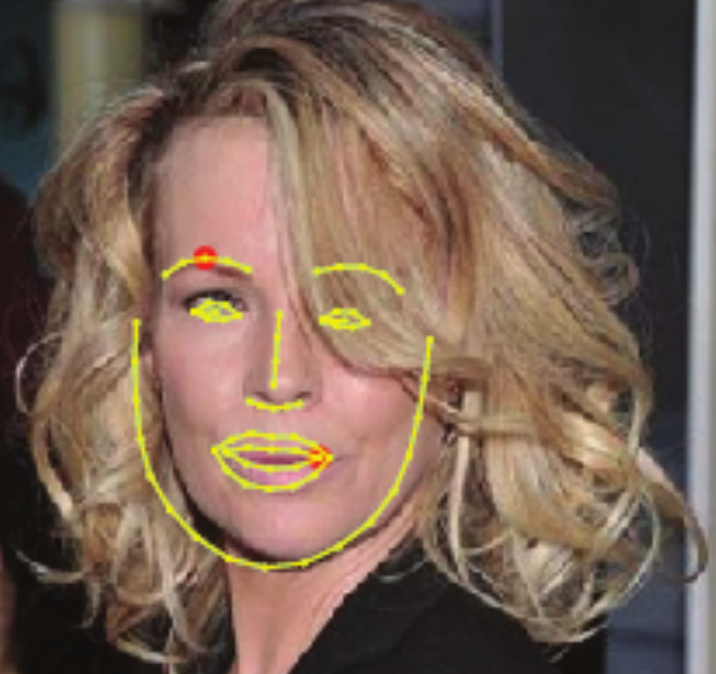}}};
		\node[right of=b, node distance=1.9cm] (c) {\reflectbox{\includegraphics[trim=30 100 20 30,clip,width=0.17\linewidth]{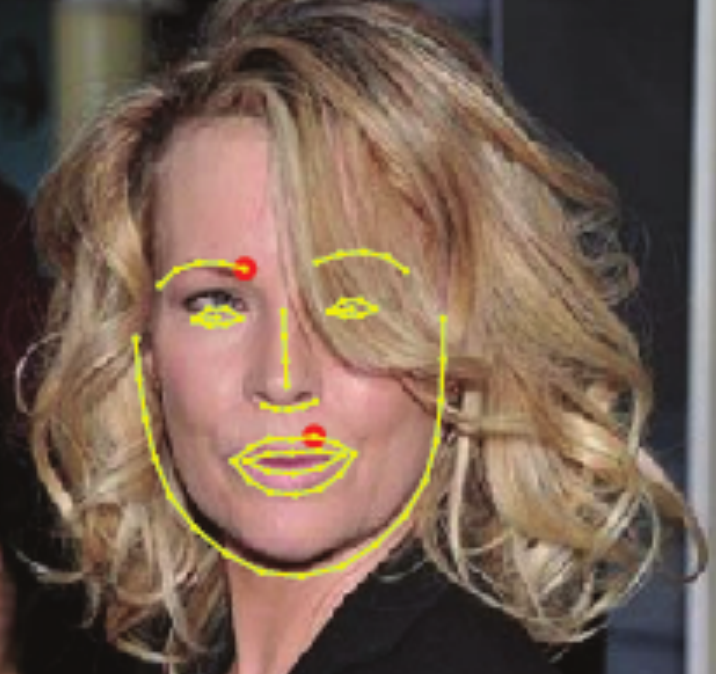}}};
		\node[right of=c, node distance=1.9cm] (d) {\includegraphics[trim=30 100 20 30,clip,width=0.17\linewidth]{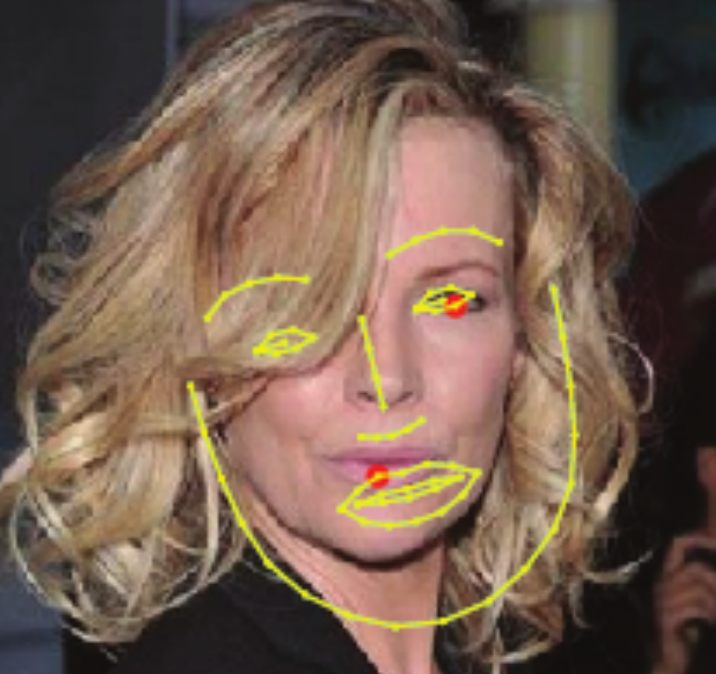}};
		\node[right of=d, node distance=1.9cm] (e) {\includegraphics[trim=30 100 20 30,clip,width=0.17\linewidth]{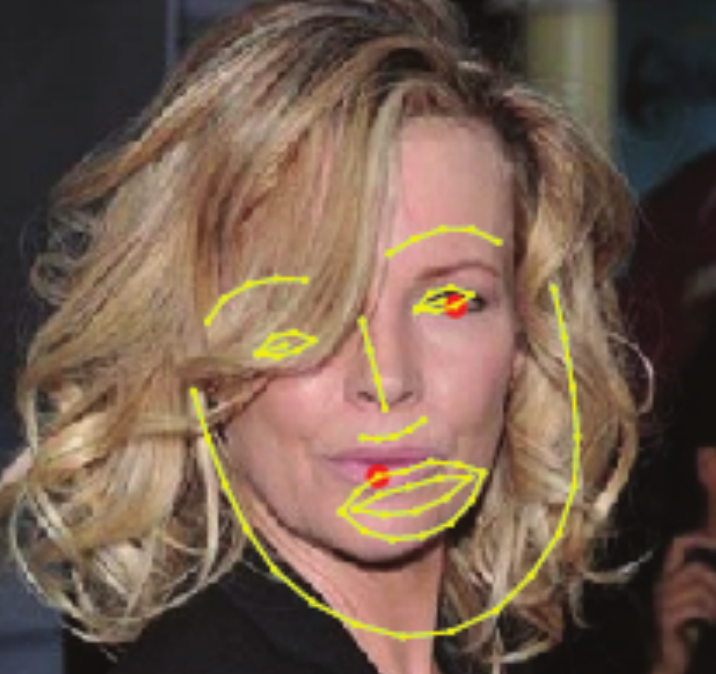}};
		\node[below of=a, node distance=1.8cm] (f) {\includegraphics[trim=30 100 20 30,clip,width=0.17\linewidth]{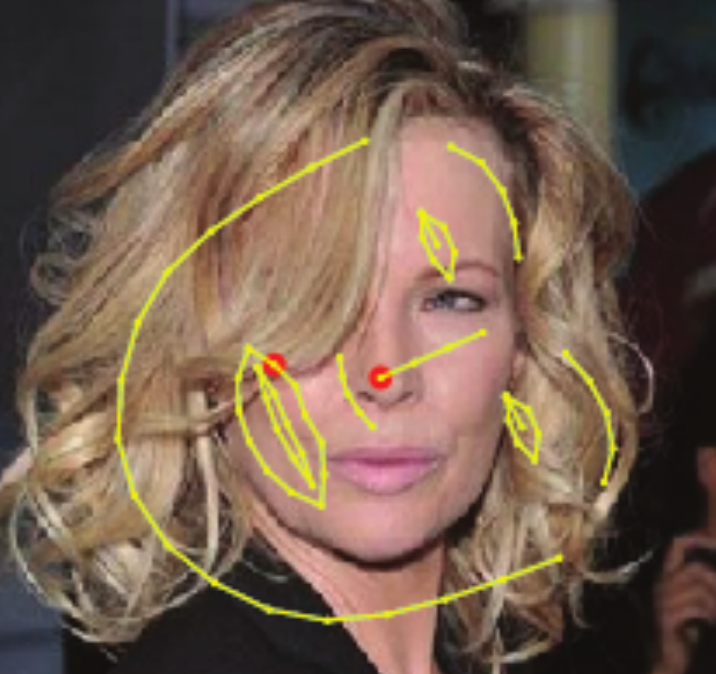}};
		\node[below of=b, node distance=1.8cm] (g) {\reflectbox{\includegraphics[trim=30 100 20 30,clip,width=0.17\linewidth]{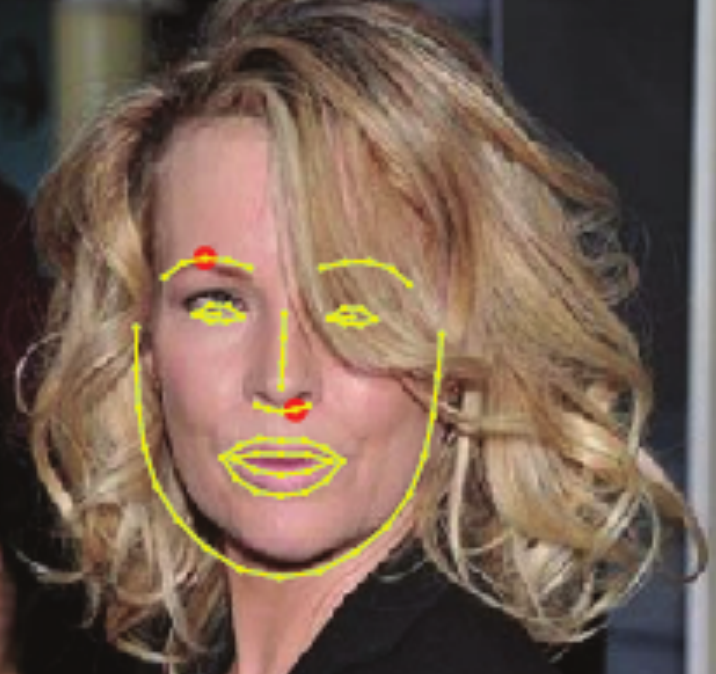}}};
		\node[below of=c, node distance=1.8cm] (h) {\includegraphics[trim=30 100 20 30,clip,width=0.17\linewidth]{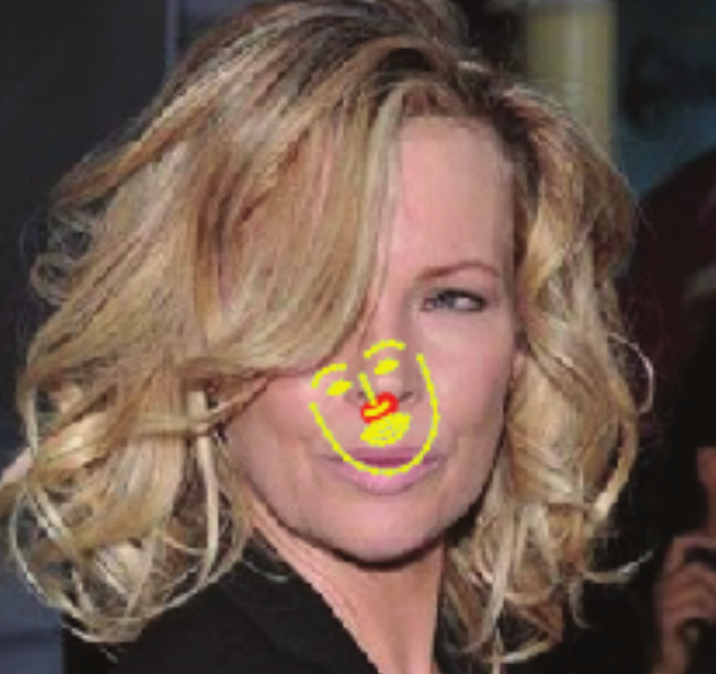}};
		\node[below of=d, node distance=1.8cm] (i) {\includegraphics[trim=30 100 20 30,clip,width=0.17\linewidth]{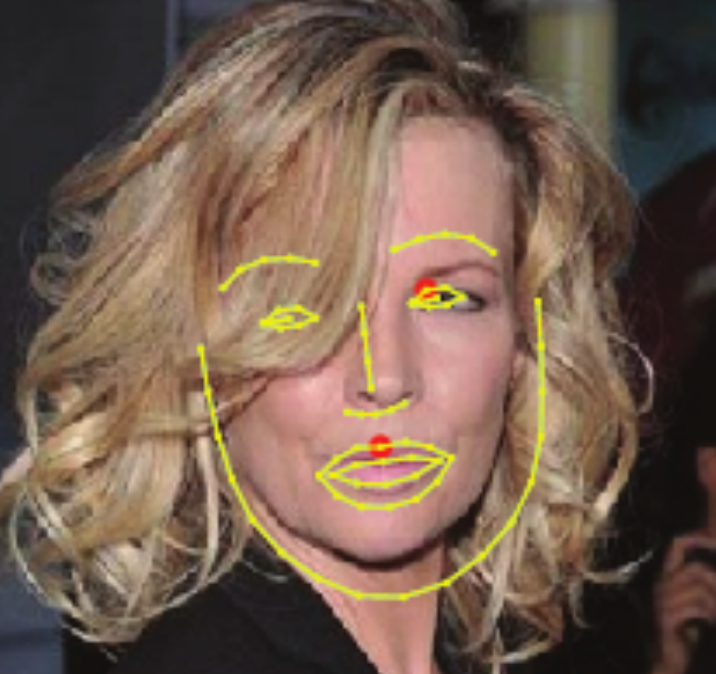}};
		\node[below of=e, node distance=1.8cm] (j) {\reflectbox{\includegraphics[trim=30 100 20 30,clip,width=0.17\linewidth]{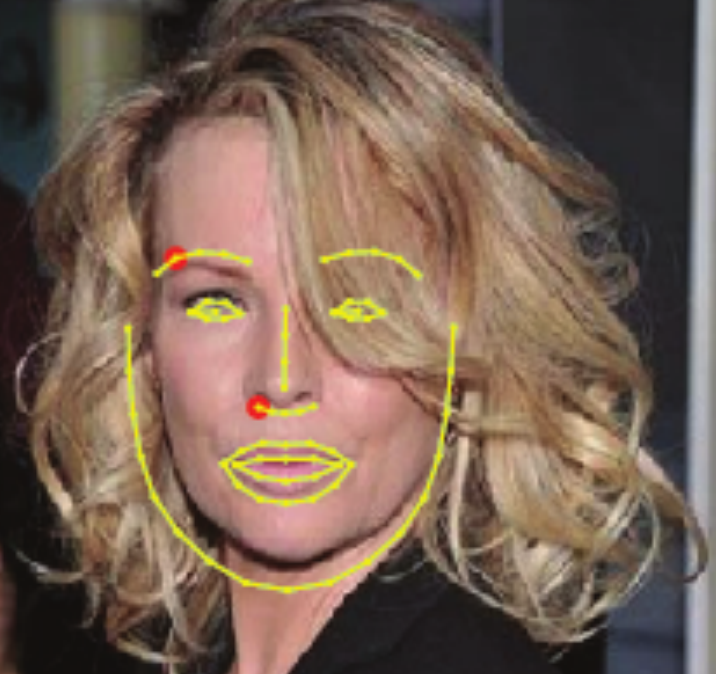}}};
	\end{tikzpicture}
	\caption{}
	\end{subfigure}
	\begin{subfigure}[b]{0.2\linewidth}
		\centering
		\reflectbox{\includegraphics[trim=30 100 20 30,clip,width=1\linewidth]{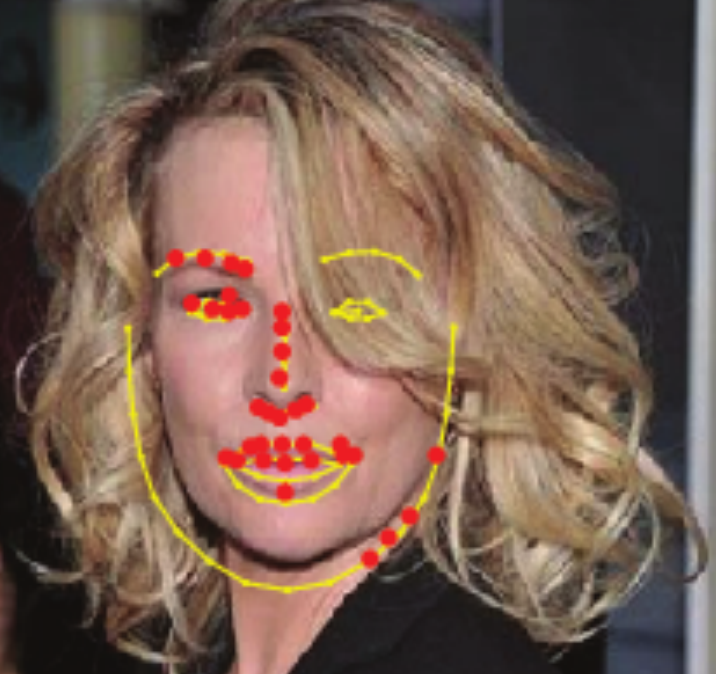}}
		\caption{}
	\end{subfigure}
	\begin{subfigure}[b]{0.2\linewidth}
		\centering
		\reflectbox{\includegraphics[trim=30 100 20 30,clip,width=1\linewidth]{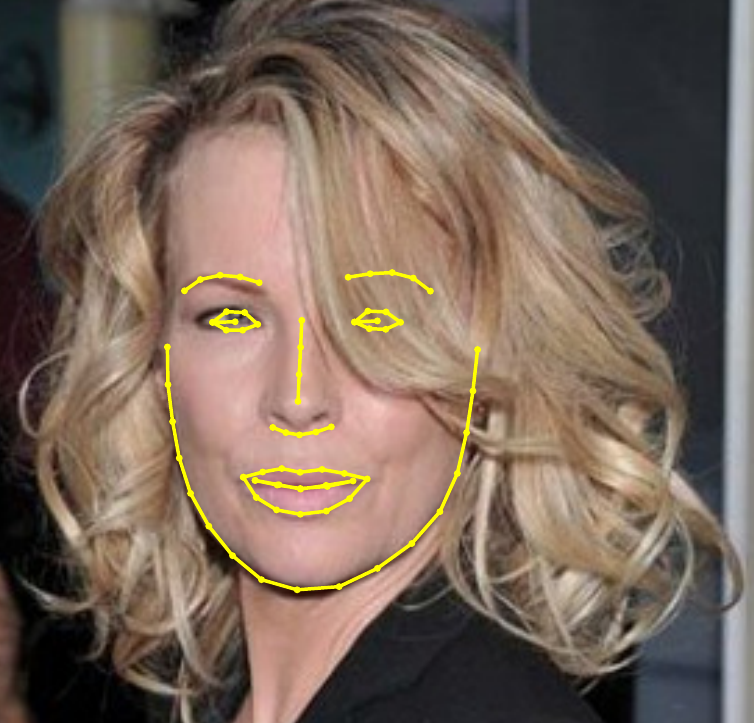}}
		\caption{}
	\end{subfigure}
	\caption{Hypothesis generation, evaluation and shape hallucination (a) Hypotheses generated over the iterations. Two landmarks (red dots) are randomly selected to estimate the scale, rotation, and translation parameters. (b) The nearest $\frac{N}{2}$ landmarks are selected to be inliers (red dots). (c) Hallucinated shape from the selected landmarks.}
	\label{fig:hypothesis}
\end{figure*}

As a reminder, let the number of landmarks be $N$. Assuming that at least half of the landmarks are visible, up to $\frac{N}{2}$ landmarks can be hypothesized to be visible in our framework. However, the hypothesis space of landmark visibilities is huge and becomes even larger when finding the correct set of candidate landmarks that are true positives and are visible. Searching this huge hypothesis space is intractable. We propose a coarse-to-fine approach to search over this space and find the best combination of candidate landmarks to align the shape. The PDM parameter $\mathbf{\Theta} = \{\mathbf{s},\mathbf{R},\mathbf{t},\mathbf{q}\}$ is progressively inferred by first estimating the geometric transformation parameters $\{\mathbf{s},\mathbf{R},\mathbf{t}\}$ followed by the shape parameter $\mathbf{q}$. Figure \ref{fig:hypothesis} shows an example illustrating our hypothesis generation, evaluation and shape hallucination stages.
\begin{enumerate}
	\item \textbf{Geometric Transformation:} The face is first aligned to the mean facial shape by estimating the scale, rotation and translation parameters.
	\item \textbf{Subset selection:} From the geometrically transformed set of candidate landmark estimates, a subset of the landmarks are selected to generate a shape hypothesis.
	\item \textbf{Shape Hallucination:} From a subset of landmarks hypothesized as visible the shape parameter is estimated and facial shape is hallucinated.
\end{enumerate}

\noindent\textbf{Geometric Transformation:} For a given shape model, the geometric transformation parameters $\{\mathbf{s},\mathbf{R},\mathbf{t}\}$ are estimated from two landmark detections associated with two different landmarks. Since the ``detection confidence" of the landmark detectors themselves are not reliable, we do not rely on them for deterministically selecting ``good" landmark detections. Instead, we resort to randomly sampling enough hypotheses such that at least one of the samples consists of ``good" detections. The sampling based nature of our hypotheses-and-test approach for occlusion reasoning optimizes ERCLM to minimize the worst case error due to occlusions (i.e., catastrophic alignment failures), instead of average case error.

\begin{figure}[tbhp]
\captionsetup{font=small}
	\centering
	\includegraphics[scale=0.5]{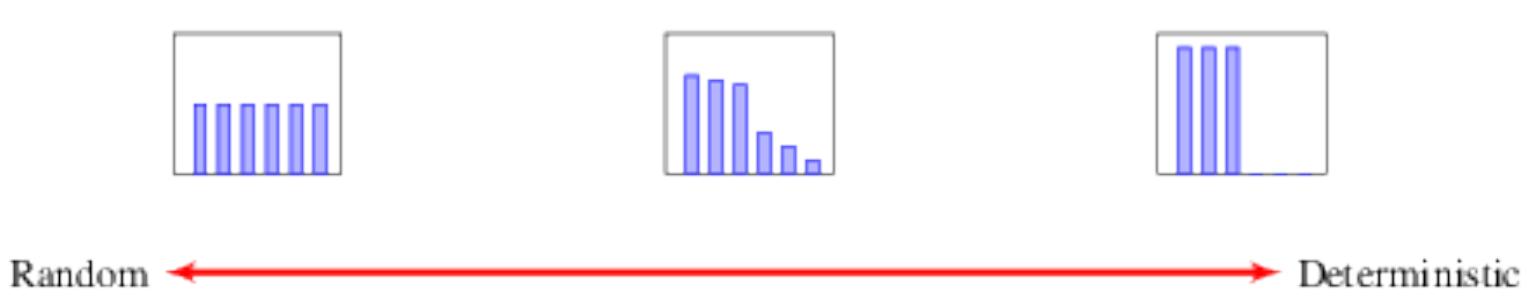}
	\caption{Sampling distributions for hypothesis generation.}
	\label{fig:sampling}
\end{figure}
Selecting the points by sampling randomly, via Random Sample Consensus (RANSAC)~\cite{fischler1981random}, from the landmark detection pool is equivalent to sampling from a uniform distribution over the hypothesis space. This results in the evaluation of a very large number of hypotheses for a given probability of sampling a ``good" hypothesis. However, by selecting the points to include landmarks with high confidence, fewer hypotheses can be evaluated to find a ``good" hypothesis with high probability. Therefore, for efficiency, we bias the samples by sampling from a probability distribution that is proportional to the local landmark detector confidence.

We use this scheme both for selecting the landmark indices as well as to select the true positives from the associated candidate landmarks i.e., we have a total of $N+1$ sampling distributions, one distribution for each landmark index (over detections for the associated landmark) and one over the landmark indices. Figure \ref{fig:sampling} shows the range of possible sampling distributions with the uniform distribution at one end of the spectrum and a deterministic sampling distribution (greedy selection) at the other end of the spectrum while the distribution in the middle corresponds to the one using detector confidences.

\noindent\textbf{Subset Selection:} The crude facial shape estimated from the geometric alignment is evaluated in terms of its ability to ``explain away" the remaining landmarks by a ``mismatch degree" metric. The ``mismatch degree" ($d$) is defined as the median Mahalanobis distance between the transformed shape and the observed landmarks:
\begin{equation}
	d = median(e(\mathbf{x}^D_{\mathcal{F}(1)},Y^1),\dots,e(\mathbf{x}^D_{\mathcal{F}(N)},Y^N))
	\label{eq:median}
\end{equation}
\begin{equation}
	\mathcal{F}(i) = \argmin_{k} E(\mathbf{x}^D_{i,k},Y^i)
	\label{eq:correct_landmarks}
\end{equation}
\begin{equation}
	E(\mathbf{x}^D_{i,k},Y^i) = min(e(\mathbf{x}^D_{i,k},y^i_1),\dots,e(\mathbf{x}^D_{i,k},y^i_{M_i}),\inf)
\end{equation}
\begin{equation}
	e(\alpha,\beta) = \sqrt{(\alpha-\beta)^T\Delta^{-1}_i(\alpha-\beta)}
\end{equation}
\noindent where $\mathbf{x}^{D}_{i,k}$ is the $k$-th hallucinated landmark of $D_i$ (Eq. \ref{eq:DPDM}), $Y^i=\{y^i_1,\dots,y^i_{M_i}\}$ is the set of $M_i$ candidate landmarks associated with the $i$-th landmark and $\Delta_i$ is the covariance matrix describing the distribution of the $i$-th landmark and is estimated from the training data. In Eq. \ref{eq:shape}, given $\{n,m\}$, the landmark selection indicator function $\mathcal{F}$ is computed by Eq. \ref{eq:correct_landmarks}. The above steps are iterated up to a maximum number of hypotheses evaluations and the best hypothesis with the lowest ``mismatch degree" $d$ is found. In our experiments, for most images, 2000 hypotheses evaluations were sufficient to find a set of correct landmark candidates.

For the best hypothesis that is selected, the closest $\frac{N}{2}$ landmark detections associated to different $\frac{N}{2}$ landmarks are selected and a shape is hallucinated using Eq. \ref{eq:hallucinate}. However, the fact that the correct facial shape can be hallucinated using only the nearest $\frac{N}{2}$ candidate landmarks is a necessary but not a sufficient condition. In practice, the selected set may consist of landmarks which are far from the hypothesized positions and may result in an incorrect facial shape estimate. To only select the appropriate landmarks for shape hallucination we filter them using representative exemplar facial shapes (obtained by clustering normalized exemplar shapes) from the training set. This procedure works as follows: from among the set of representative exemplar facial shapes (cluster centers) find an exemplar shape with the lowest mean error between the landmarks and the exemplar shape and find a new set of landmarks within a distance threshold. 

Our approach, unlike most other approaches, does not depend solely on detection confidences for occlusion reasoning. It instead leverages both the discriminative appearance model (detection confidence) and the generative shape model (``mismatch degree") to determine the unoccluded detections. Due to the nature of our randomized hypotheses generation and evaluation, and exemplar filtering process, even high confidence detections may be interpreted as occluded (outliers) if the observation lies outside the shape space. Similarly, even low confidence detections can possibly be interpreted as unoccluded (inliers) if they fall within the shape space. This also results in our occlusion labeling being asymmetrical i.e., the selected landmarks are likely unoccluded but the non-selected landmarks could either be occluded or non-salient. The non-selected points serve as a proxy for occluded landmarks.

\noindent\textbf{Shape Hallucination:} Given a hypothesis with the selected landmark candidates and their occlusion labels, $O = \{o_1,\dots,o_N\}$, where $o_i \in \{0,1\}$ (setting the landmark occlusion label i.e., $o_i=1$ if the $i$-th landmark is hypothesized to be visible), we use the Convex Quadratic Curve Fitting method introduced in~\cite{wang2008enforcing} to compute the shape parameter $\mathbf{q}$ in Eq. \ref{eq:shapemodel} by a closed form expression.
\begin{eqnarray}
	\mathbf{q} = (\Phi^{\mathbf{T}}\mathbf{A}\Phi)^{-1}\Phi^{\mathbf{T}}\mathbf{b}
	\label{eq:hallucinate}
\end{eqnarray}
\noindent where \[\mathbf{A}=\left[\begin{array}{ccc} o_1\mathbf{A}_1 & \cdots & 0 \\ \vdots & \ddots & \vdots\\ 0 & \cdots & o_N\mathbf{A}_N \end{array}\right]\label{cf_D} \mbox{ and } \mathbf{b}= \left[\begin{array}{c}o_1\mathbf{b}_1 \\ \vdots \\ o_N\mathbf{b}_N \end{array}\right]\] 

and $\mathbf{A}_i$ and $\mathbf{b}_i$ are computed using Eq. \ref{eq:meanshift}. This shape parameter $\mathbf{q}$ is used to hallucinate the full facial shape.

\begin{figure*}[!ht]
\captionsetup{font=small}
	\centering
	\begin{subfigure}[b]{0.15\linewidth}
		\reflectbox{\includegraphics[trim=30 100 20 30,clip,width=1\textwidth]{figs/1_2_1_hallucinated.pdf}}
		\caption{$0^{\circ}$}
	\end{subfigure}
	\begin{subfigure}[b]{0.15\linewidth}
		\reflectbox{\includegraphics[trim=30 100 20 30,clip,width=1\textwidth]{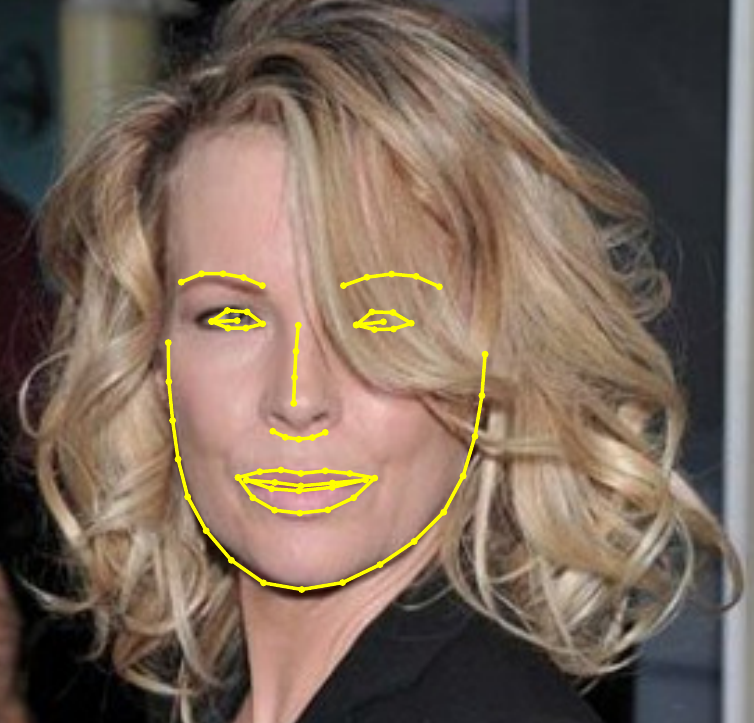}}
		\caption{$15^{\circ}$}
	\end{subfigure}
	\begin{subfigure}[b]{0.15\linewidth}
		\reflectbox{\includegraphics[trim=30 100 20 30,clip,width=1\textwidth]{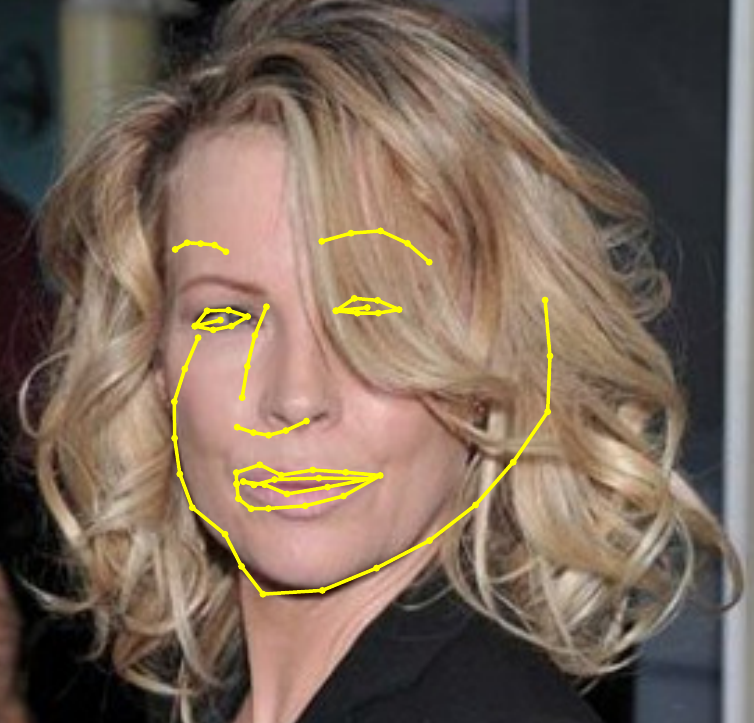}}
		\caption{$45^{\circ}$}
	\end{subfigure}
	\begin{subfigure}[b]{0.15\linewidth}
		\reflectbox{\includegraphics[trim=30 100 20 30,clip,width=1\textwidth]{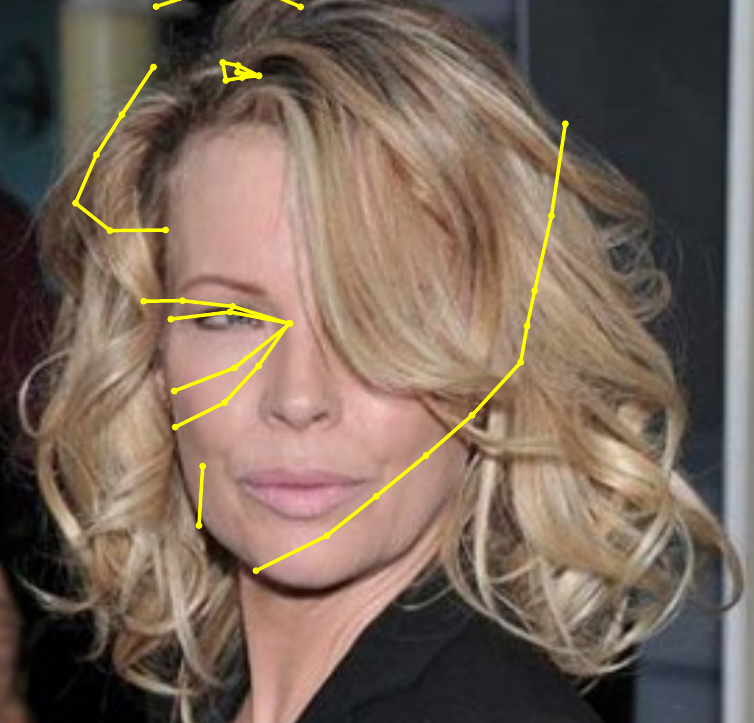}}
		\caption{$75^{\circ}$}
	\end{subfigure}
	\begin{subfigure}[b]{0.15\linewidth}
		\reflectbox{\includegraphics[trim=30 100 20 30,clip,width=0.99\textwidth]{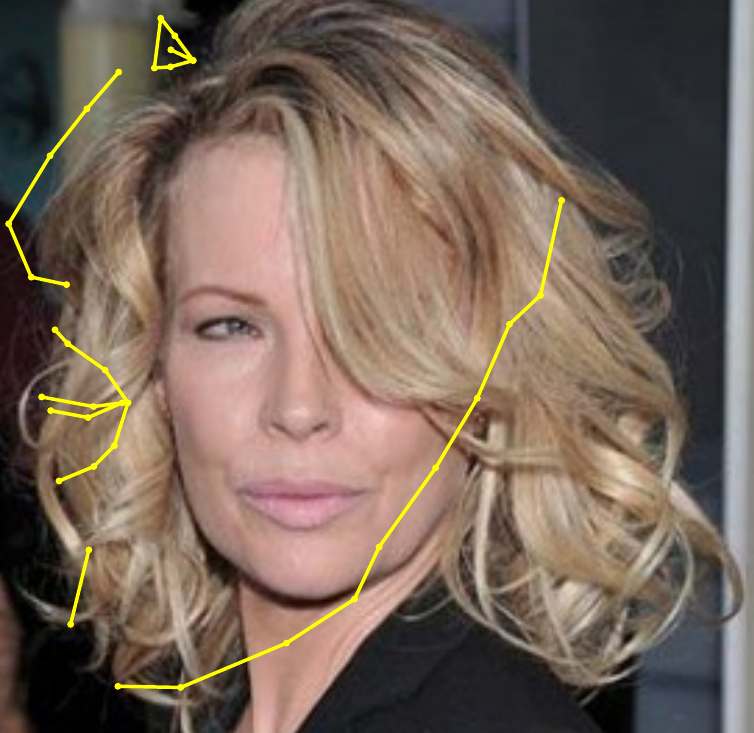}}
		\caption{$90^{\circ}$}
	\end{subfigure}
	\begin{subfigure}[b]{0.15\linewidth}
		\includegraphics[trim=30 100 20 30,clip,scale=0.38]{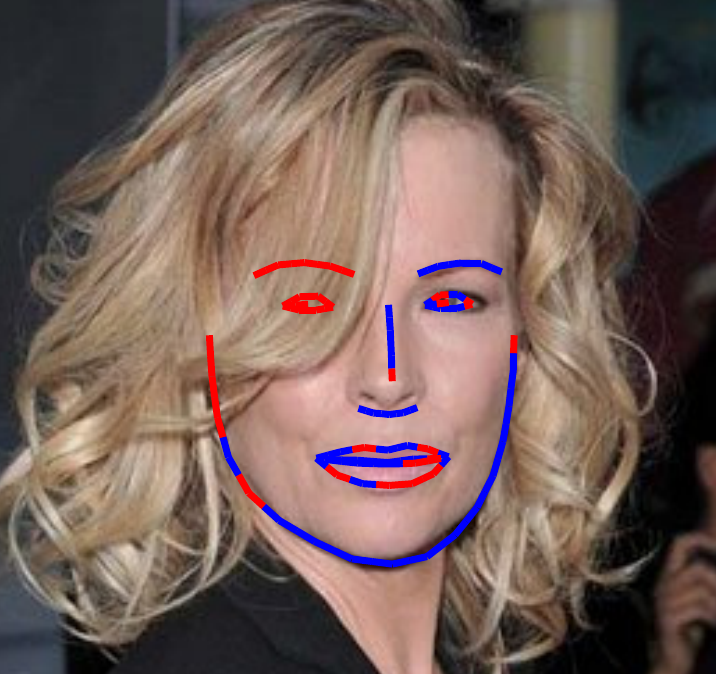}
		\caption{Refined}
	\end{subfigure}
	\caption{Hallucinated shapes from different models (a)-(e) $0^{\circ}$ to $90^{\circ}$. In this example, the first shape model is chosen as the best hallucinated shape, and (f) final refined shape, landmarks predicted as visible and occluded are shown in blue and red, respectively.}
	\label{fig:hallucinate}
\end{figure*}
\subsubsection{Shape Model Evaluation and Selection}
For each given facial pose $n$ and expression $m$ and the corresponding shape model $\{\mathbf{\bar{x}}_i(n,m), \Phi_i(n,m)\}$, the correct landmarks, $\mathcal{F}$, are estimated from Eq. \ref{eq:correct_landmarks} and the shape parameters, $\mathbf{q}$, from Eq. \ref{eq:hallucinate} to hallucinate a shape. Figure \ref{fig:hallucinate} shows some of the hallucinated shapes spanning pose $0^{\circ}$ to $90^{\circ}$. These shapes are evaluated to select the pose and expression mode that best fits the observed shape. For the $n$-th pose model and $m$-th expression model, let $V^n_m$ be the number of inliers and let $E^n_m$ be the mean error of inliers. The pose model is chosen by Eq. \ref{eq:choose_pose} (maximizing the number of inliers while minimizing the mean error) and the expression model by Eq. \ref{eq:choose_cluster} (maximizing the number of inliers).
\begin{equation}
	n_0 = \argmax_{n} \sum_{m=1}^{E(n)} \frac{V^n_m}{E^n_m}
	\label{eq:choose_pose}
\end{equation}
\noindent where the $E(n)$ is the number of shape clusters over the $n$-th facial angle. From the set of hallucinated shape of $n_0$-th facial angle, a best shape is chosen as follows:
\begin{equation}
	m_0 = \argmax_{m} V^{n_0}_m
	\label{eq:choose_cluster}
\end{equation}

\subsubsection{Shape Refinement}
To refine the shape alignment result, the local landmark detectors responses are re-calculated with the scale, rotation and translation parameters estimated from the shape model selected ($S_{0}$ with parameters $\{n_0,m_0\}$) in the previous stage. During the shape refinement process we add more inliers to the set of landmarks which were used to hallucinate the facial shape $S_{0}$. To select the inliers we adopt the idea of finding peaks along the tangent line of each landmark~\cite{zhou2003bayesian}. In our model, the tangent-line-search is adopted only for the contour features, such as jawline, eye-brows, lips, and nose bridge features. For each landmark, the highest peak on the tangent search line, within a search region, is found and included in our inlier set if the peak value is above a given threshold. The final shape is hallucinated using this new set of inlier landmarks.

For the $i$-th landmark, let $\mathbf{x}^m_i$, $\mathbf{x}^p_i$, and $\mathbf{x}^h_i$ be the positions of the mean shape of the chosen facial pose and expression model, the detected landmark locations, and the hallucinated shape. Then the parameters $\mathbf{A}$ and $\mathbf{b}$ required to estimate the shape parameters $\mathbf{q}$ in Eq. \ref{eq:hallucinate} are defined as follows:
\vspace{2mm}
$\mathbf{A}=\left[\begin{array}{ccc} \mathbf{A}^{'}_1 & \cdots & 0 \\ \vdots & \ddots & \vdots\\ 0 & \cdots & \mathbf{A}^{'}_N \end{array}\right]\label{cf_D}$ and $\mathbf{b}= \left[\begin{array}{c}\mathbf{b}^{'}_1 \\ \vdots \\ \mathbf{b}^{'}_N \end{array}\right]$ where, $\mathbf{A}^{'}_i = \left\{
  \begin{array}{lr}
  	o_i\mathbf{I}_{2\times2} & : {\mathbf{x}_i \in \Omega}\\
  	o_i\mathbf{A}_i & : {\mathbf{x}_i \in \Upsilon}\\
  \end{array}\right.$ and 

 \[\mathbf{b}^{'}_i = \left\{
  \begin{array}{lr}
	\mathbf{x}^p_i-\mathbf{x}^m_i & : {o_i=1 \mbox{ and } \mathbf{x}_i \in \Upsilon} \\
    \mathbf{b}_i & : {o_i=1 \mbox{ and } \mathbf{x}_i \in \Omega} \\
    \mathbf{x}^h_i-\mathbf{x}^m_i & : \mbox{otherwise} \\
  \end{array}\right.\]
Figure \ref{fig:hallucinate}(f) shows the refined shape of our running example where landmarks shown in blue are predicted to be visible and those shown in red are deemed to be occluded. Algorithm \ref{alg:algorithm} describes our complete ``Face Alignment Robust to Pose, Expressions and Occlusions" procedure.

\begin{algorithm}[!ht]
 \KwData{Image $I$}
 \KwResult{PDM Parameter $\mathbf{\Theta}$, Occlusion Labels $O$}
 Run Face Detector\;
 \For{$face = 1:\# faces$}{	
 	\For{$pose = 1:n$}{
 		Run Landmark Detectors\;
 		Estimate $\{\mathbf{A}_1,\dots\,\mathbf{A}_N\}$ and $\{\mathbf{b}_1,\dots,\mathbf{b}_N\}$ from Eq. \ref{eq:meanshift}\;
 		\While{$\#$ hypothesis $\leq$ MAX-ITER}{
  			Sample two landmark indices\;
  			Estimate geometric parameters $\{\mathbf{s},\mathbf{R},\mathbf{t}\}$\;
  			Compute ``mismatch degree" (d) from Eq. \ref{eq:median}\;
 		}
 		Select best hypothesis with lowest ``mismatch degree"\;
 		Filter candidate landmarks using exemplar facial shapes\;
 		Estimate shape parameters $\mathbf{q}$ from Eq. \ref{eq:hallucinate}\;
 	}
 	Select best pose ($n_0$) from Eq. \ref{eq:choose_pose}\;
 	Select best expression ($m_0$) from Eq. \ref{eq:choose_cluster}\;
 	Refine facial shape using best selected model parameters\;
 }
 \caption{\scriptsize Face Alignment Robust to Pose, Expressions and Occlusions}
 \label{alg:algorithm}
\end{algorithm}

\section{Experiments and Analysis \label{sec:experiments}}
In this section we describe the experimental evaluation of ERCLM, our proposed pose, expression and occlusion robust face alignment method and many strong face alignment baselines. We compare and demonstrate the efficacy of these face alignment approaches via extensive large scale experiments on many different datasets of face images, both occluded and unoccluded, and spanning a wide range of facial poses and expressions.

\subsection{Datasets}
\noindent \textbf{LFPW: } The Labeled Face Parts in the Wild~\cite{belhumeur2011localizing} consists of images collected from the web and have various expressions, facial poses (excluding profile or near profile faces) and partial occlusions. The original dataset contained 1132 training images and 300 test images. Unfortunately, many URLs have expired and we were able to download only 776 images from the training subset and 208 images from the testing subset. While the original dataset has 29 annotated landmarks, this dataset was re-annotated with 68 landmarks~\cite{ibug-dataset}.

\noindent \textbf{AFW: } The Annotated Faces In-The-Wild~\cite{zhu2012face} is a dataset with images downloaded from Flickr consisting of 205 images with 468 faces each annotated with 6 landmarks (the center of eyes, tip of nose, the two corners and center of mouth). The images contain cluttered backgrounds with large variations in both face viewpoint and appearance (aging, sunglasses, make-ups, skin color, expression, etc.). Some images from this dataset have been re-annotated with 68 landmarks~\cite{ibug-dataset}.

\noindent \textbf{Helen: } The HELEN dataset~\cite{le2012interactive} is a collection of 2,330 high resolution face portraits downloaded from Flickr with pose, illumination, expression and occlusion variations. While the original dataset is densely annotated with 194 landmarks, this dataset was re-annotated with 68 landmarks~\cite{ibug-dataset}.

\noindent \textbf{IBUG: } IBUG~\cite{ibug-dataset} is a dataset of real-world face images. It consists of 135 images publicly available and taken in highly unconstrained settings with non-cooperative subjects and annotated with 68 landmarks.

\noindent \textbf{300W: } The 300W~\cite{sagonas2015300} is a dataset of real-world face images released as part of a challenge. It consists of 600 indoor and outdoor faces captured under highly unconstrained settings and annotated with 68 landmarks.

\noindent \textbf{COFW: } The Caltech Occluded Faces in the Wild~\cite{burgos2013robust} has faces showing large variations in shape and occlusions due to differences in pose, expression, use of accessories such as sunglasses and hats and interactions with objects (e.g. food, hands, microphones, etc.). It consists of 1,007 images annotated the 29 landmarks positions along with an occluded/unoccluded label.

\subsection{Training}
We learn an ensemble of independent CLMs spanning a wide range of pose and expression variations. Both the local landmark detectors and the facial shape models were trained using a subset of the CMU Multi-PIE~\cite{gross2010multi} dataset, about 10,000 images with manually annotated pose, expression and landmark locations. Each face is annotated with 68 facial landmarks for frontal faces ($-45^{\circ}$ to $45^{\circ}$) and 40 landmarks for profile faces ($45^{\circ}$ to $90^{\circ}$). This dataset was captured in a controlled environment without any facial occlusions but under different illumination conditions over multiple days. 

We trained multiple independent CLMs, both appearance and shape models, spanning $P=5$ pose and $E(n)=2$ expression modes for a total of 10 models. The pose modes correspond to $0^{\circ}\sim 15^{\circ}$, $15^{\circ}\sim 30^{\circ}$, $30^{\circ}\sim 60^{\circ}$, $60^{\circ}\sim 75^{\circ}$, $75^{\circ}\sim 90^{\circ}$, spanning the camera angles from $0^{\circ}$ to $90^{\circ}$ in the dataset. The same local landmark detectors and facial shape models learned from the CMU Multi-PIE dataset are used to align faces across all the other datasets for evaluation.

To train the local landmark detectors, both positive patches of the landmarks and the background patches are harvested from the training images which are normalized by Generalized Procrustes Analysis (GPA). The positive patches\footnote{The width of the face region is normalized to 150 pixels and local patch's size is $35 \times 35$, so each local patch covers almost $\frac{1}{4}$ of the face width.} are centered at the ground-truth landmark locations, and negative patches are sampled in a large region around the ground-truth landmark location. For improved robustness to image rotations, we augment the positive patches by sampling them from $\pm10^{\circ}$ rotated training images as well.

To train the shape models we first normalize the training shapes using GPA~\cite{goodall1991procrustes}. Conventionally all the points in the shape model are used in the normalization process. However, this process can be biased by the distribution of the points. For instance, the mouth region has many more points than the other parts of the face, so conventional GPA shape normalization is biased by the points in the mouth region. To overcome this bias, we use only a few select points to normalize the shapes. For the frontal pose, we use the three least morphable points on the face to normalize the shape, centers of both eyes and the center of the nostril. Similarly, for the profile face pose, we use the center of the visible eye, center of the nostril and the tip of the lip to normalize the shape.
\begin{table}[!ht]
	\captionsetup{font=footnotesize}
	\centering
	\caption{Comparison of the number of eigenvectors that preserve 95\% of the training data.}
	\label{table:procrustes}
	\scalebox{0.75}{
	\begin{tabular}{|c|c|c|c|}
		\hline
		& $0^{\circ}$ face point & $45^{\circ}$ face point & $90^{\circ}$ face point \\
		& (70 points) & (70 points) & (40 points) \\
		\hline
		Conventional GPA & 21 & 19 & 18 \\
		Subset GPA & 17 & 15 & 18 \\
		\hline
		Conventional GPA (dense) & 14 & 12 & 13 \\
		Subset GPA (dense) & 10 & 9 & 13 \\
		\hline
	\end{tabular}}	
\end{table}

Learning the shape models using a subset of the landmarks results in fewer eigenvectors required to preserve 95\% of the training data in comparison to using all the facial landmarks. Table \ref{table:procrustes} shows a comparison of the number of eigenvectors that preserve 95\% of the training data for the conventional GPA normalization and the proposed landmark subset GPA normalization. The results show that 1) the subset GPA normalization can normalize the shape very effectively and 2) the dense point shape provides even further compression.

\subsection{Evaluation}
\textbf{Metrics:} We report the Mean Normalized Landmark Error (MNLE) and face alignment Failure Rate (FR). Errors are normalized with respect to the interocular distance \cite{ibug-dataset} (euclidean distance between the outer corners of the eyes) and we consider any alignment error, defined as the mean error of all the landmarks, above 10\% to be a failure, as proposed in~\cite{dantone2012real}.

\noindent\textbf{Baselines:} We evaluate and compare against many strong face alignment baselines. Deformable parts based model (DPM)\footnote{We use the publicly available implementation using the best performing pre-trained model with 1,050 parts.} proposed by Zhu et.al. \cite{zhu2012face} that is trained using images only from the CMU Multi-PIE dataset. DPM consists of a mixture of trees spanning the entire range of facial pose but does not explicitly model occlusions. We also consider multiple regression based approaches, Explicit Shape Regression (ESR) \cite{cao2012face}, Supervised Descent Method (SDM) \cite{xiong2013supervised} and Robust Cascaded Pose Regression (RCPR) \cite{burgos2013robust} which explicitly models occlusions. We retrain ESR and RCPR using the publicly available implementations using the same face detection bounding boxes at train and test time. To train RCPR with occlusion labels, we generate occluded faces and labels virtually following the procedure in \cite{ghiasi2014occlusion}. Lastly since there is no publicly available code for training SDM, we simply use the executable made available by the authors.

\begin{figure*}[!ht]
	\captionsetup{font=small}
	\begin{subfigure}[b]{0.24\linewidth}
		\centering
		\includegraphics[width=0.95\linewidth]{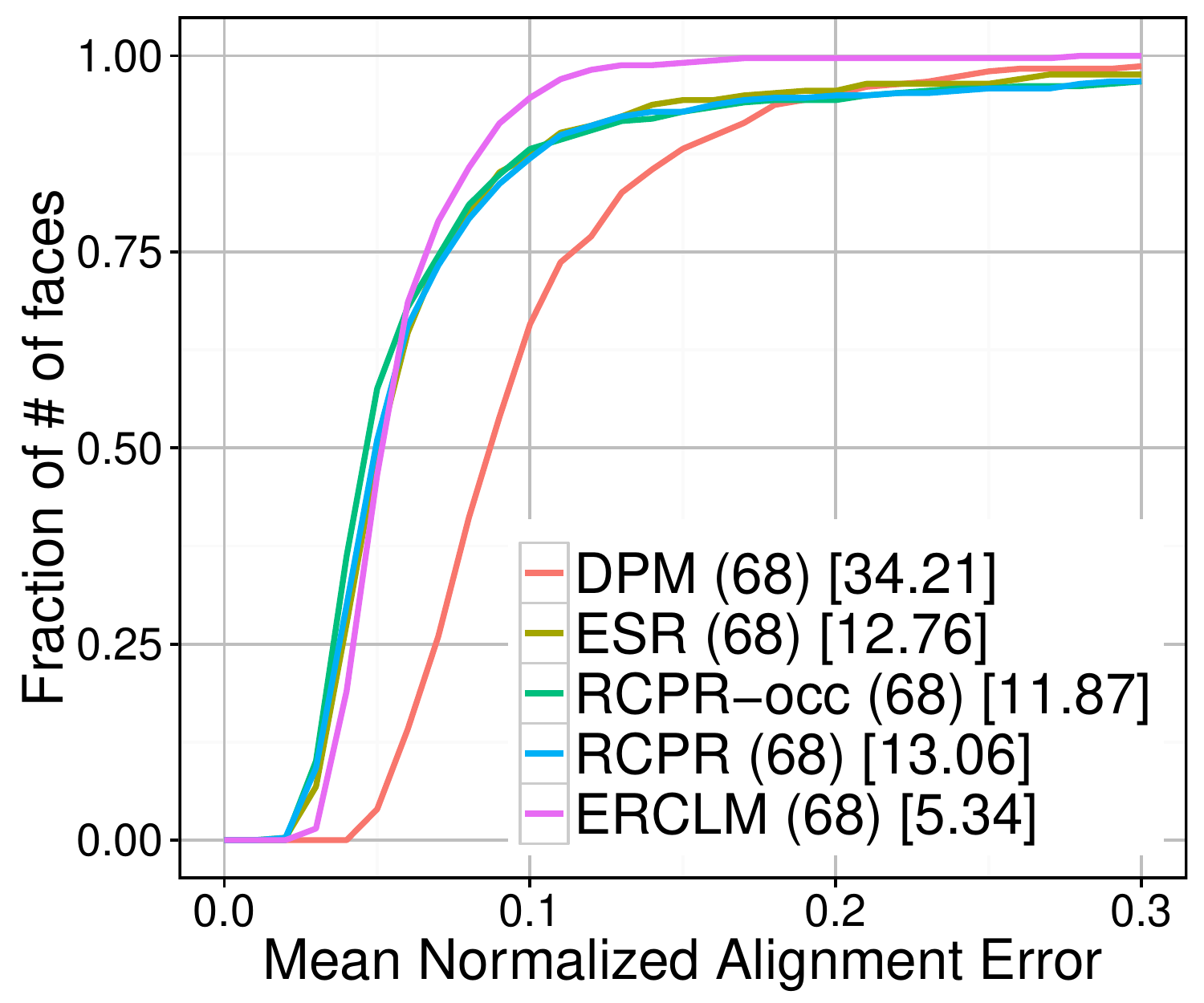}\\
		\includegraphics[width=0.95\linewidth]{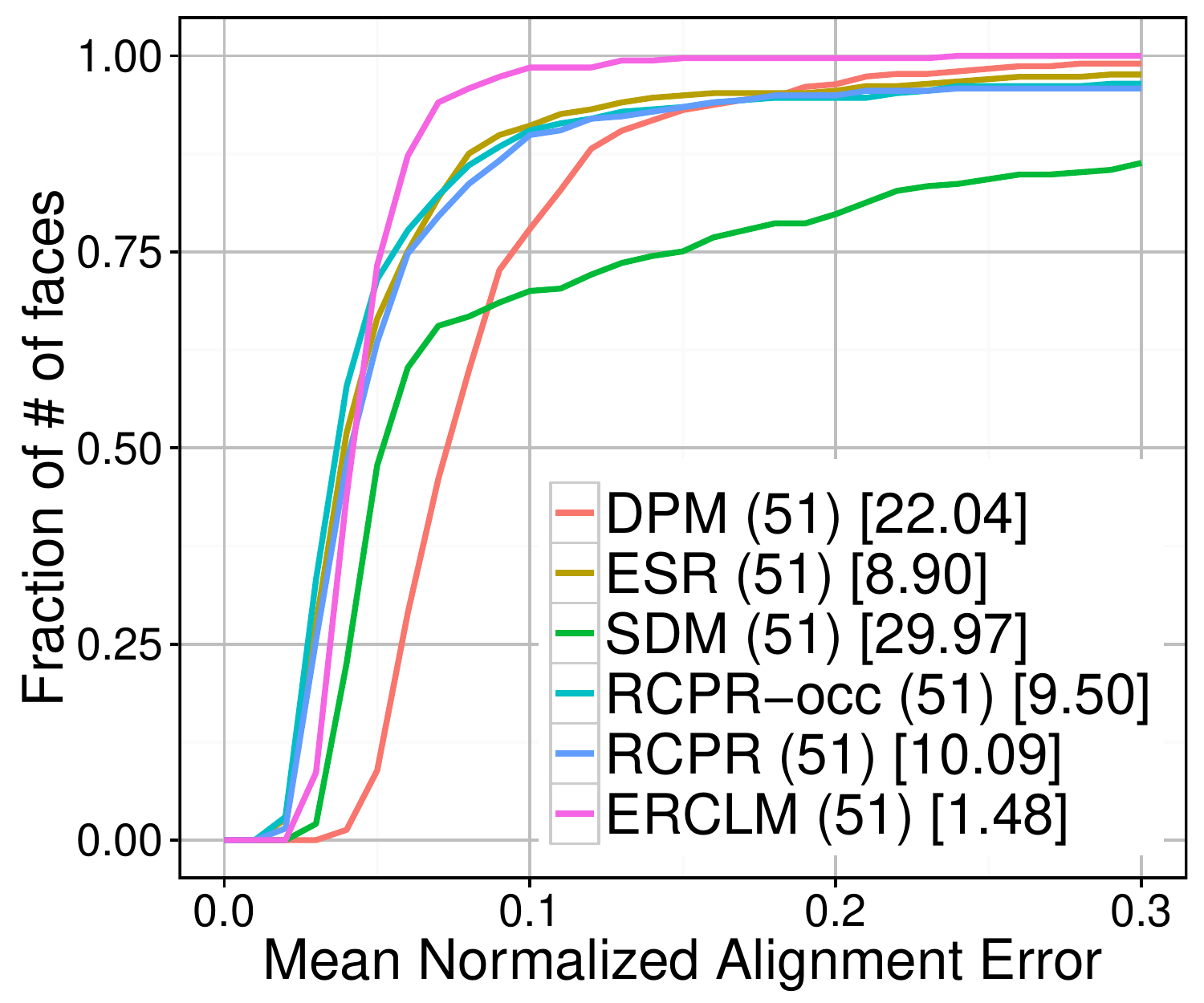}
		\caption{AFW}
	\end{subfigure}
	\begin{subfigure}[b]{0.24\linewidth}
		\centering
		\includegraphics[width=0.95\linewidth]{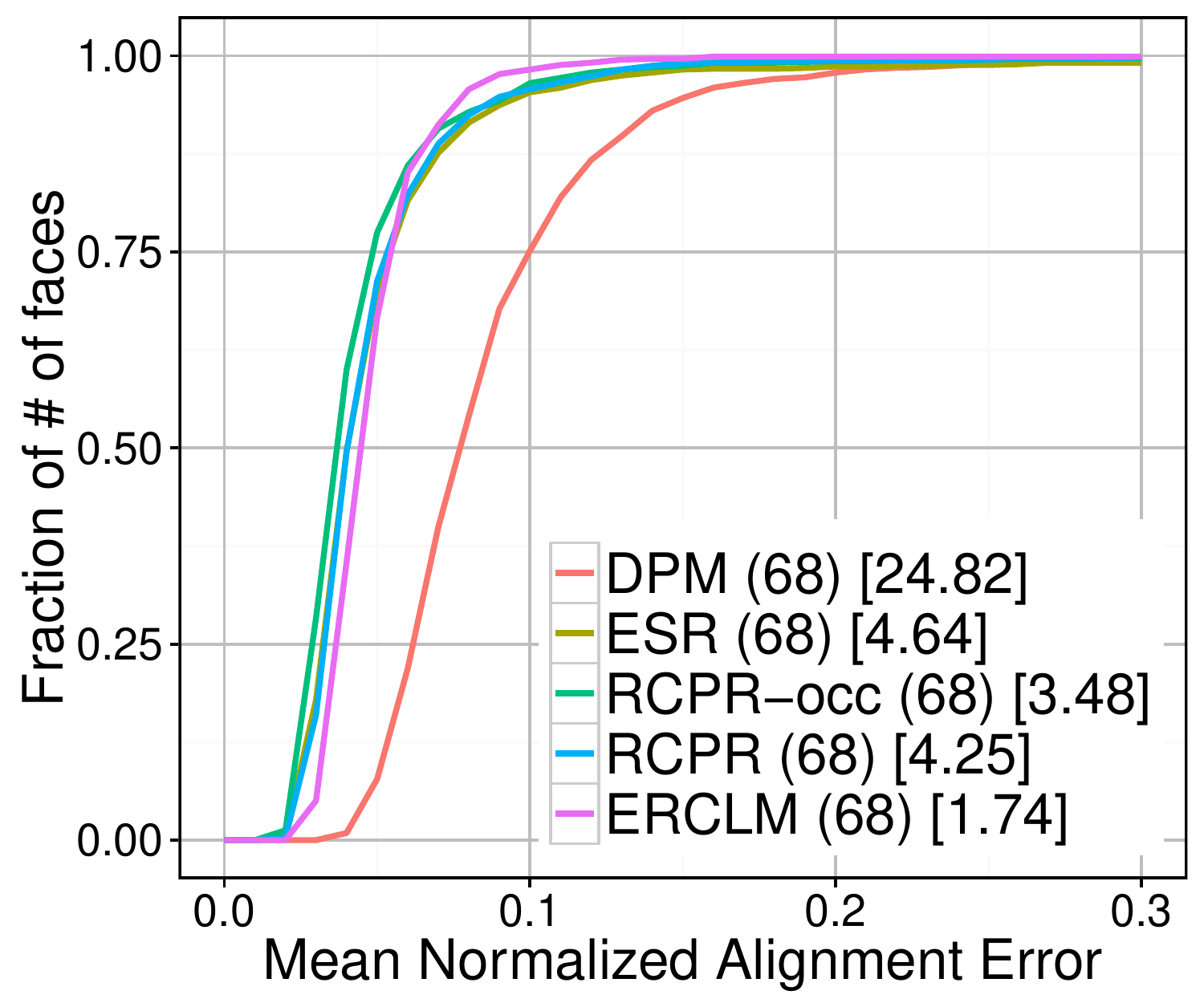}\\
		\includegraphics[width=0.95\linewidth]{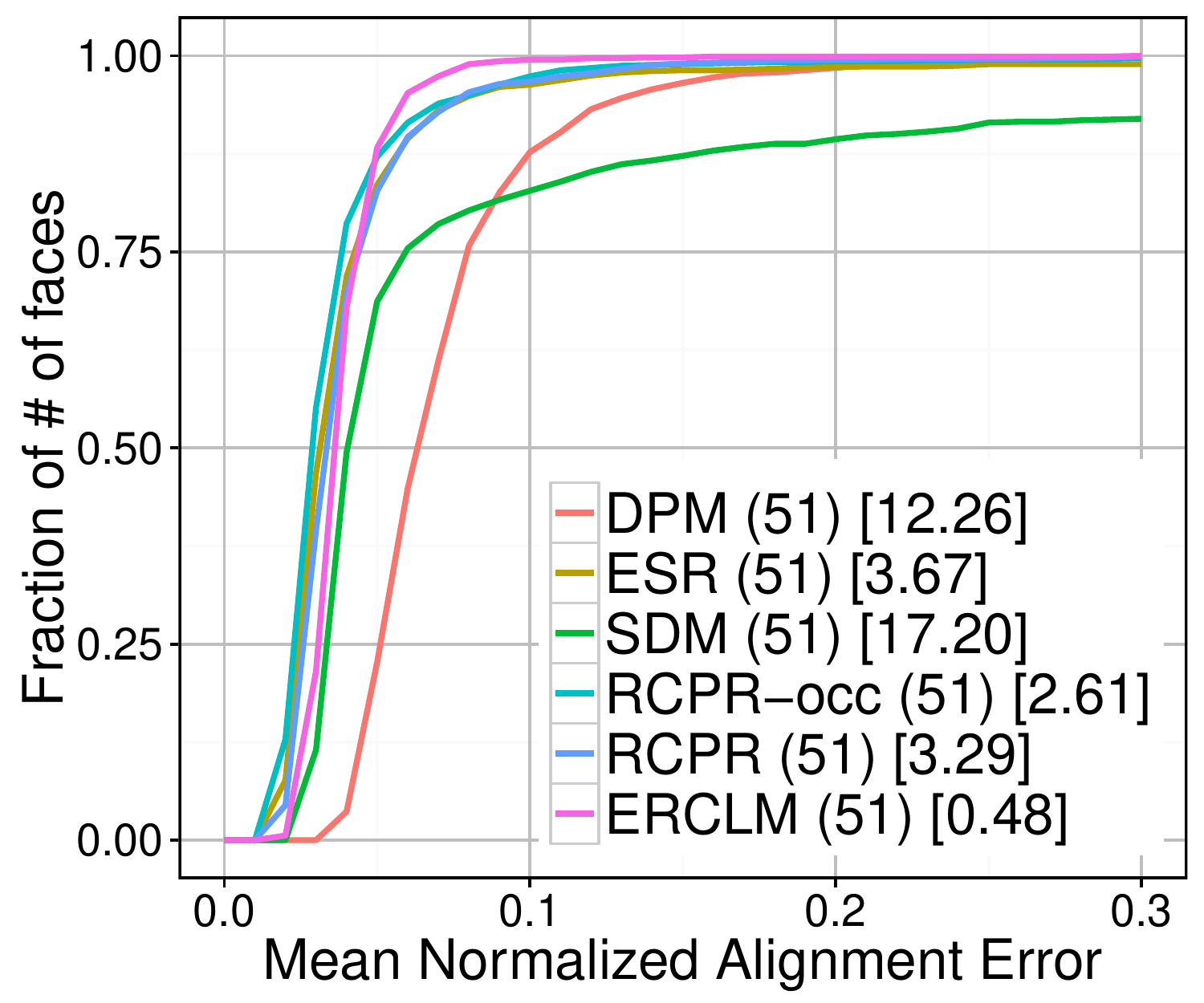}
		\caption{LFPW}
	\end{subfigure}
	\begin{subfigure}[b]{0.24\linewidth}
		\centering
		\includegraphics[width=0.95\linewidth]{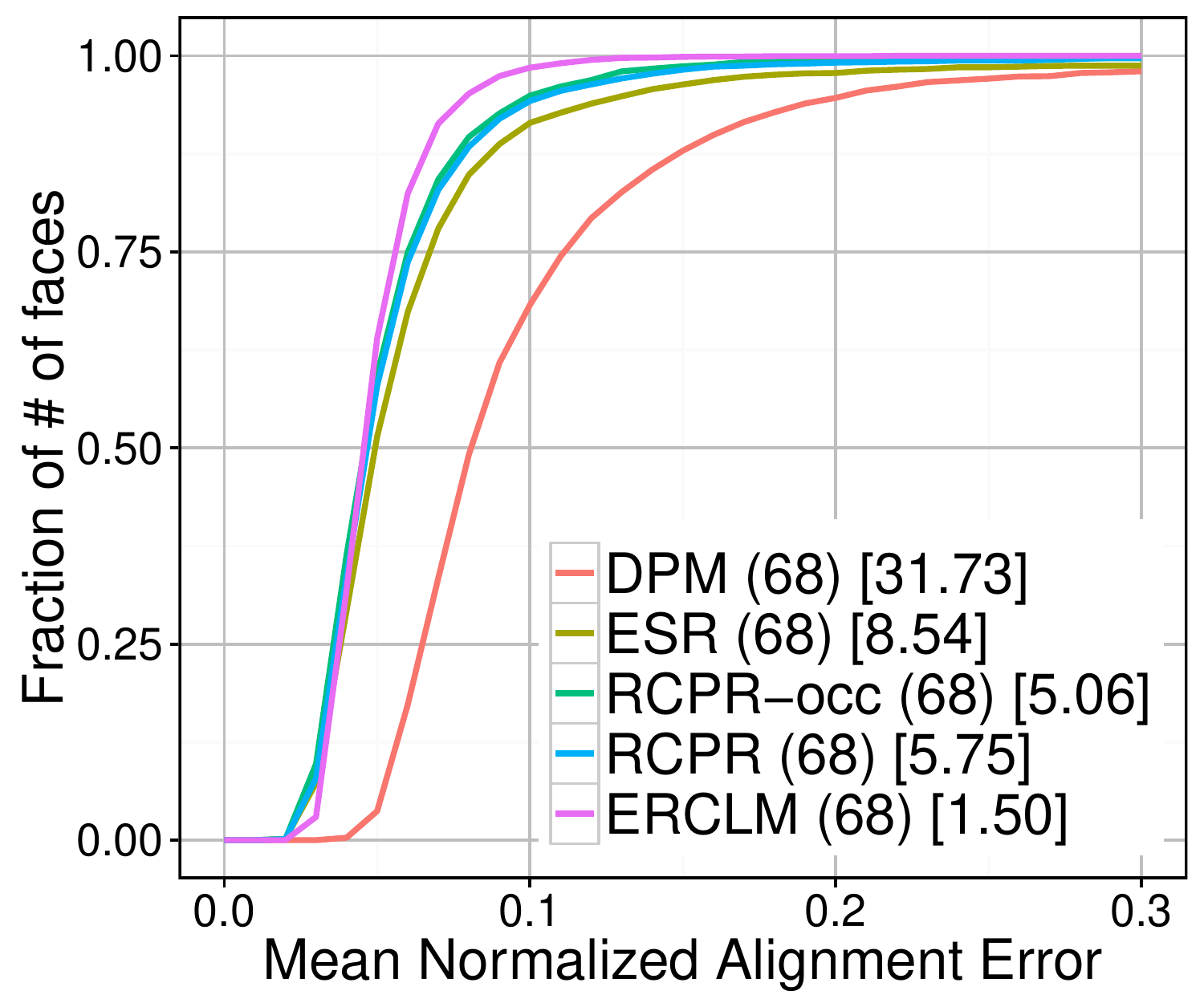}\\
		\includegraphics[width=0.95\linewidth]{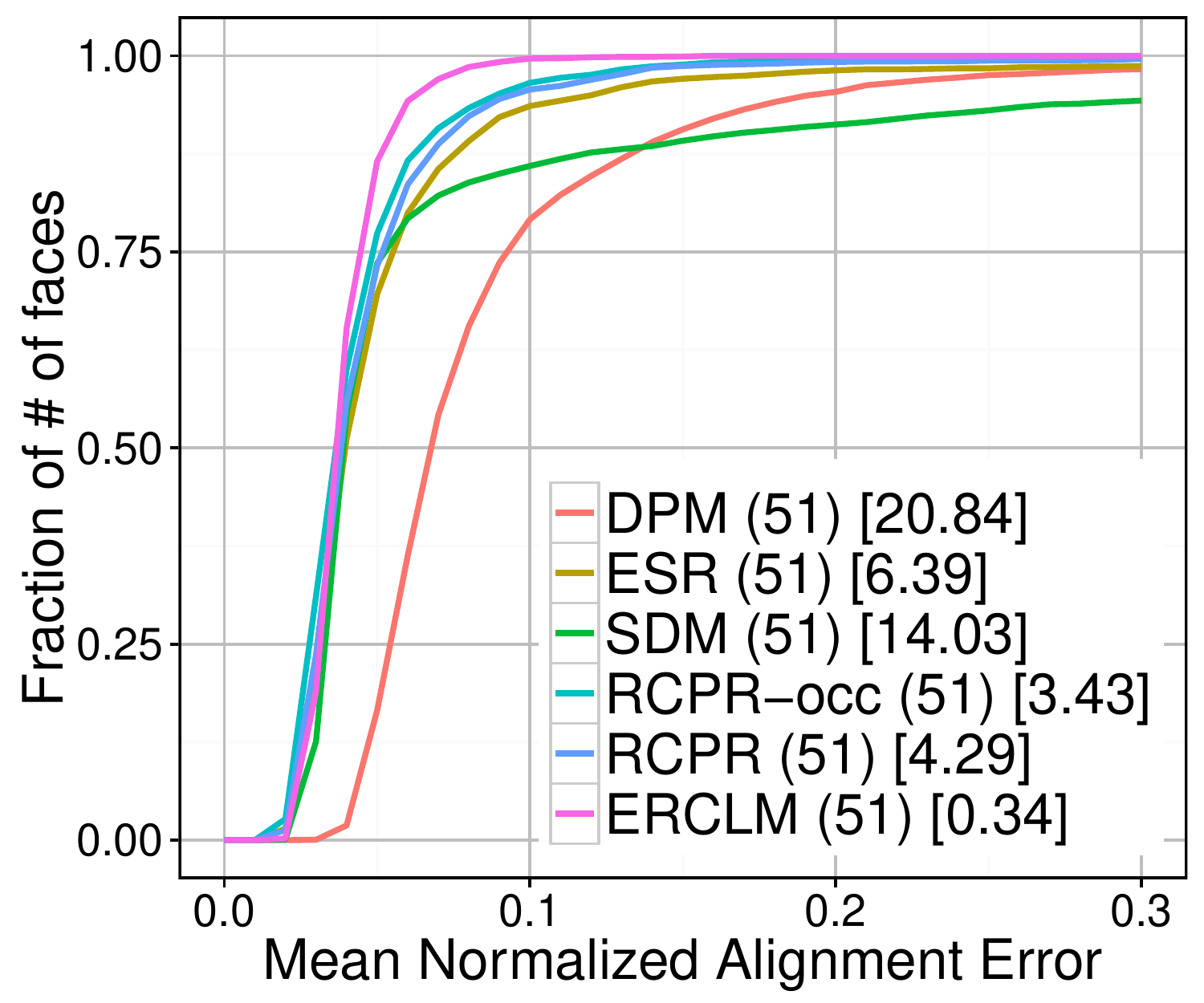}
		\caption{HELEN}
	\end{subfigure}
	\begin{subfigure}[b]{0.24\linewidth}
		\centering
		\includegraphics[width=0.95\linewidth]{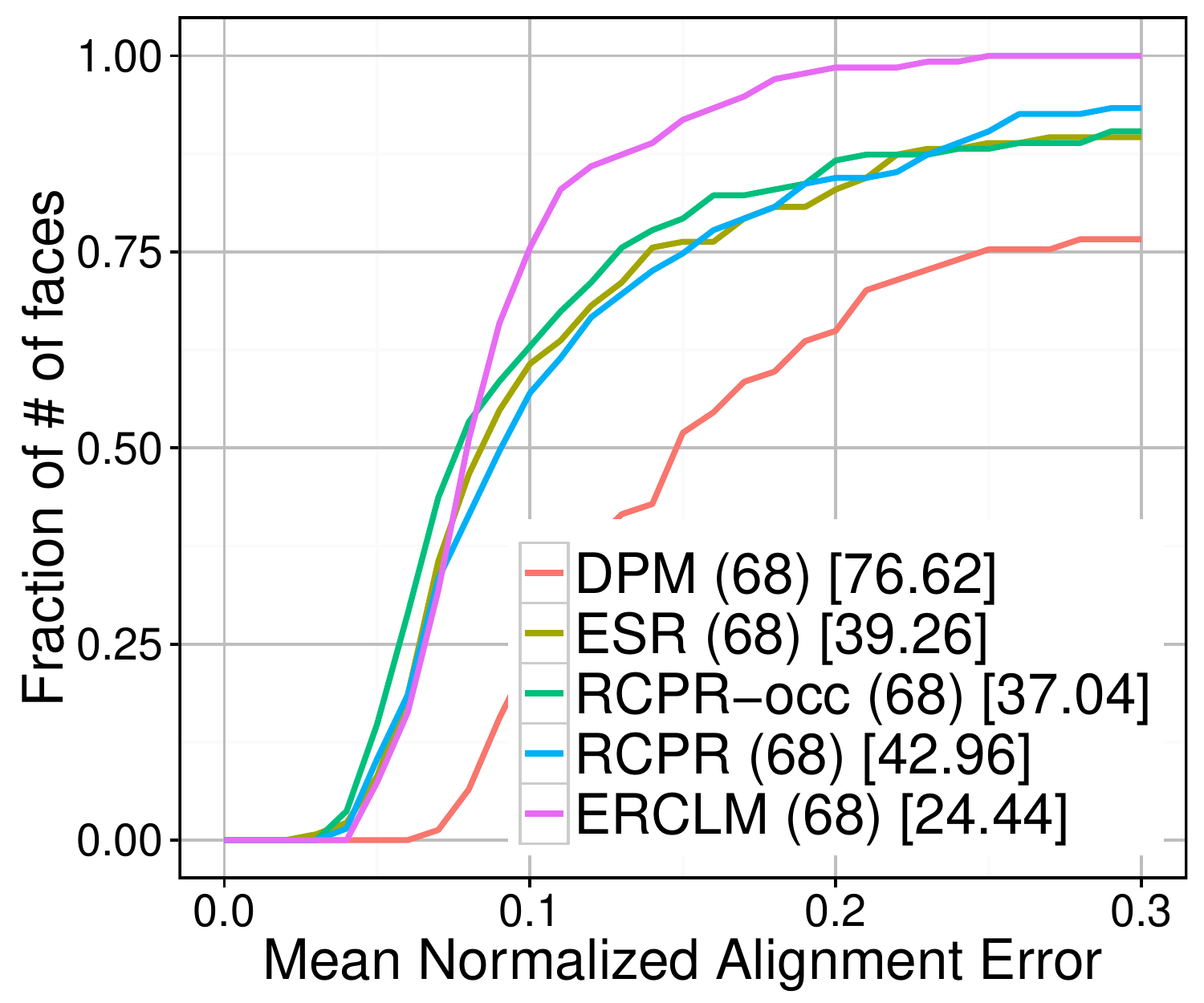}\\
		\includegraphics[width=0.95\linewidth]{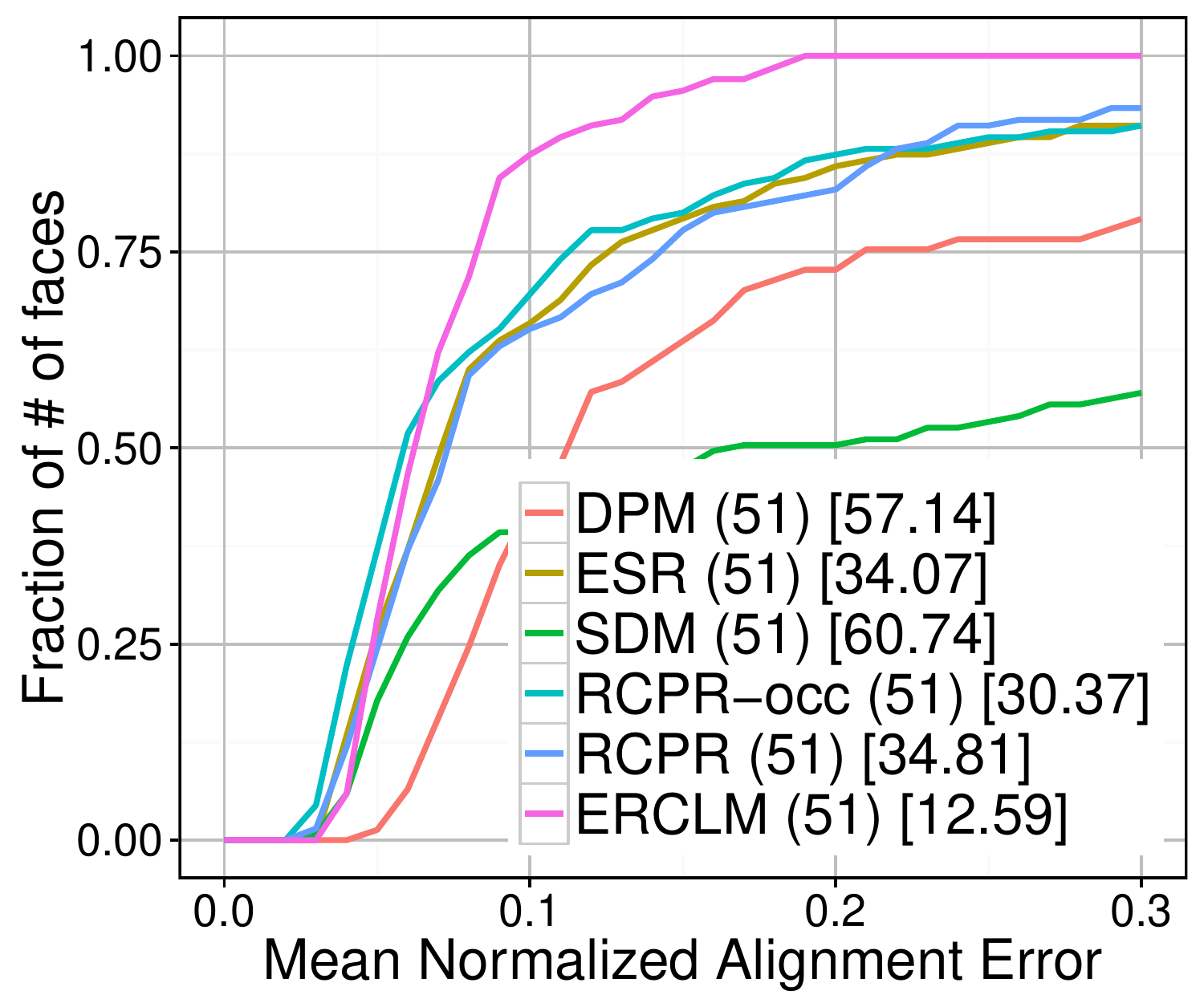}
		\caption{IBUG}
	\end{subfigure}
	\caption{Cumulative error distribution curves for face alignment showing the proportion of images that have the Mean Normalized Alignment Error below a given threshold on the \textbf{AFW}, \textbf{LFPW}, \textbf{HELEN} and \textbf{IBUG} datasets. We compare our proposed method to a baseline tree-structured Deformable Parts Model (DPM) \cite{zhu2012face}, Explicit Shape Regression (ESR) \cite{cao2012face}, Robust Pose Regression (RCPR) \cite{burgos2013robust} and Supervised Descent Method (SDM) \cite{xiong2013supervised}. We show face alignment results both including (68) and excluding (51) the points on the jawline. The legend reports the failure rate (in \%) at a threshold of 0.1. Our method, ERCLM, shows good alignment performance, especially in the presence of severe occlusions and demonstrates robust generalization across datasets.}
	\label{fig:all_plot}
\end{figure*}

\noindent \textbf{Quantitative Results:} We first report results on the AFW, HELEN, LFPW and IBUG datasets. For each of these datasets we retrain the baseline regression based approaches using images from the other three datasets. Due to the cross-dataset nature of our training and evaluation protocol we report results on all (training and testing) the images in each dataset. Finally, due to the relative difficulty of aligning the jawline, we report results both including (68) and excluding (51) the facial landmarks on the jawline.

\begin{table*}[!ht]
\captionsetup{font=footnotesize}
\centering
\scalebox{0.65}{
\begin{tabular}{ccccccccccccccccccccc}
& & & & \multicolumn{2}{c}{\large \textbf{AFW}} & & \multicolumn{5}{c}{\large \textbf{LFPW}} & & \multicolumn{5}{c}{\large \textbf{HELEN}} & & \multicolumn{2}{c}{\large \textbf{IBUG}} \\
& & & & & & & & & & & & & & & & & & & & \\
& & & & \multicolumn{2}{c}{\large \textbf{all}} & & \multicolumn{2}{c}{\large \textbf{test}} & & \multicolumn{2}{c}{\large \textbf{all}} & & \multicolumn{2}{c}{\large \textbf{test}} & & \multicolumn{2}{c}{\large \textbf{all}} & & \multicolumn{2}{c}{\large \textbf{all}}\\
& & & & & & & & & & & & & & & & & & & & \\
\# of \\ landmarks & & {\Large \textbf{Method}} & & MNLE(\%) & FR (\%) & & MNLE(\%) & FR (\%) & & MNLE(\%) & FR (\%) & & MNLE(\%) & FR (\%) & & MNLE(\%) & FR (\%) & & MNLE(\%) & FR (\%)\\
\cline{1-1} \cline{3-3} \cline{5-6} \cline{8-9} \cline{11-12} \cline{14-15} \cline{17-18} \cline{20-21}
& & & & & & & & & & & & & & & & & & & &\\
& & \large DPM & & \Large 10.2 & \Large 34.2 & & \Large 8.3 & \Large 23.2 & & \Large 8.6 & \Large 24.8 & & \Large 8.8 & \Large 24.7 & & \Large 11.3 & \Large 29.8 & & \Large 19.7 & \Large 70.1 \\
\multirow{4}{*}{\Large 68} & & & & & & & & & & & & & & & & & & & &\\
& & \large ESR & & \Large 7.2 & \Large 12.8 & & \Large 4.9 & \Large 4.0 & & \Large 5.0 & \Large 4.6 & & \Large 5.6 & \Large 7.0 & & \Large 6.2 & \Large 8.5 & & \Large 12.8 & \Large 39.3 \\
& & & & & & & & & & & & & & & & & & & &\\
& & \large RCPR-occ & & \large 7.1 & \large 11.9 & & \large \textbf{4.1} & \large 1.8 & & \large \textbf{4.5} & \large 3.5 & & \large 5.0 & \large 4.2 & & \large 5.3 & \large 5.1 & & \large 12.1 & \large 37.0 \\
& & & & & & & & & & & & & & & & & & & &\\
& & \large RCPR & & \large 7.4 & \large 13.1 & & \large 4.8 & \large 3.6 & & \large 4.8 & \large 4.3 & & \large 5.1 & \large 4.8 & & \large 5.5 & \large 5.8 & & \large 12.6 & \large 42.9 \\
& & & & & & & & & & & & & & & & & & & &\\
& & \large ERCLM & & \large \textbf{5.7} & \large \textbf{4.7} & & \large 4.4 & \large \textbf{0.0} & & \large 4.8 & \large \textbf{1.7} & & \large \textbf{4.7} & \large \textbf{1.5} & & \large \textbf{4.9} & \large \textbf{1.6} & & \large \textbf{8.9} & \large \textbf{26.7}\\
\cline{1-1} \cline{3-3} \cline{5-6} \cline{8-9} \cline{11-12} \cline{14-15} \cline{17-18} \cline{20-21}
& & & & & & & & & & & & & & & & & & & &\\
& & \large DPM & & \large 8.8 & \large 22.0 & & \large 6.8 & \large 10.9 & & \large 7.2 & \large 12.3 & & \large 6.6 & \large 10.8 & & \large 8.1 & \large 17.4 & & \large 17.6 & \large 53.2 \\
& & & & & & & & & & & & & & & & & & & &\\
& & \large ESR & & \large 6.3 & \large 8.9 & & \large 4.0 & \large 3.1 & & \large 4.1 & \large 3.7 & & \large 4.7 & \large 3.9 & & \large 5.4 & \large 6.4 & & \large 11.7 & \large 34.1 \\
& & & & & & & & & & & & & & & & & & & &\\
\large 51 & & \large SDM & & \large 15.9 & \large 30.0 & & \large 7.9 & \large 14.7 & & \large 9.3 & \large 17.2 & & \large 8.1 & \large 13.0 & & \large 8.1 & \large 14.0 & & \large 30.4 & \large 60.7 \\
& & & & & & & & & & & & & & & & & & & &\\
& & \large RCPR-occ & & \large 6.3 & \large 9.5 & & \large \textbf{3.2} & \large 1.8 & & \large \textbf{3.6} & \large 2.6 & & \large 4.1 & \large 3.3 & & \large 4.4 & \large 3.4 & & \large 11.1 & \large 30.4 \\
& & & & & & & & & & & & & & & & & & & &\\
& & \large RCPR & & \large 6.8 & \large 10.1 & & \large 4.1 & \large 2.7 & & \large 4.0 & \large 3.2 & & \large 4.3 & \large 3.6 & & \large 4.7 & \large 4.3 & & \large 11.7 & \large 34.8 \\
& & & & & & & & & & & & & & & & & & & &\\
& & \large ERCLM & & \large \textbf{4.5} & \large \textbf{2.1} & & \large 3.5 & \large \textbf{0.0} & & \large 3.9 & \large \textbf{0.6} & & \large \textbf{3.7} & \large \textbf{0.9} & & \large \textbf{3.9} & \large \textbf{0.6} & & \large \textbf{7.1} & \large \textbf{14.8}\\
\cline{1-1} \cline{3-3} \cline{5-6} \cline{8-9} \cline{11-12} \cline{14-15} \cline{17-18} \cline{20-21}
\end{tabular}}
\caption{Face alignment results on the \textbf{AFW}, \textbf{LFPW}, \textbf{HELEN} and \textbf{IBUG} datasets evaluated over both 68 (includes jawline)  and 51 (excludes jawline) landmarks. We report both the Mean Normalized Landmark Error (MNLE) and the alignment Failure Rate (FR). Due to the robustness of our algorithm (ERCLM) to occlusions the face alignment failure rate is significantly reduced on all the datasets.}
\label{table:all}
\end{table*}

Table \ref{table:all} presents the aggregate results on the AFW, LFPW, HELEN and IBUG datasets, both the test subset as well as the full dataset for the LFPW and HELEN datasets. Figure \ref{fig:all_plot} shows the cumulative face alignment Failure Rate (FR) as a function of the Mean Normalized Alignment Error (MNAE). Unsurprisingly, both our method and the baselines achieve better performance when excluding the jawline from the evaluation. ERCLM achieves significantly lower face alignment error and face alignment failure rate especially on difficult datasets like AFW and IBUG. DPM, despite using many local detectors and explicit modeling of the continuous variation in facial pose performs poorly on the difficult datasets due to the lack of explicit occlusion modeling.

Regression based approaches perform excellently on datasets with near frontal pose and free of occlusion. However, regression based face alignment approaches are extremely sensitive to initialization \cite{yan2013learn} and often perform very poorly if there is a mismatch between the initializations used at train and test time. This is exemplified by the poor performance of pre-trained SDM on all the datasets since its training face detector is different (we were unable to use the OpenCV face detector used by the authors since it failed on most of the images in these datasets) from the one used for evaluation.  CLM based approaches, the proposed method as well as DPM, on the other hand is very robust to the initialization from the face detector. Surprisingly, RCPR trained with virtually occluded faces and labels performs worse in comparison, suggesting possible over-fitting.

\begin{figure*}
\captionsetup{font=small}
	\centering
	\begin{subfigure}[b]{0.39\linewidth}
		\centering
		\includegraphics[width=0.49\linewidth]{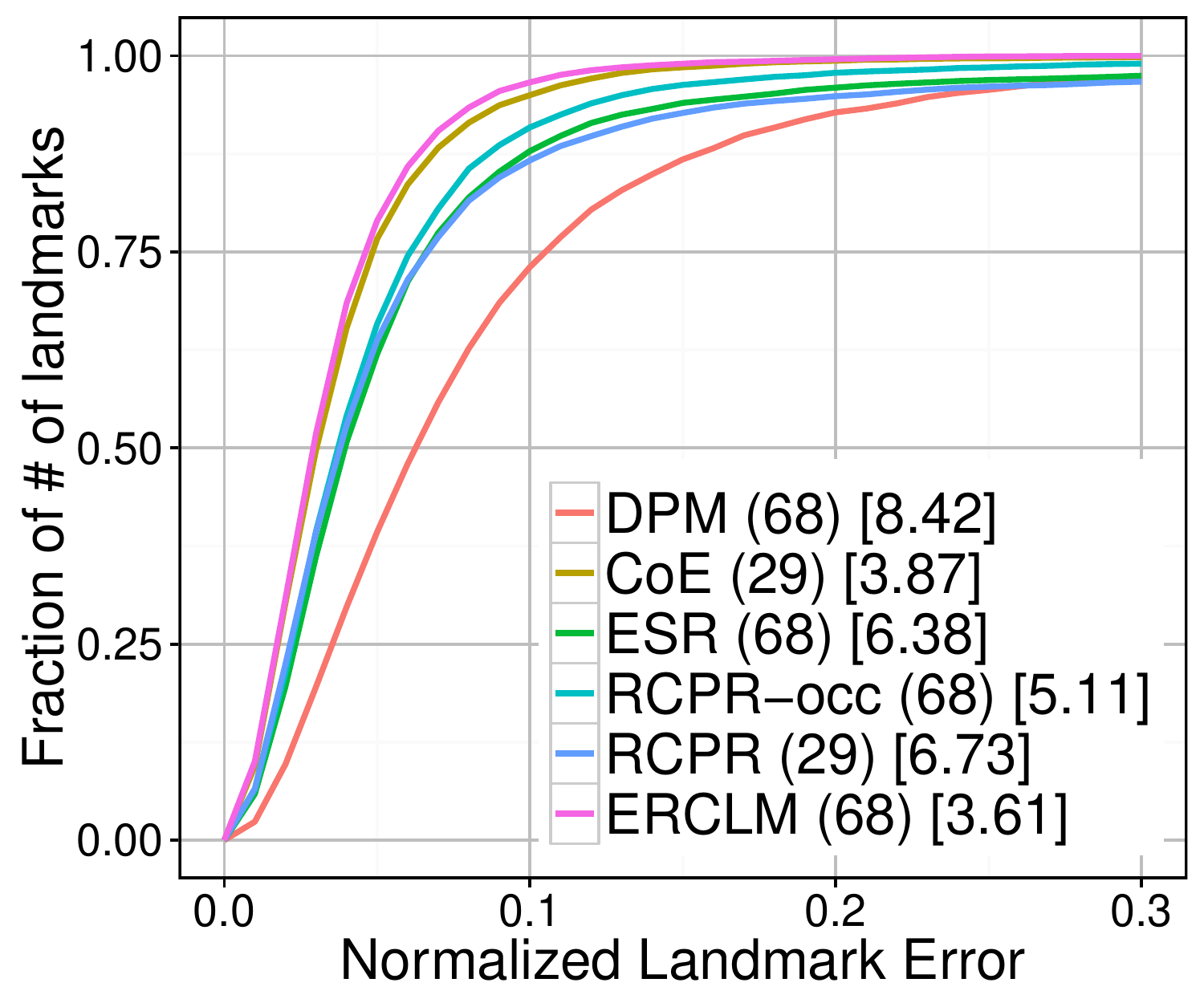}
		\includegraphics[width=0.49\linewidth]{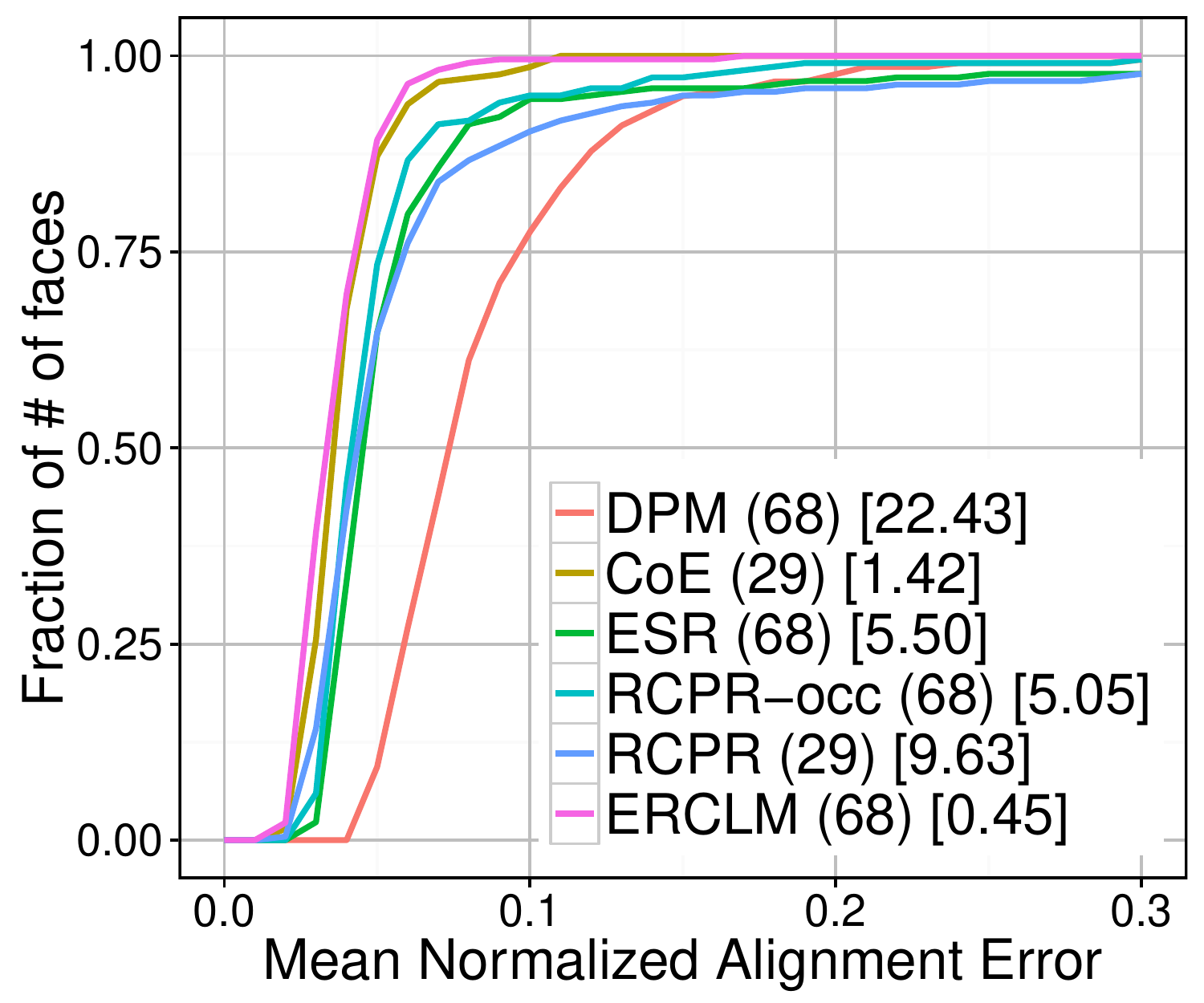}
		\caption{LFPW}
	\end{subfigure}
	\begin{subfigure}[b]{0.59\linewidth}
		\centering
		\includegraphics[width=0.325\linewidth]{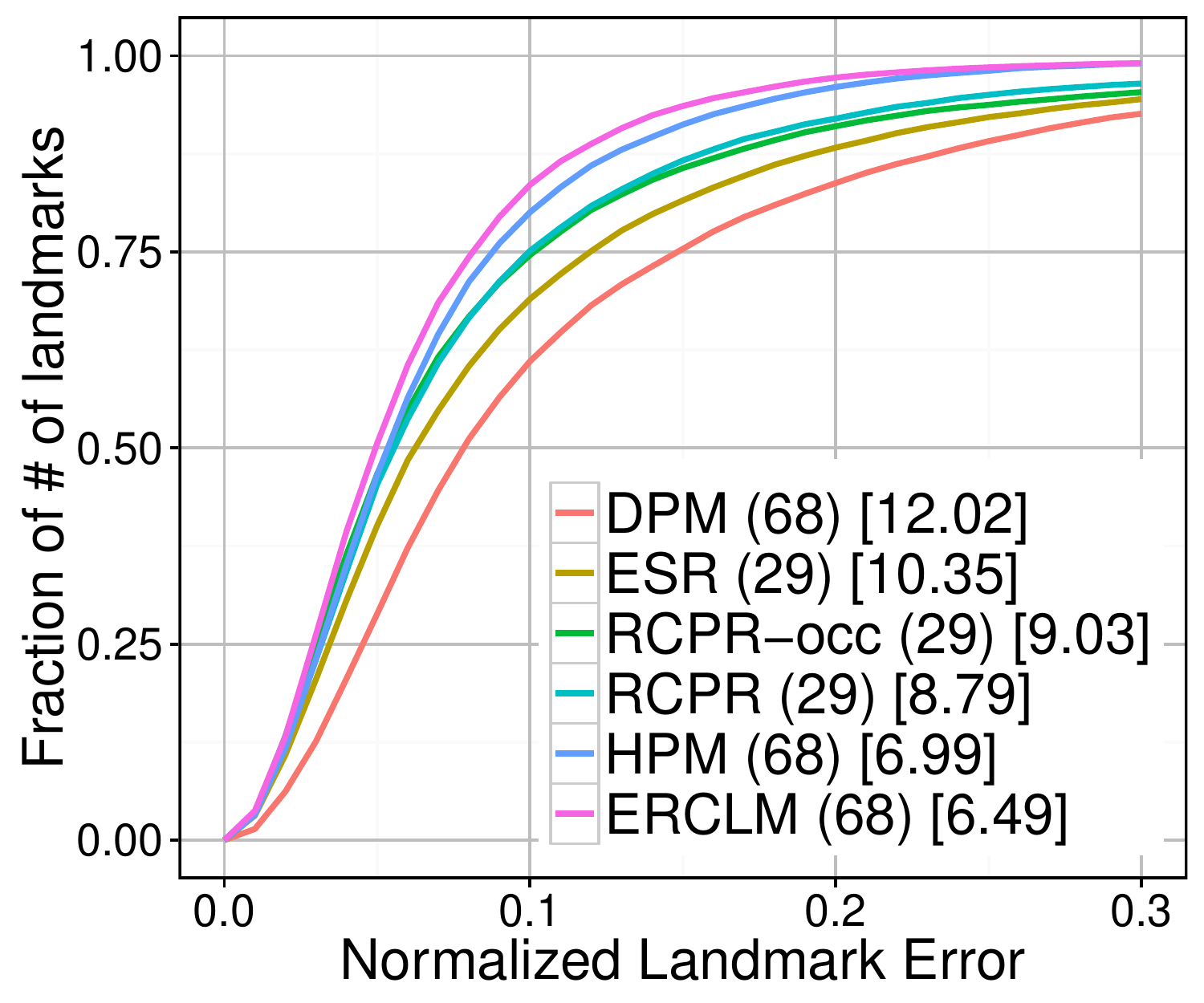}
		\includegraphics[width=0.325\linewidth]{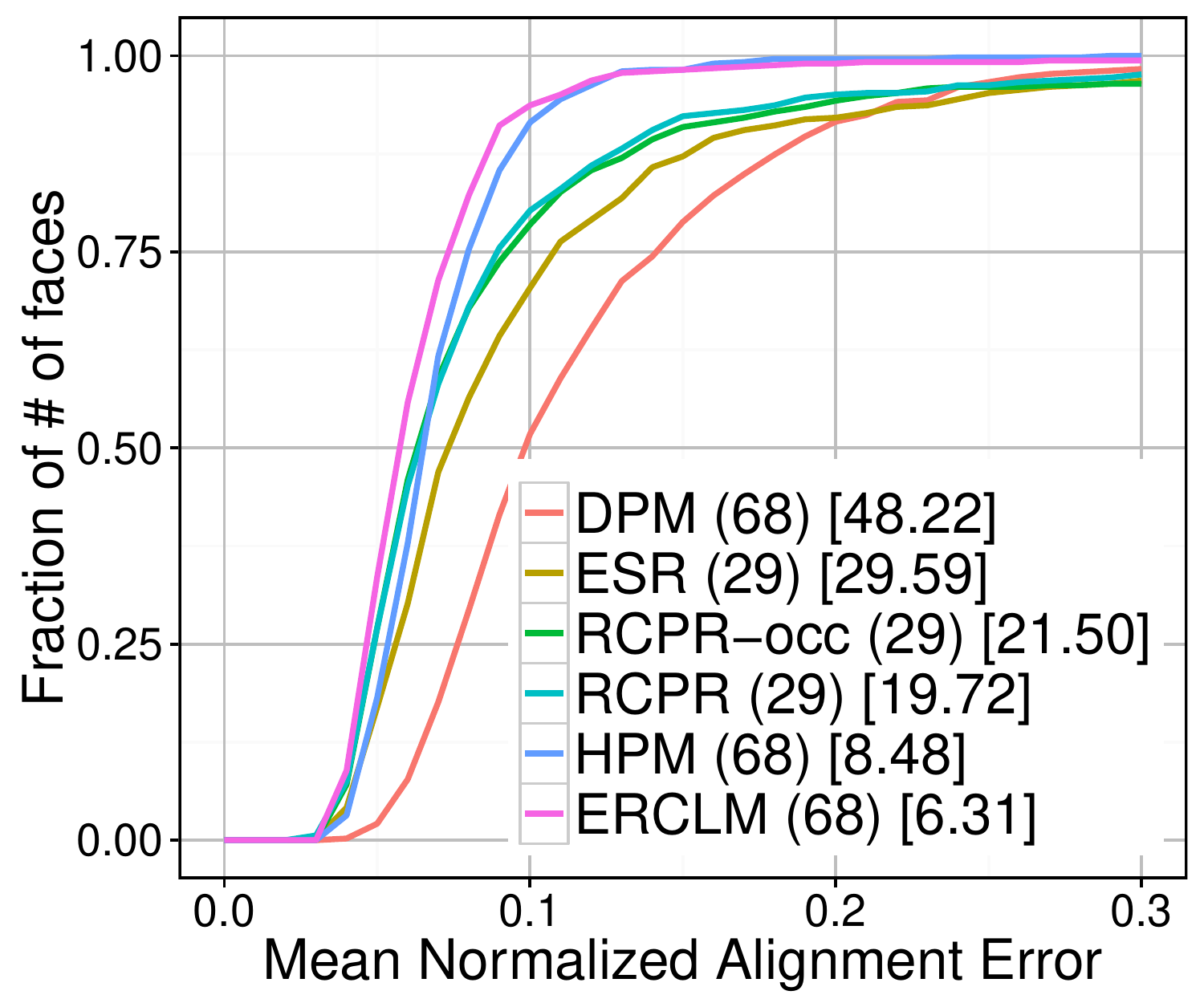}
		\includegraphics[width=0.325\linewidth]{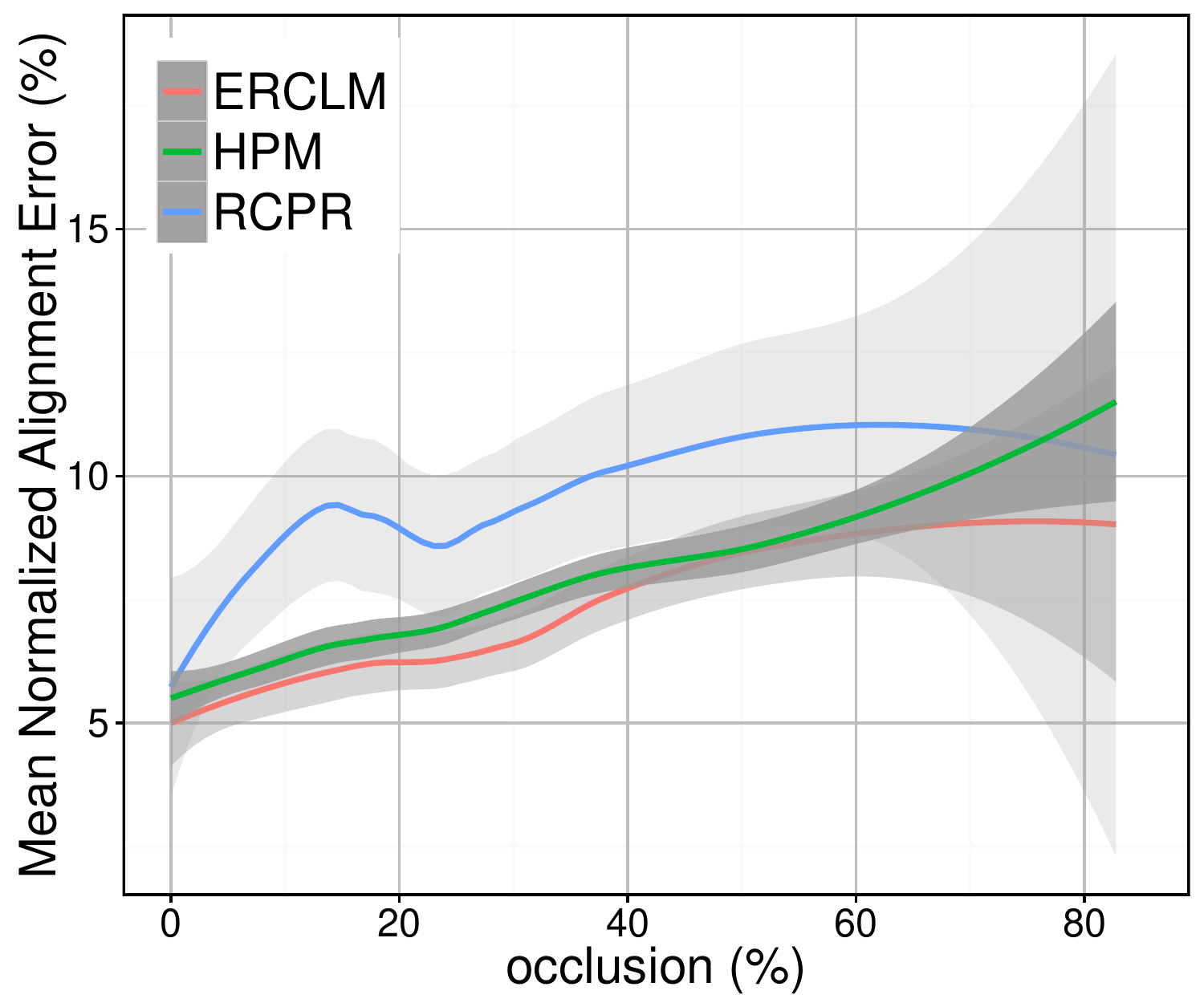}
		\caption{COFW}
	\end{subfigure}	
	\caption{Face alignment results on the \textbf{LFPW} and \textbf{COFW} dataset evaluated over 29 landmarks. We report the mean landmark localization error and the face alignment failure rate. On the \textbf{COFW} dataset we also compare the face alignment performance of RCPR, HPM and ERCLM as a function of the amount of facial occlusion. Due to the robustness of ERCLM to facial pose and facial occlusions the face alignment failure rate is significantly reduced.}
	\label{fig:landmarks_29}
\end{figure*}

We also evaluate ERCLM for predicting 29 landmarks on the LFPW test set and the COFW dataset by mapping our 68 point shape to the 29 point configuration using the linear regressor learned in \cite{ghiasi2014occlusion}. For the LFPW test set we also report the original results of the Consensus of Exemplars (CoE) \cite{belhumeur2011localizing} approach. Figure \ref{fig:landmarks_29} compares the cumulative landmark localization failure rate as a function of normalized landmark error and the cumulative face alignment failure rate as a function of MNAE. Additionally, for the COFW dataset we also report the MNAE as a function of the amount of facial occlusion. Our method consistently achieves lower and more stable localization error across all degrees of occlusions in comparison to RCPR and Hierarchical Parts Model (HPM) \cite{ghiasi2014occlusion}. On the COFW dataset with significant facial occlusion our method achieves a face alignment FR of 6.31\% and average landmark localization error of 6.49\% compared to 8.48\% FR and mean error of 6.99\% achieved by HPM. Our explicit (combinatorial) search over landmark occlusion labels during inference is more effective at handling occlusions compared to RCPR and HPM which rely on learning occlusion patterns at the training stage only. On the LFPW dataset, where face alignment performance is saturating and reaching or exceeding human performance~\cite{burgos2013robust}, our results are comparable to the CoE and HPM approach.

Finally, we note that our results have been achieved by training on the Multi-PIE dataset which neither exhibits facial occlusions nor as much variation in facial shape (especially no variation in facial pitch) while the baselines (except DPM) has been trained on images similar to the test set and also requires occlusion labels (only RCPR) at training time. This demonstrates the generalization capability of our face alignment framework.

\begin{figure*}
	\captionsetup{font=footnotesize}
	\includegraphics[scale=0.075]{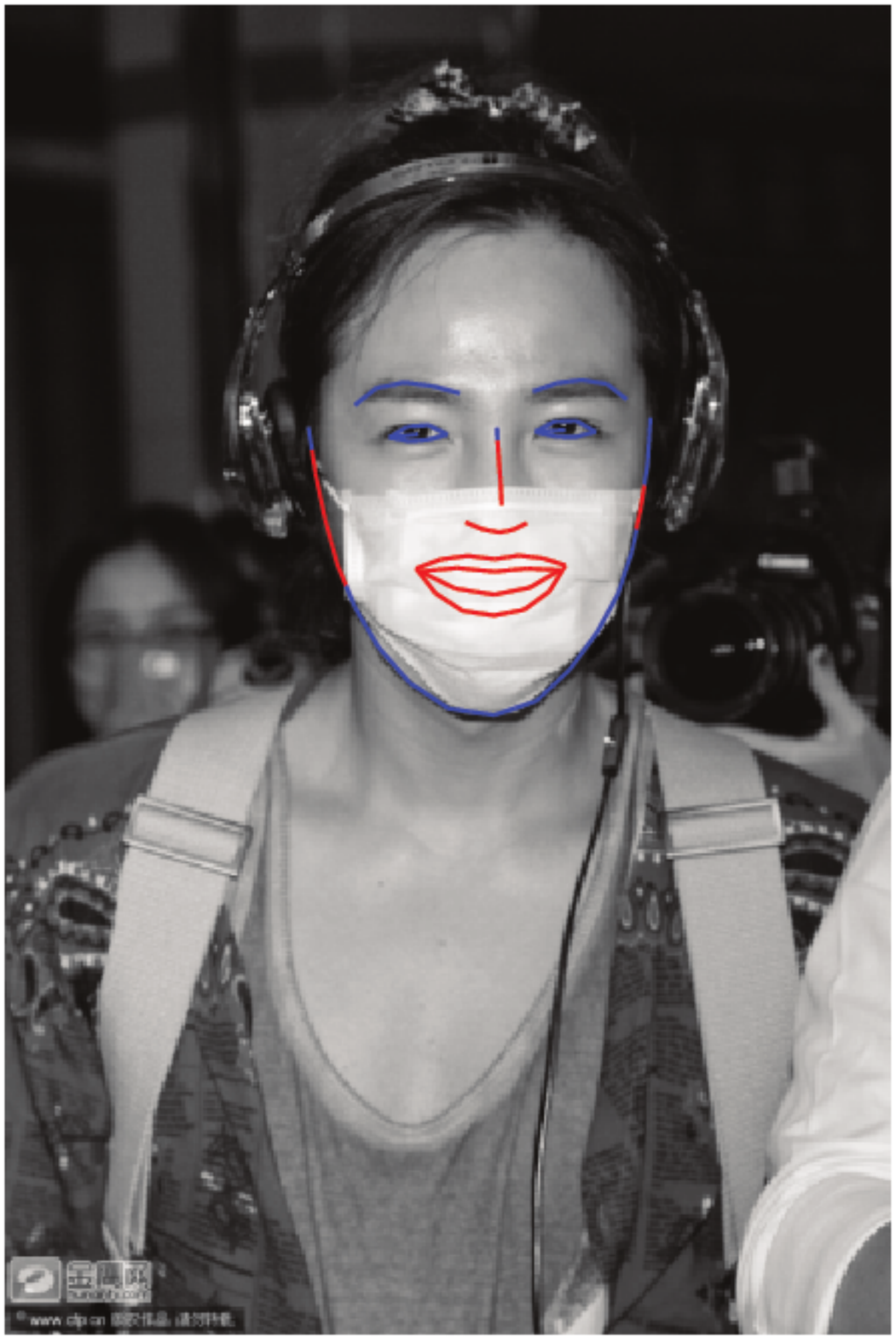}
	\includegraphics[scale=0.081]{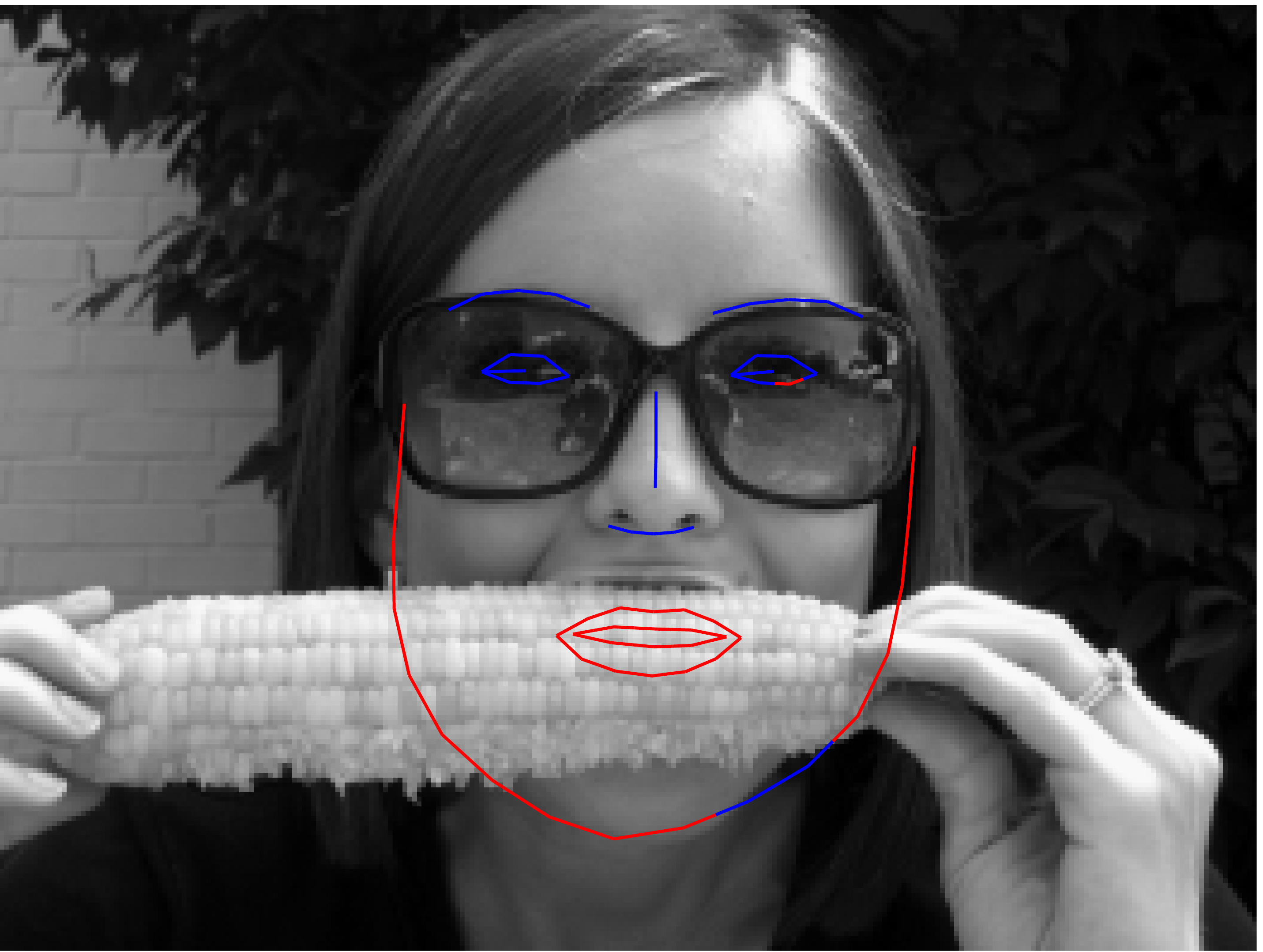}
	\includegraphics[scale=0.08]{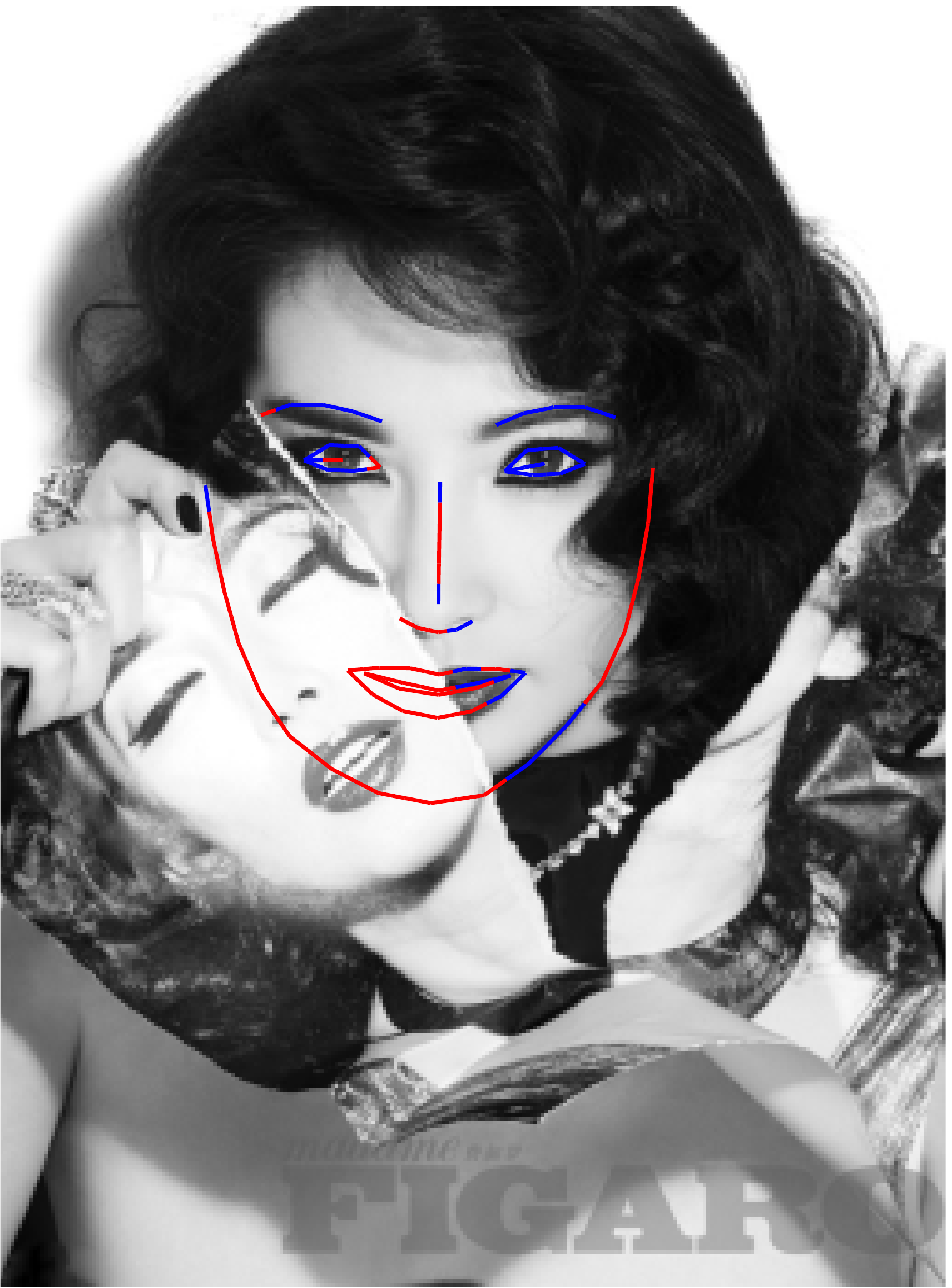}
	\includegraphics[scale=0.081]{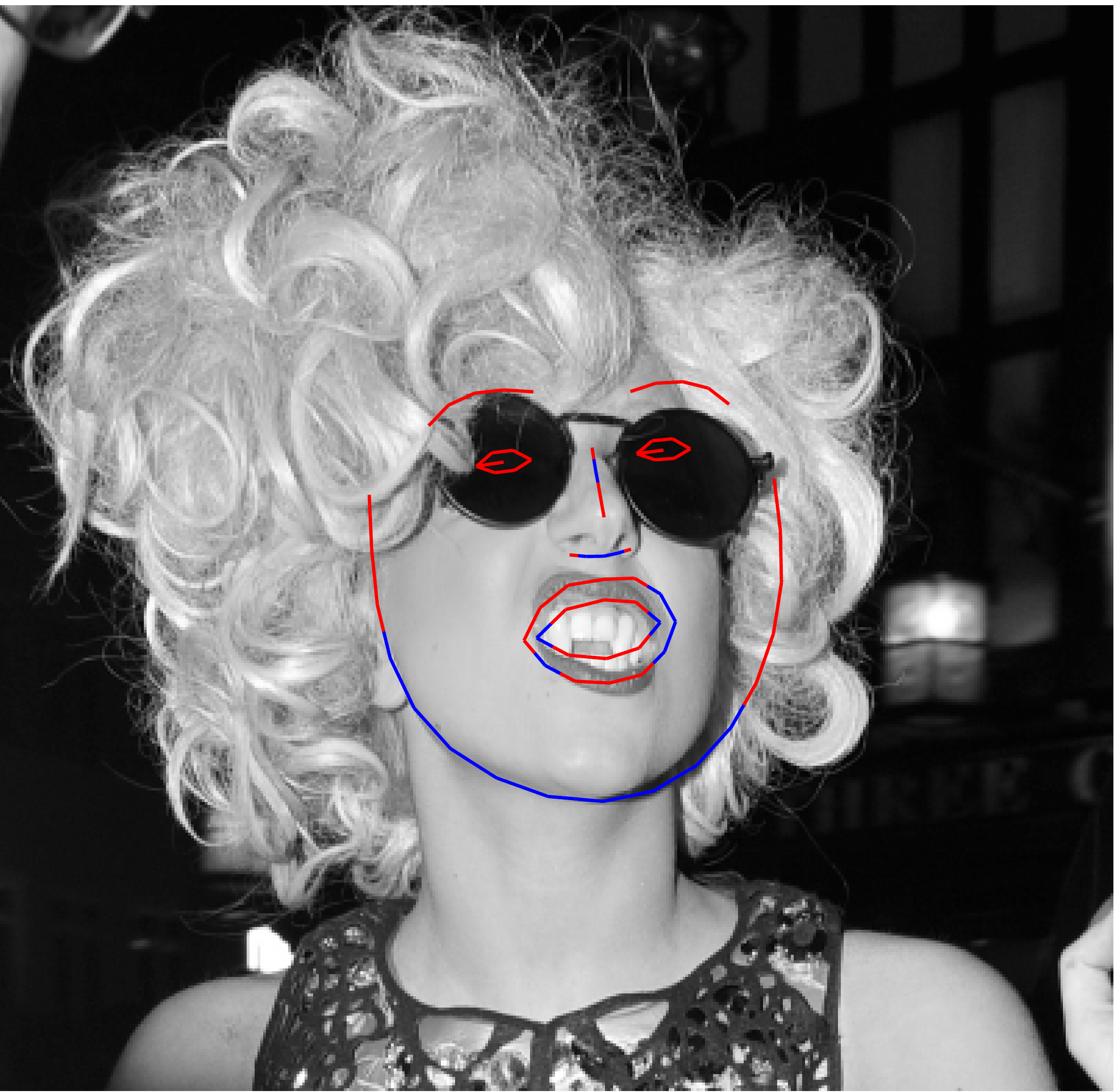}
	\includegraphics[scale=0.08]{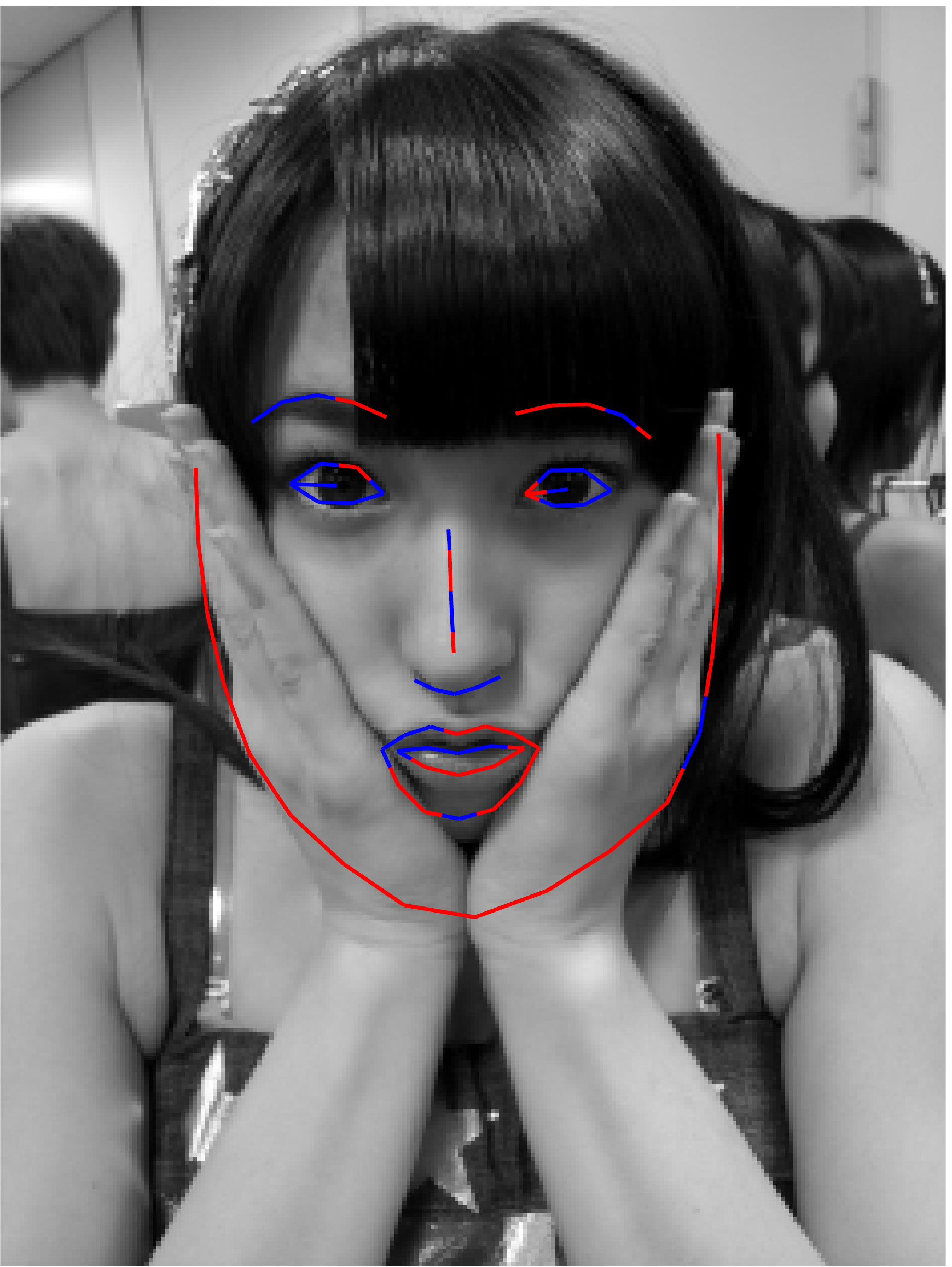}
	\includegraphics[scale=0.085]{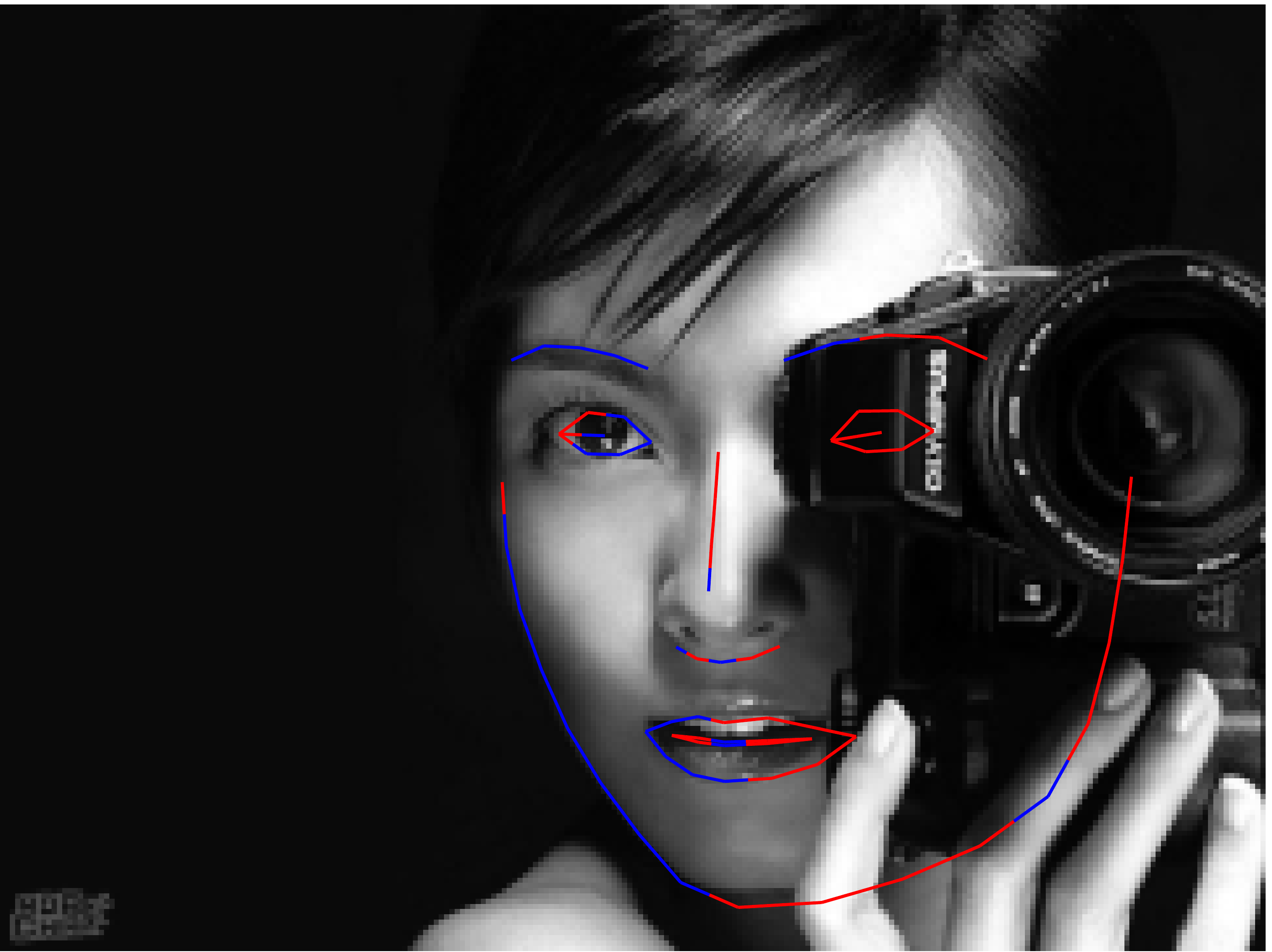}
	\includegraphics[scale=0.085]{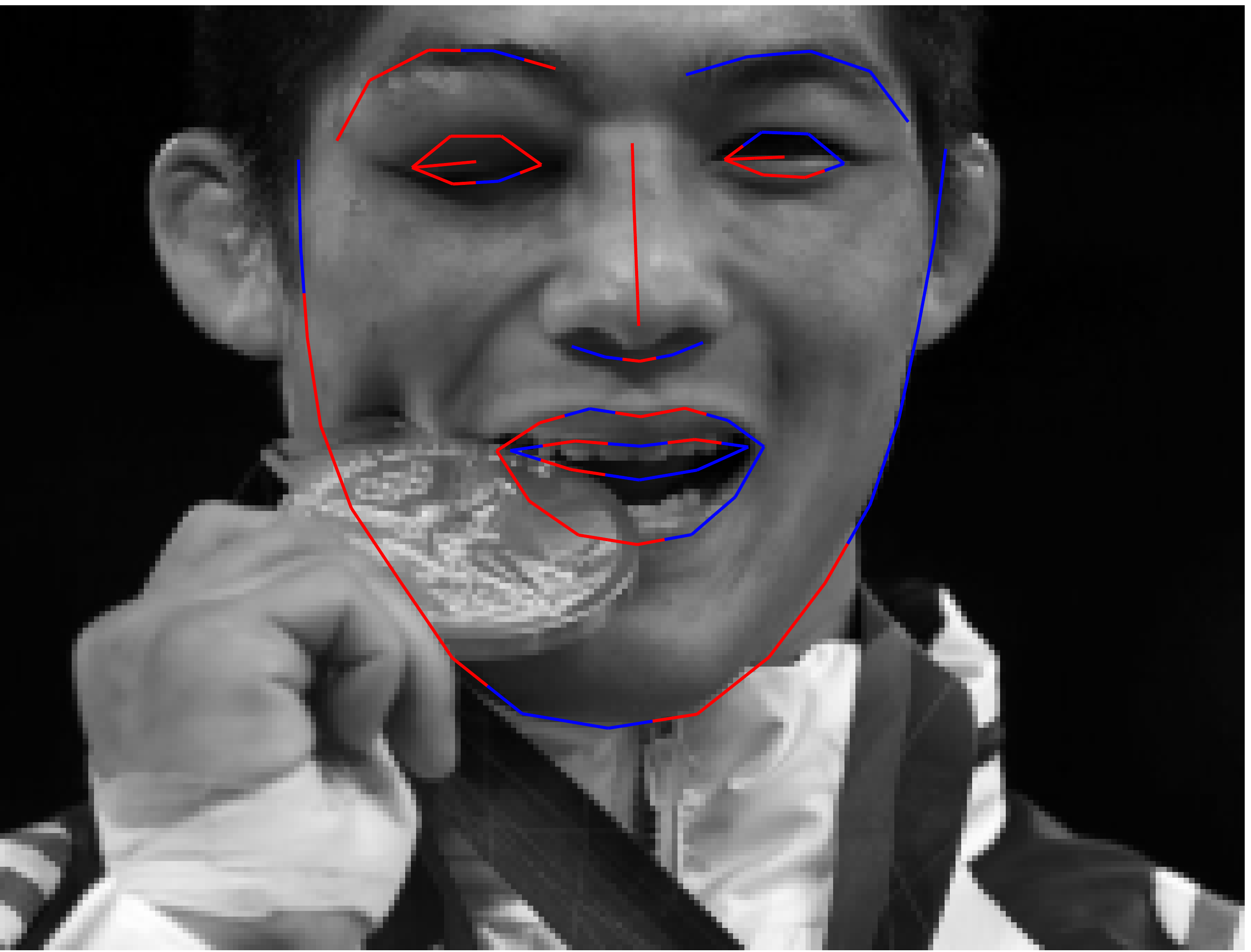}
	\includegraphics[scale=0.082]{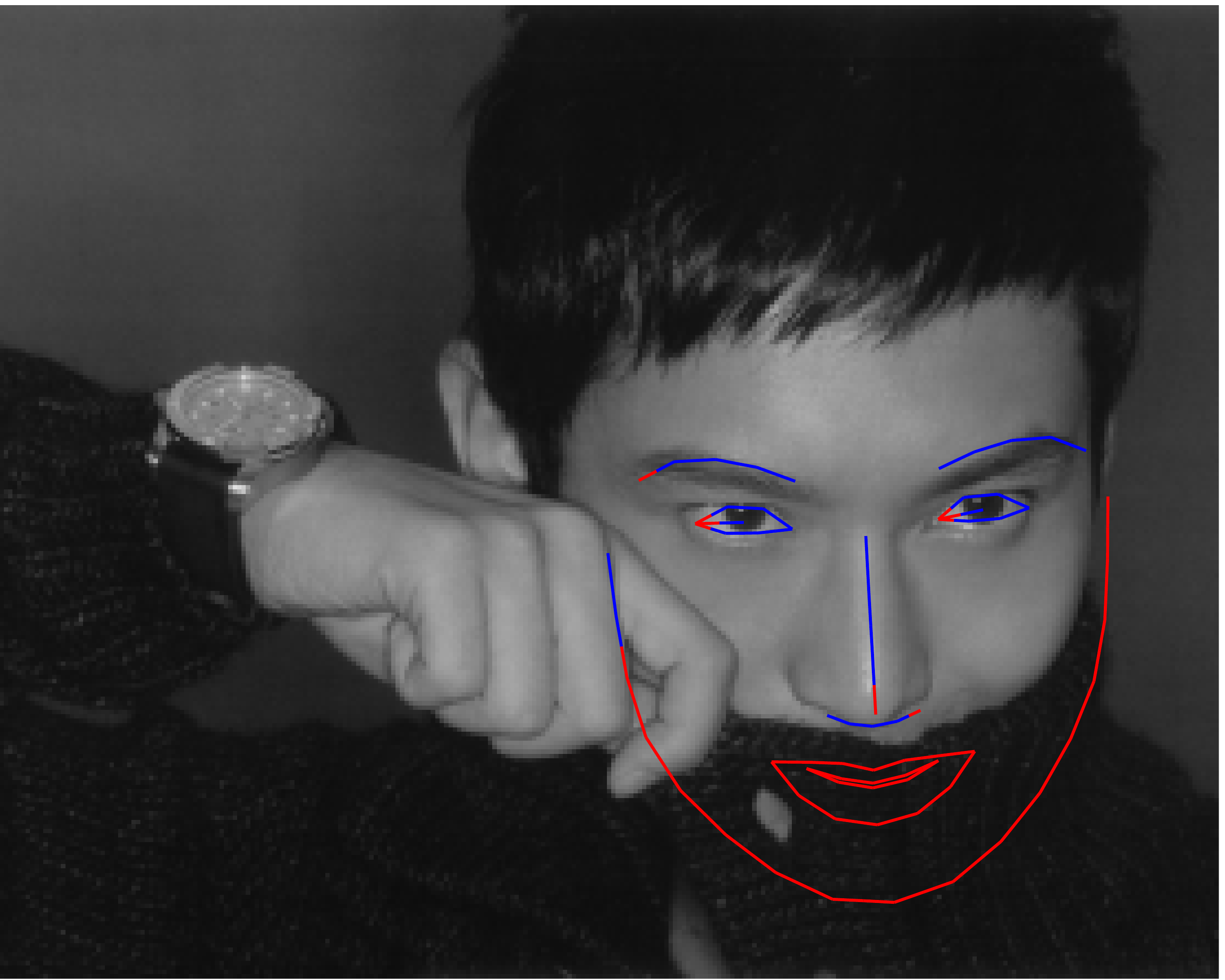}	
	\includegraphics[scale=0.067]{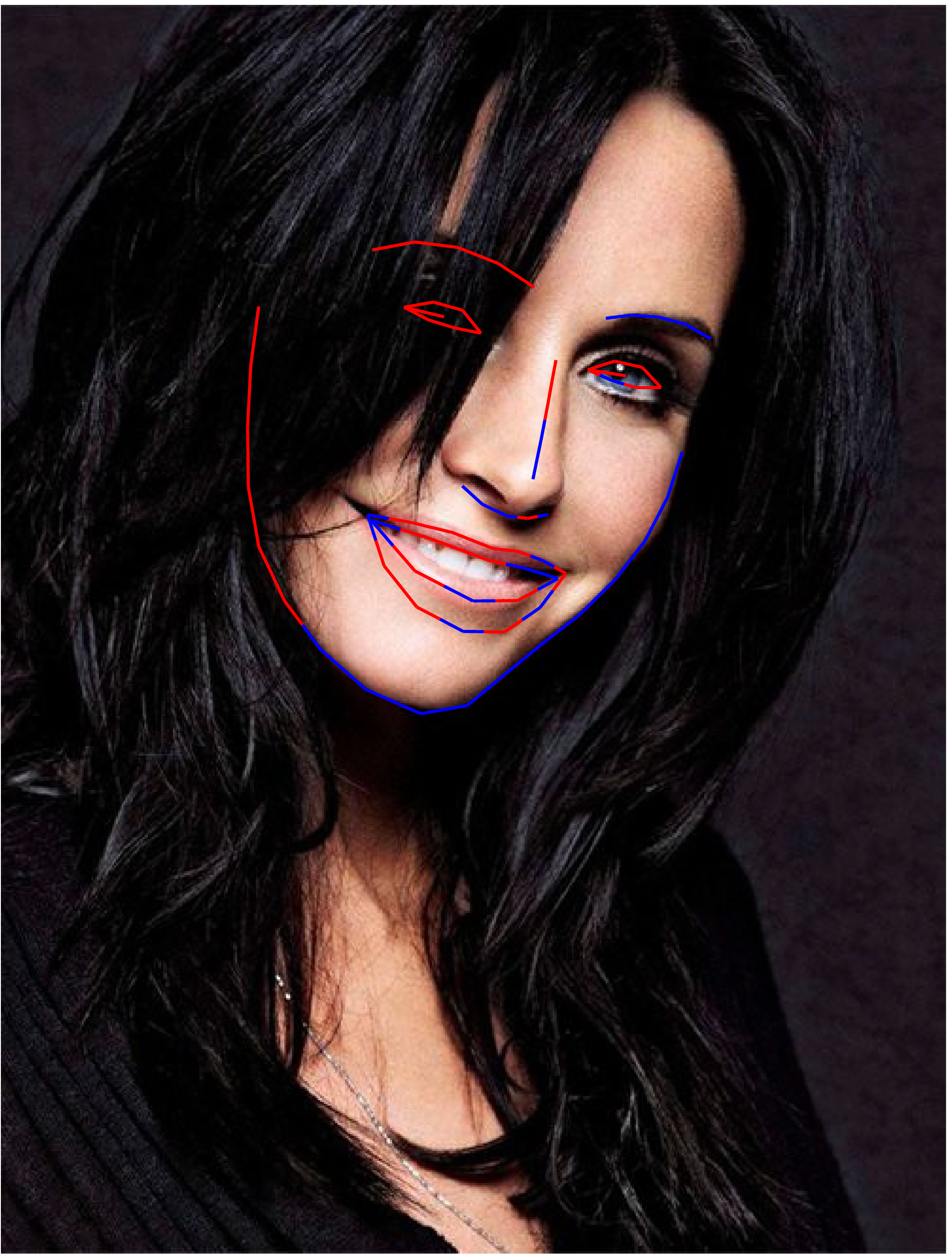}
	
	\includegraphics[scale=0.082]{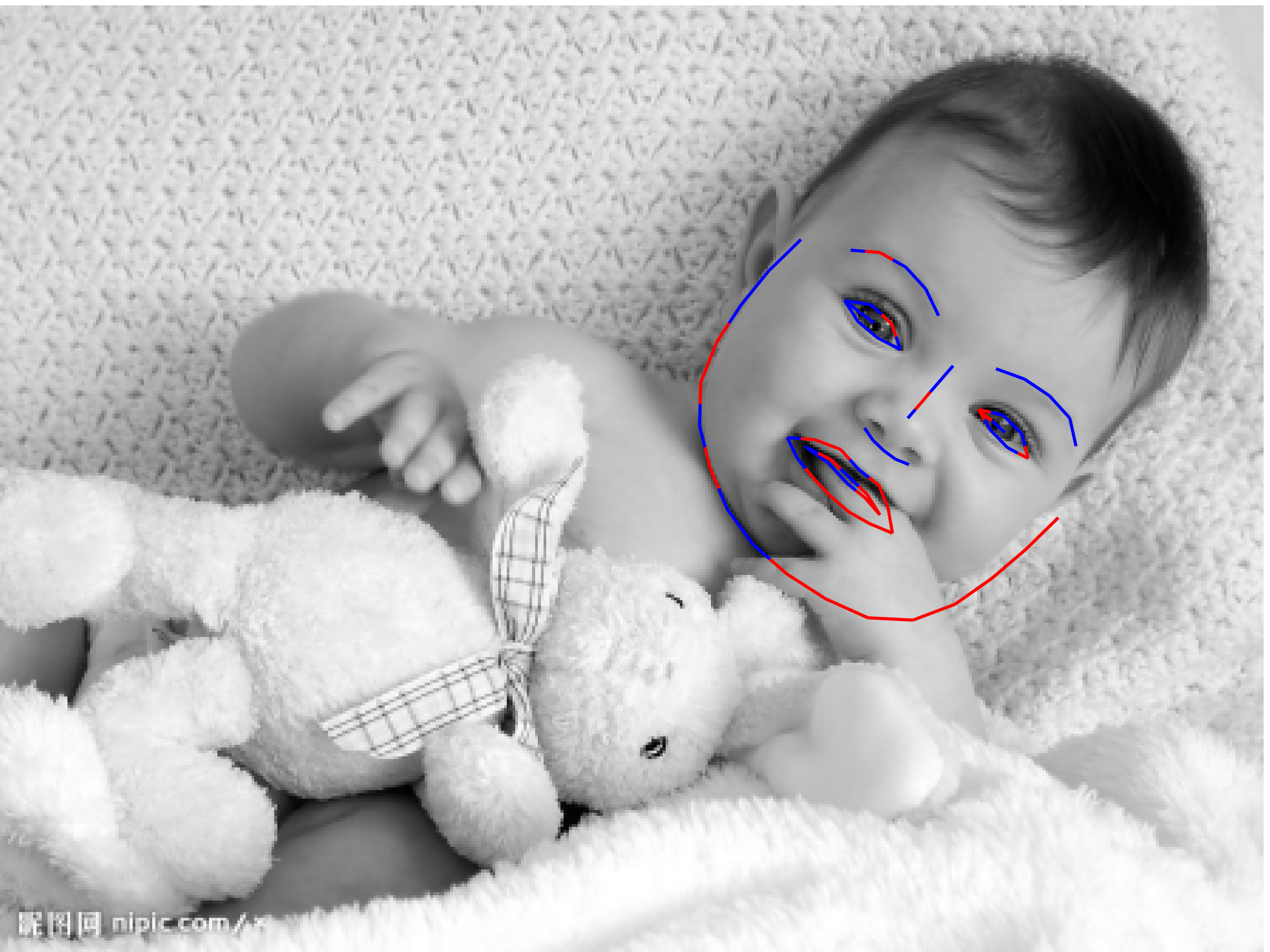}
	\includegraphics[scale=0.092]{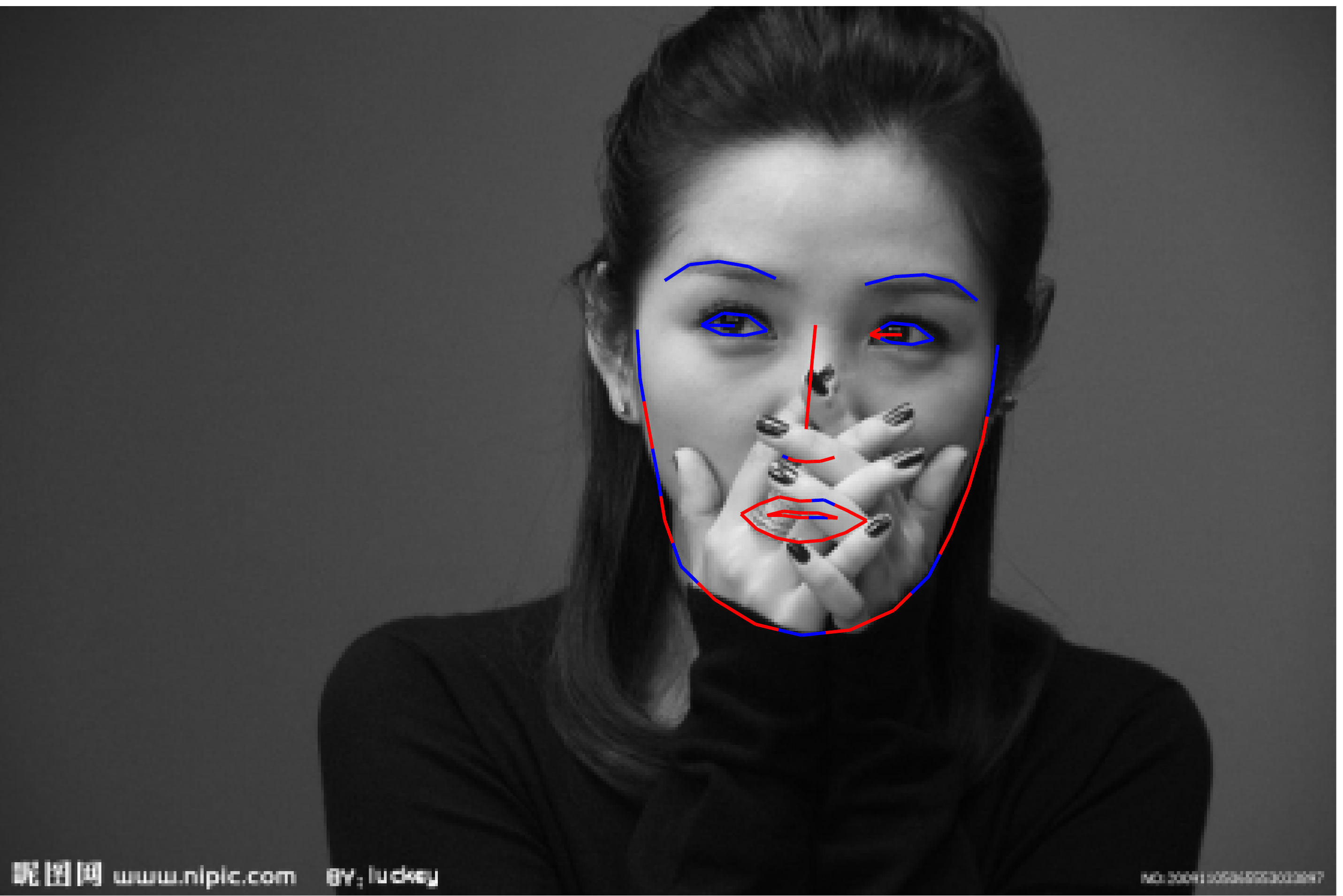}
	\includegraphics[scale=0.092]{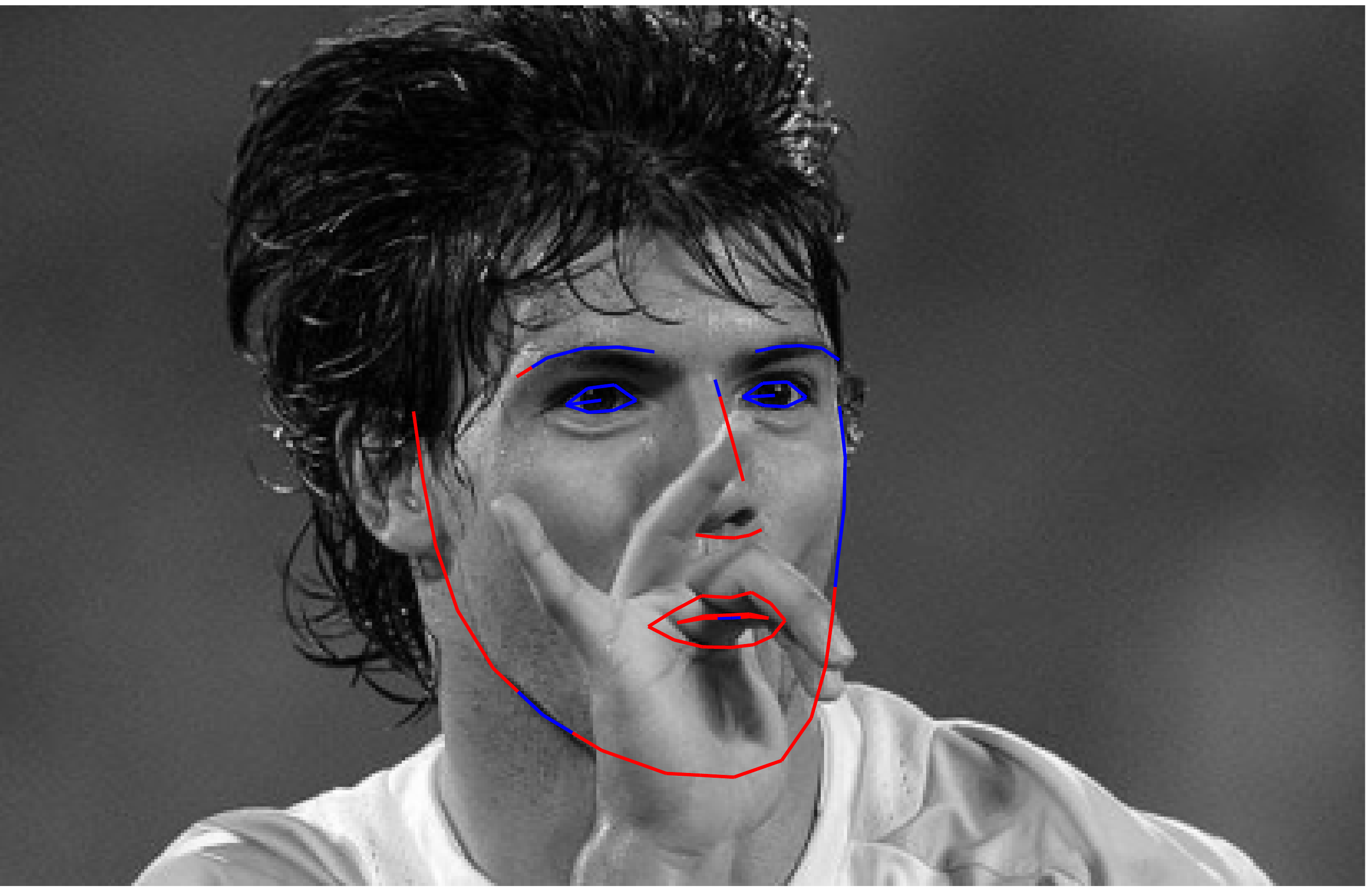}
	\includegraphics[scale=0.09]{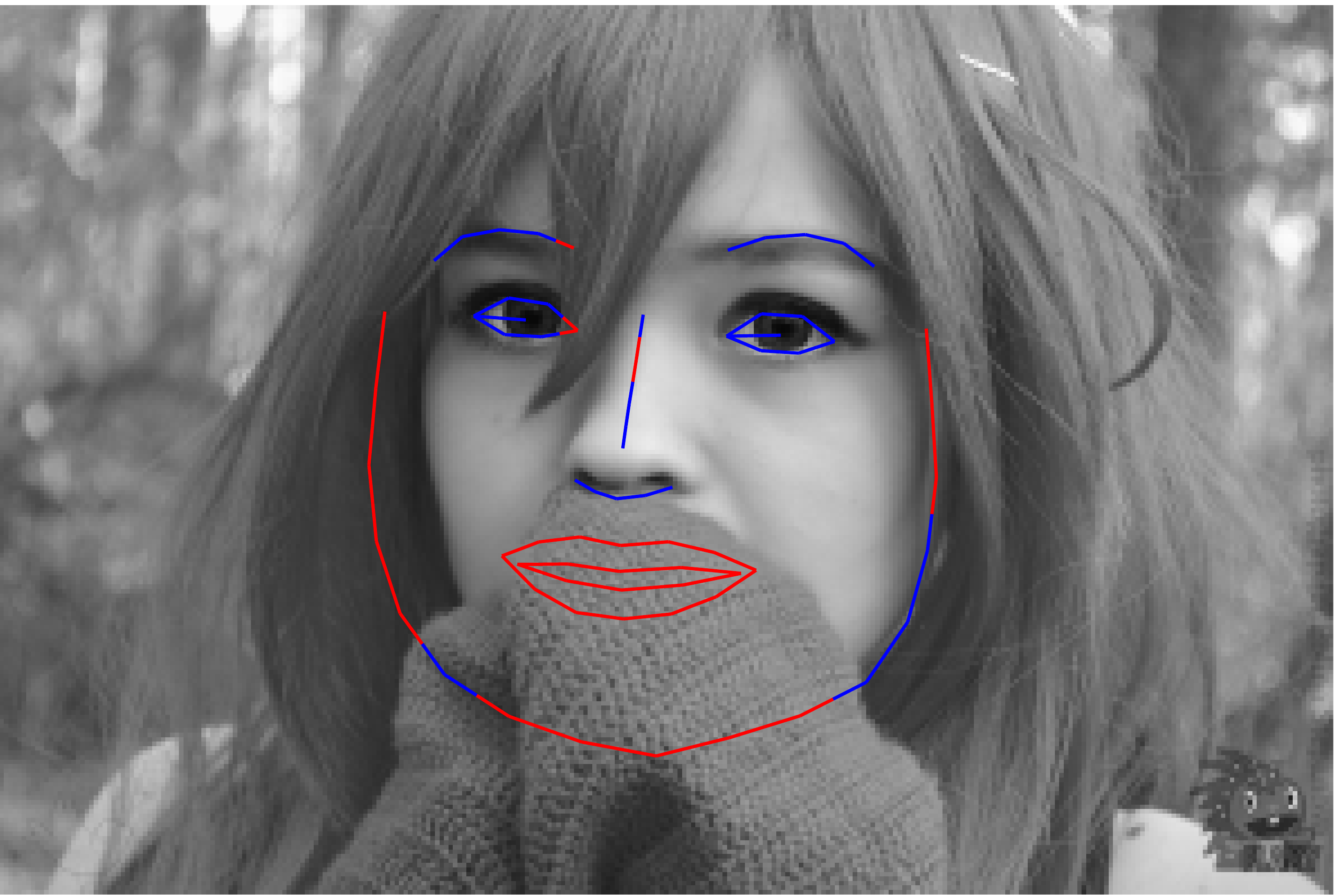}
	\includegraphics[scale=0.08]{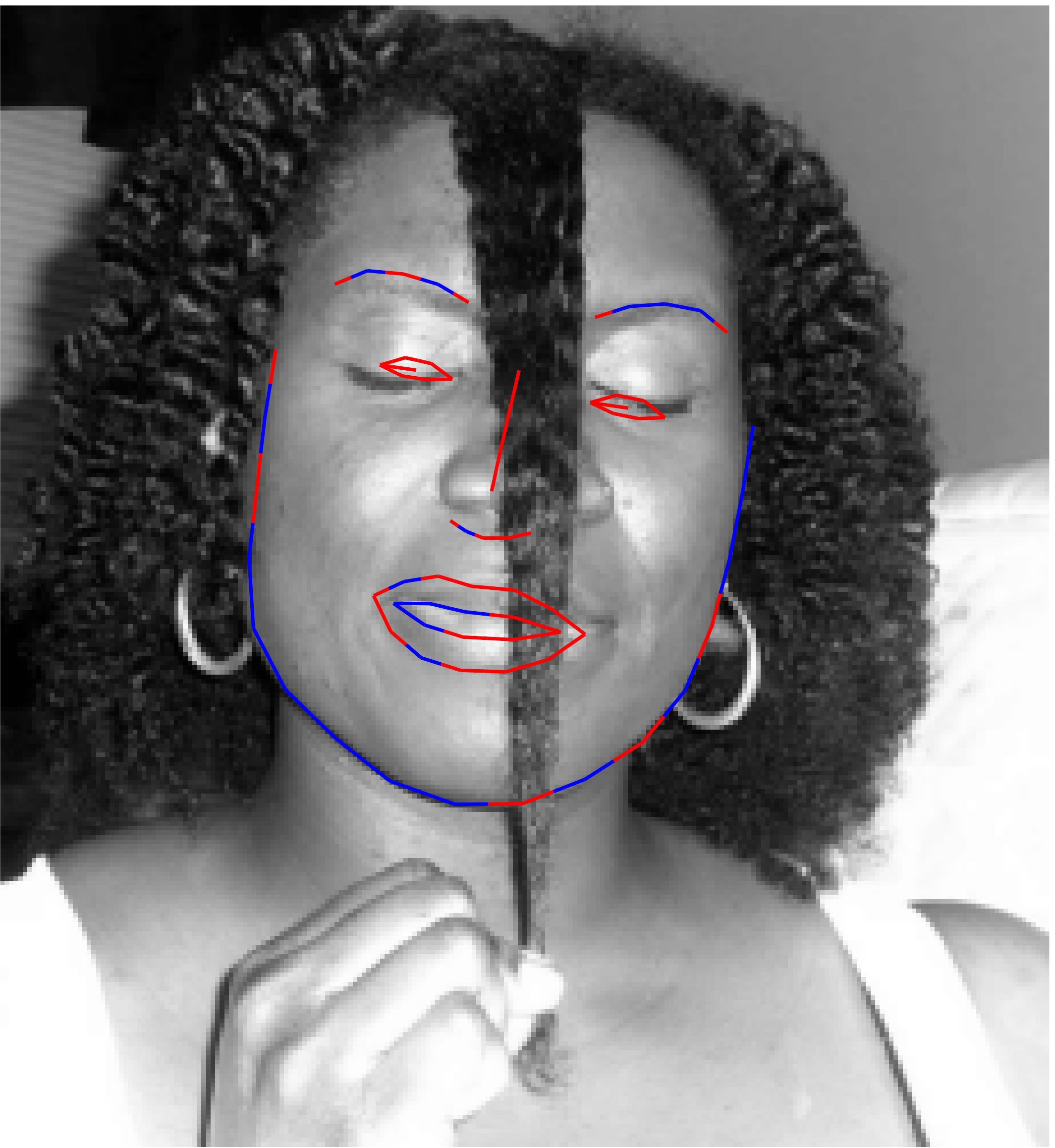}
	\includegraphics[scale=0.085]{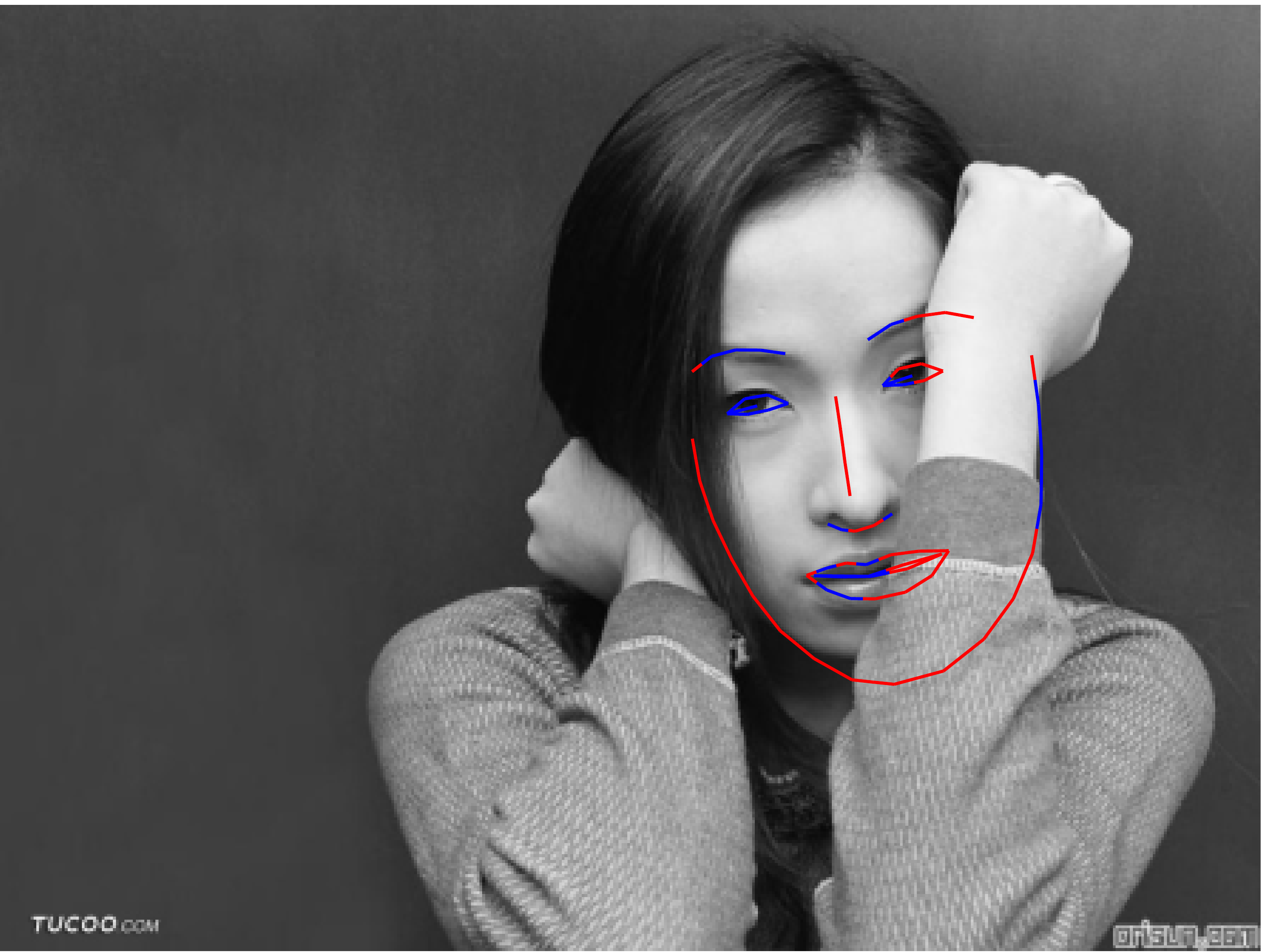}	
	\includegraphics[scale=0.1]{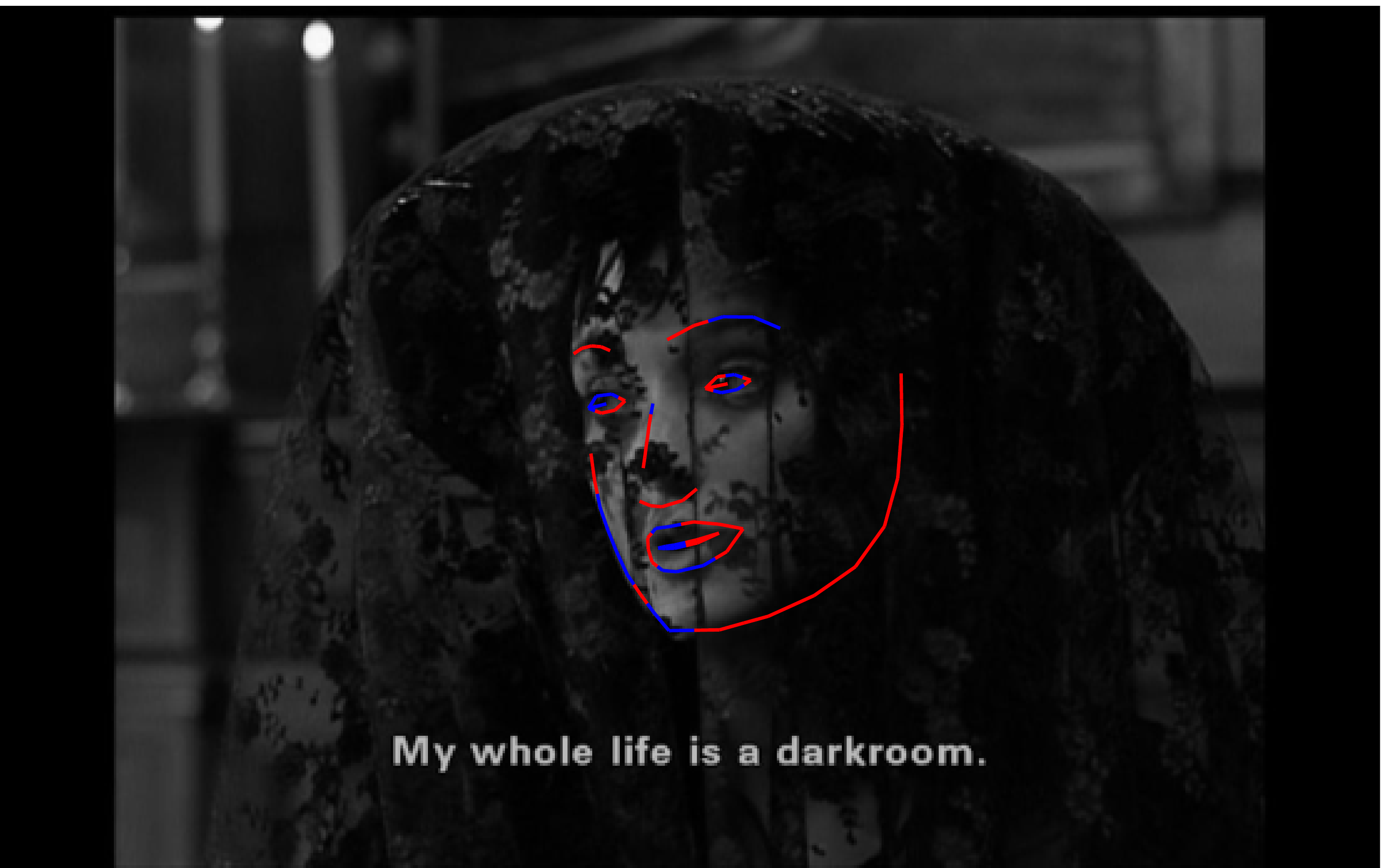}
	
	\includegraphics[scale=0.08]{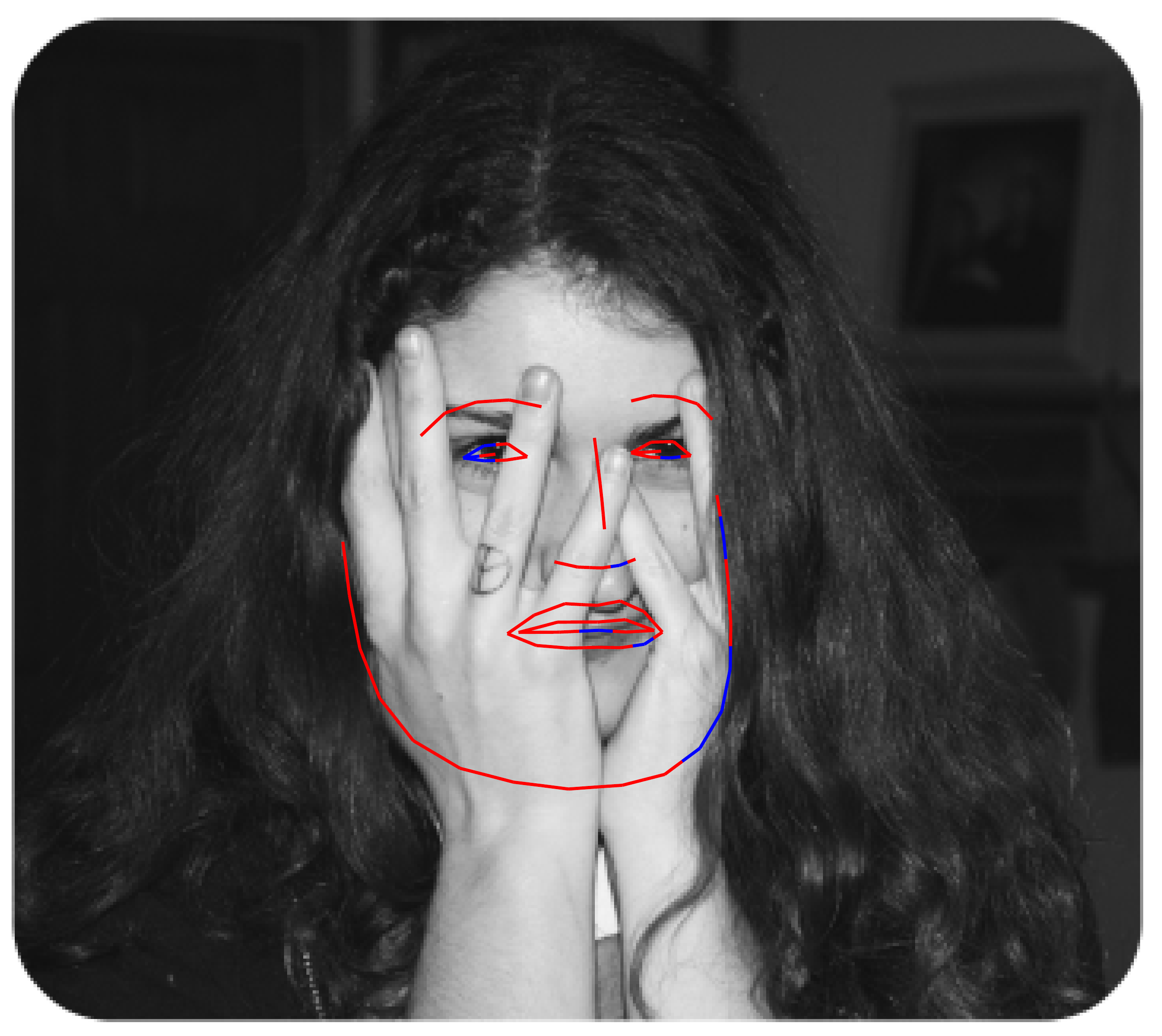}
	\includegraphics[scale=0.067]{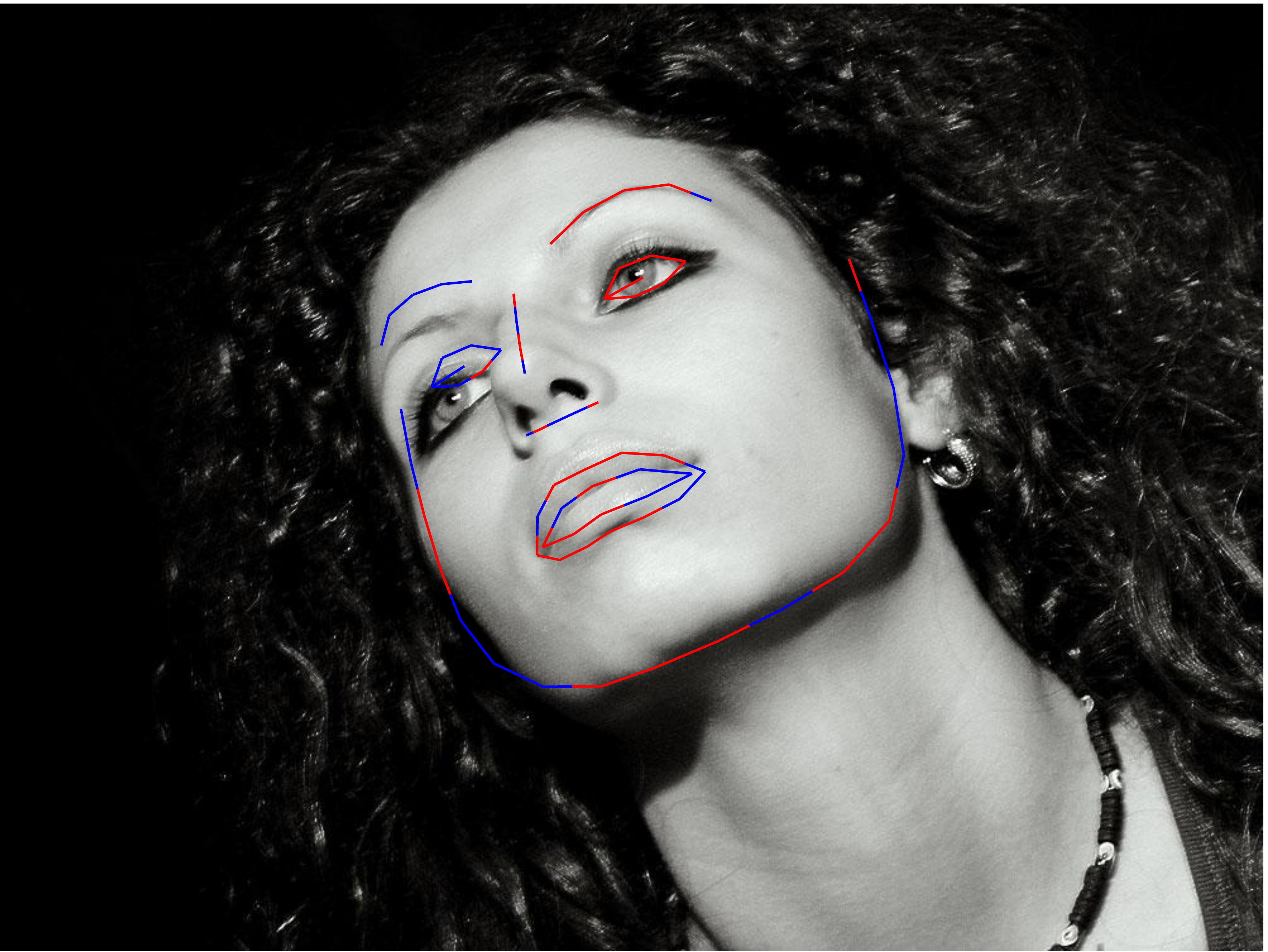}
	\includegraphics[scale=0.069]{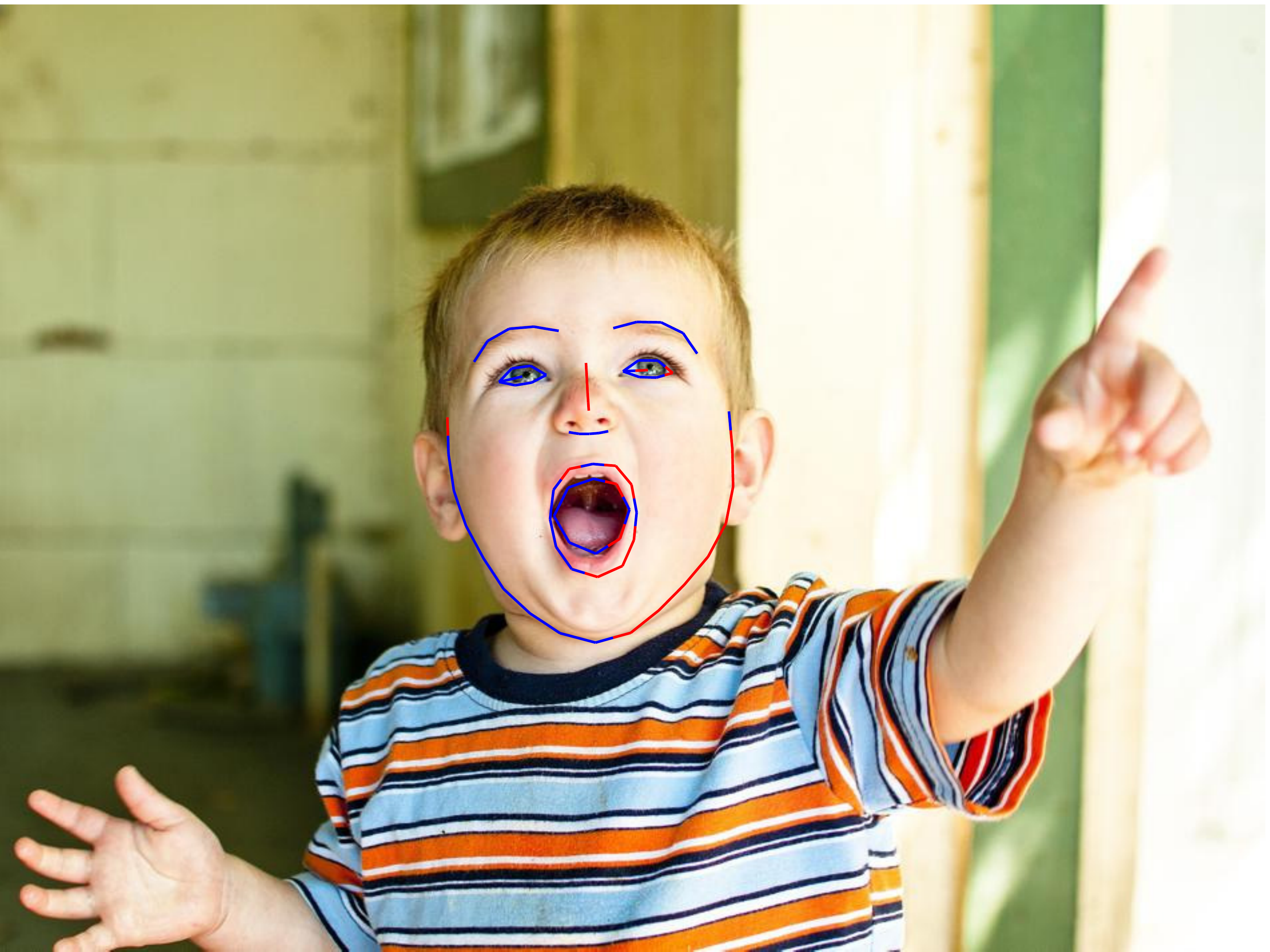}
	\includegraphics[scale=0.071]{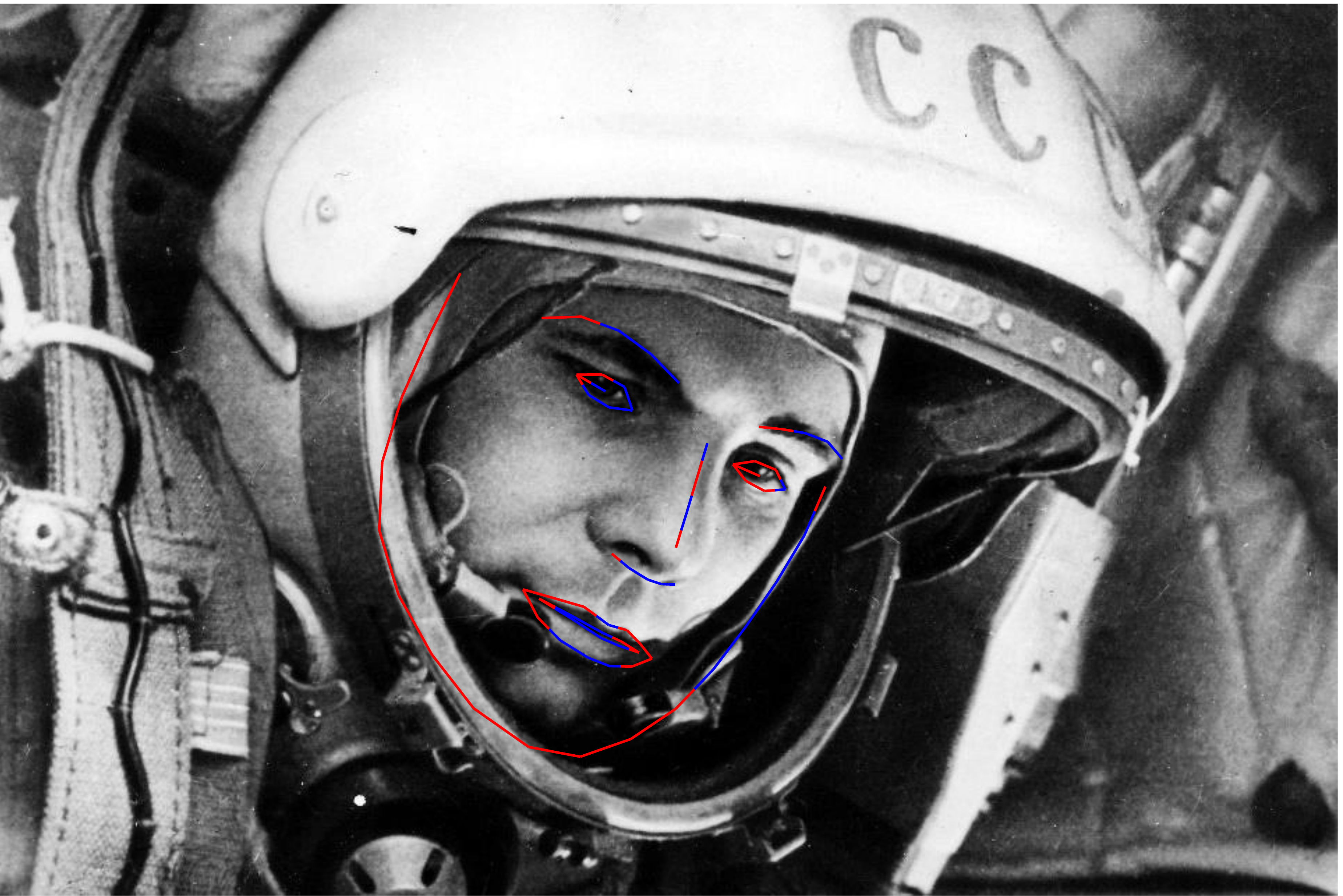}	
	\includegraphics[scale=0.08]{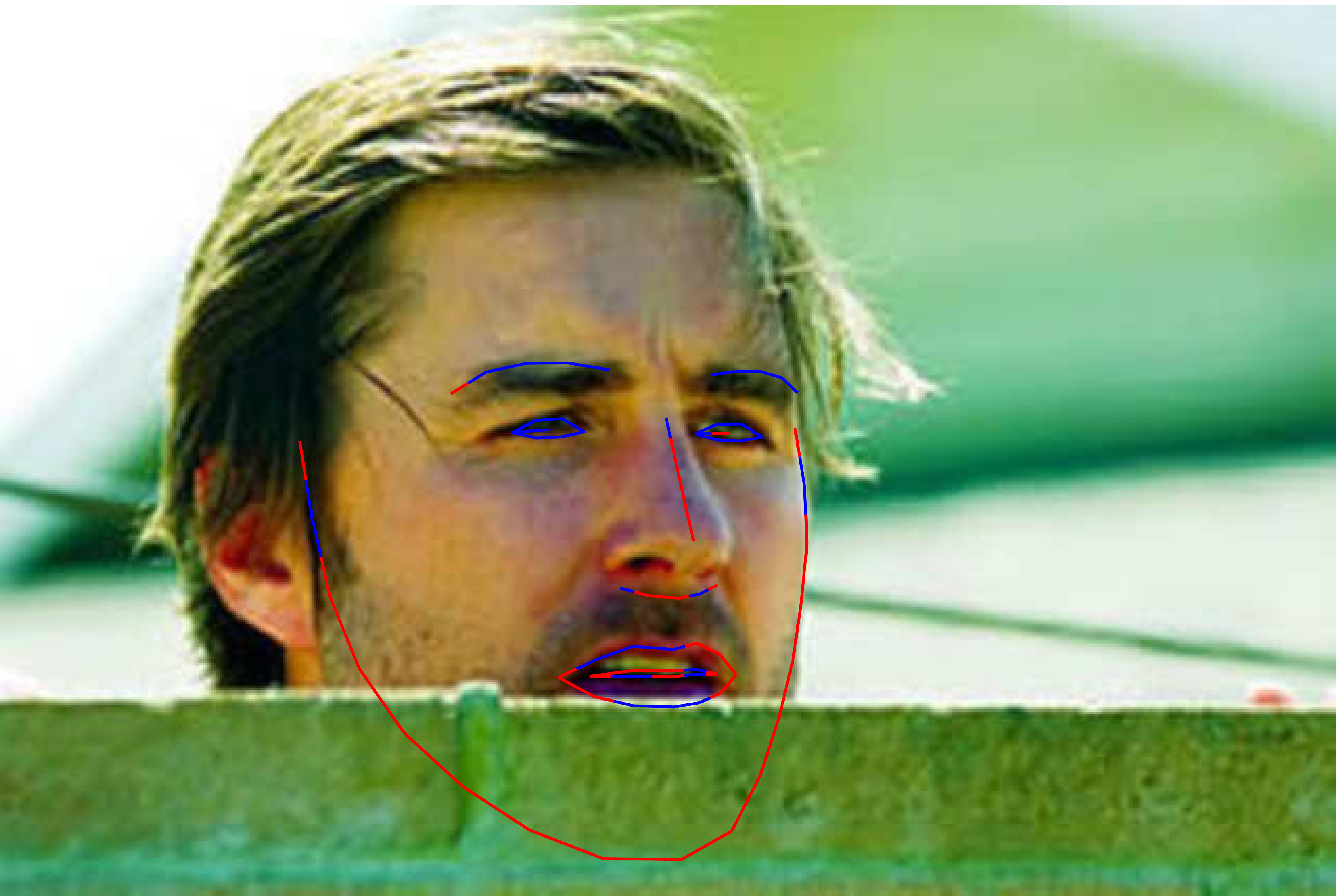}
	\includegraphics[scale=0.08]{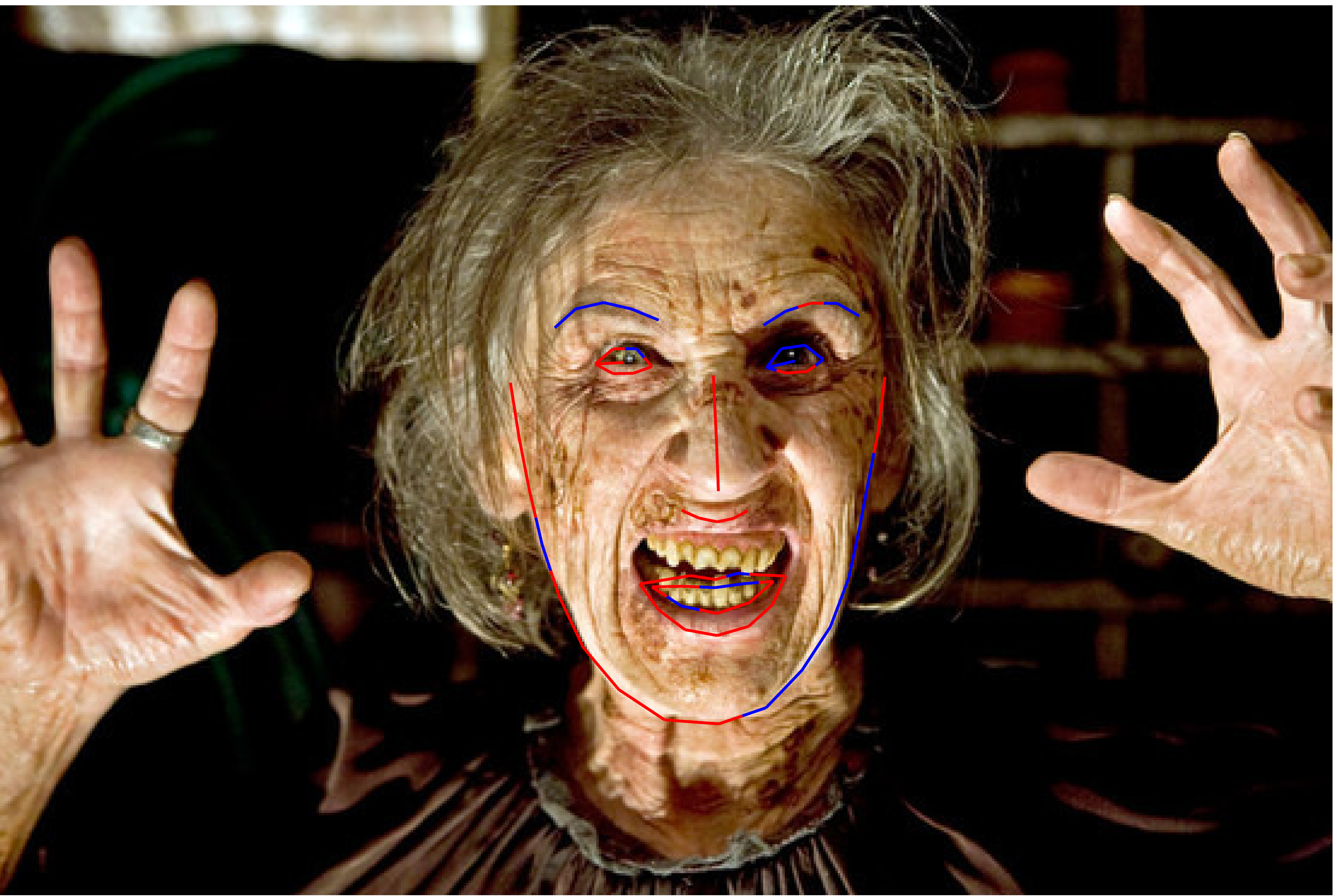}
	\includegraphics[scale=0.072]{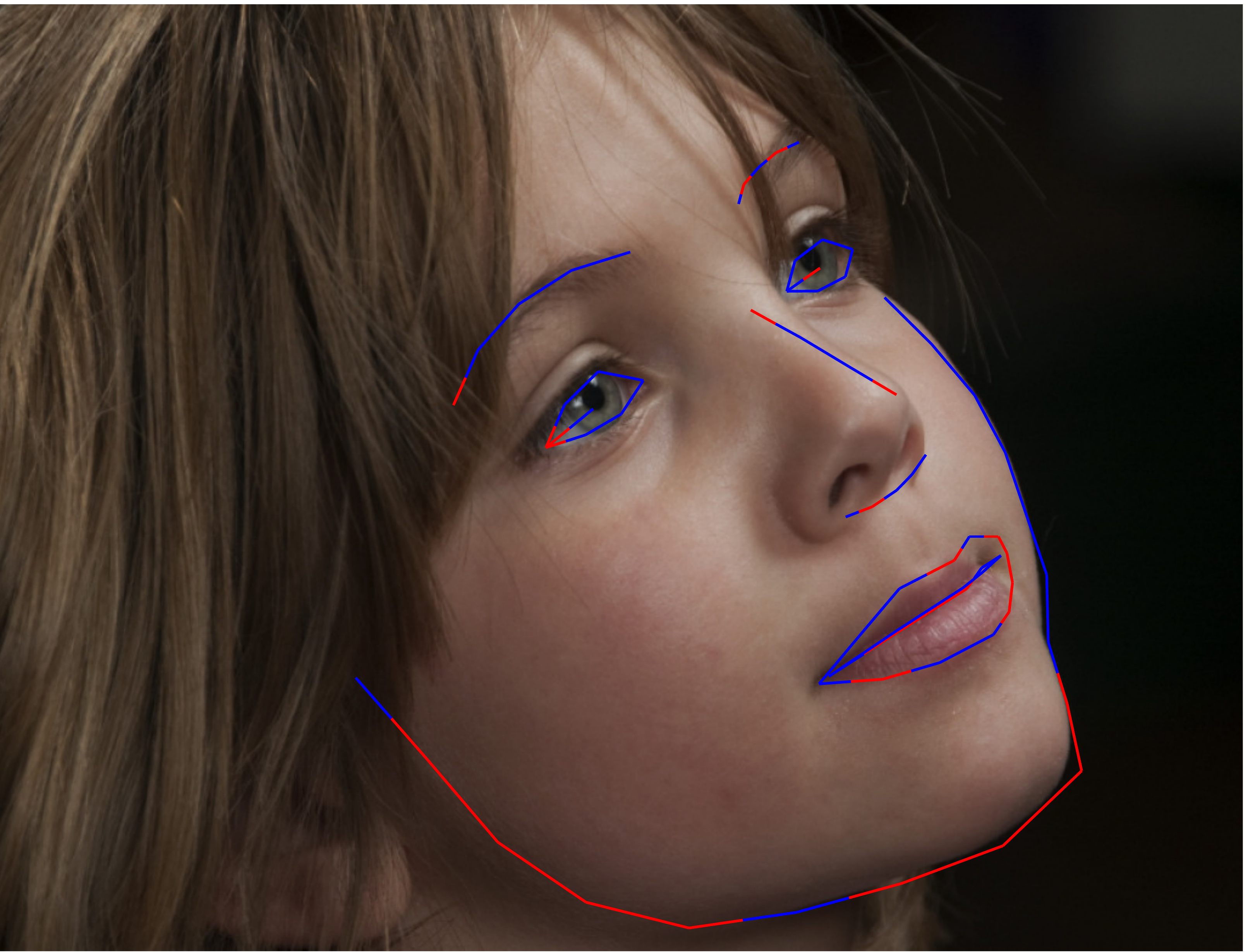}

	\includegraphics[scale=0.095]{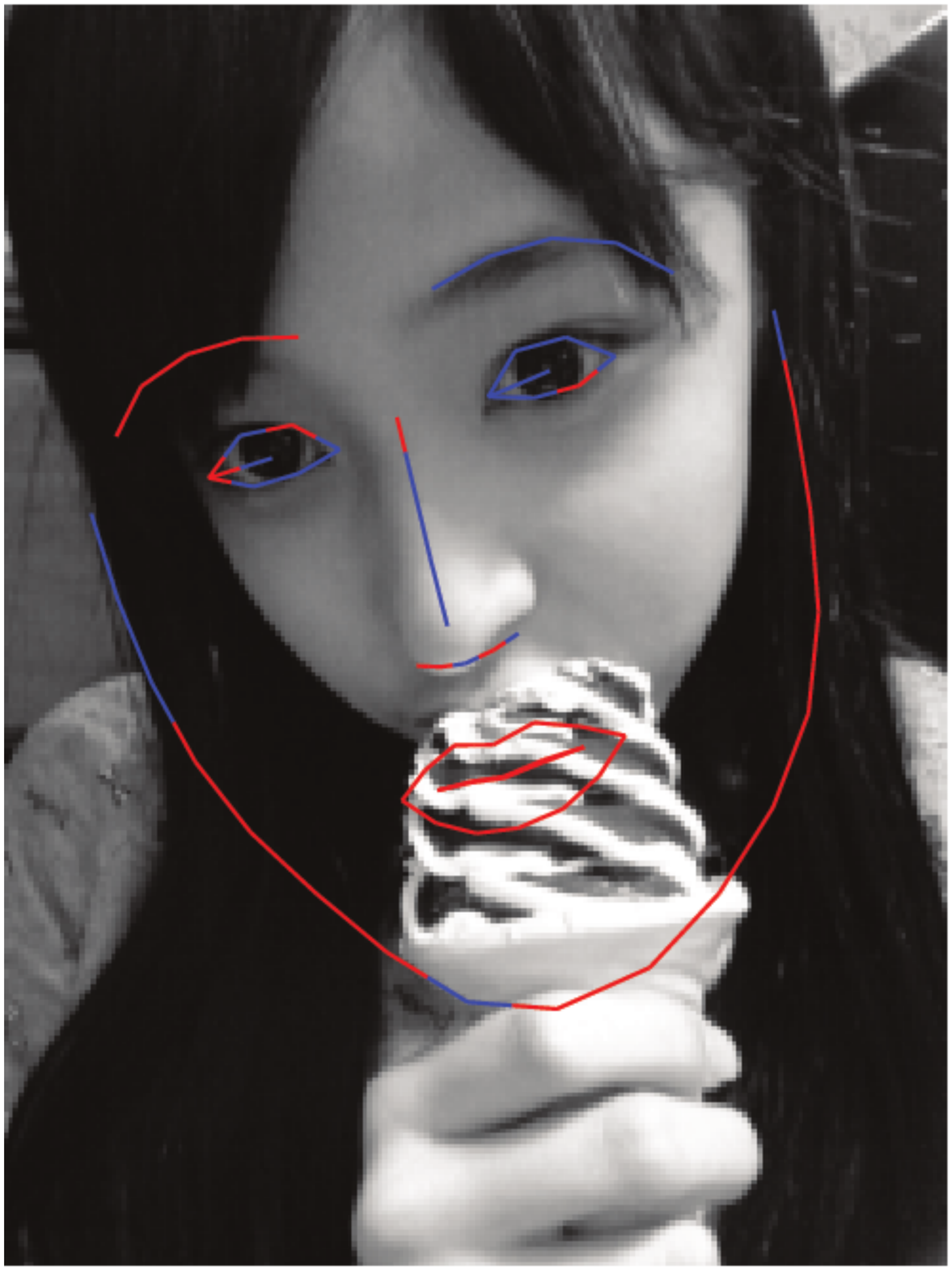}
	\includegraphics[scale=0.105]{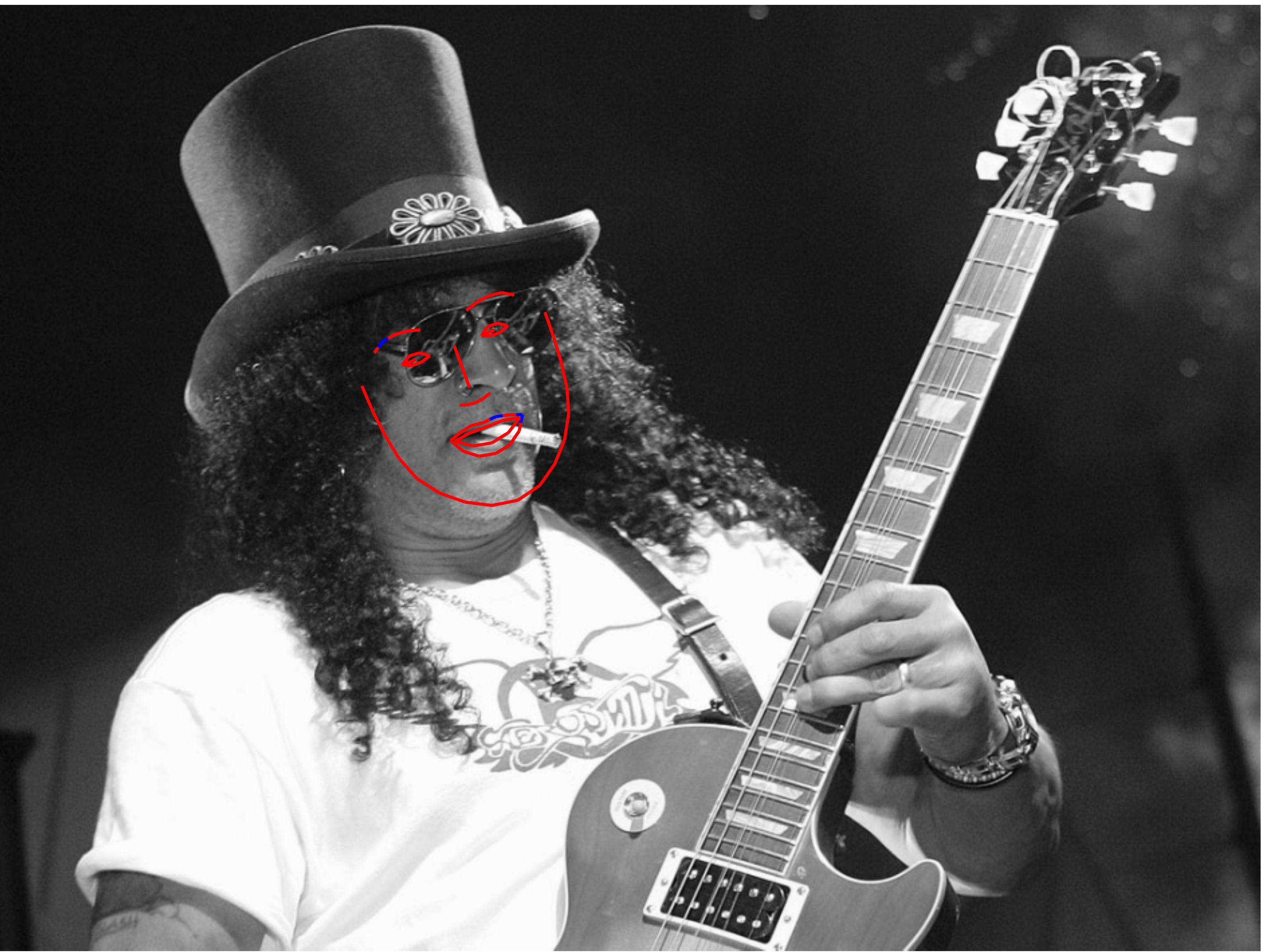}
	\includegraphics[scale=0.1]{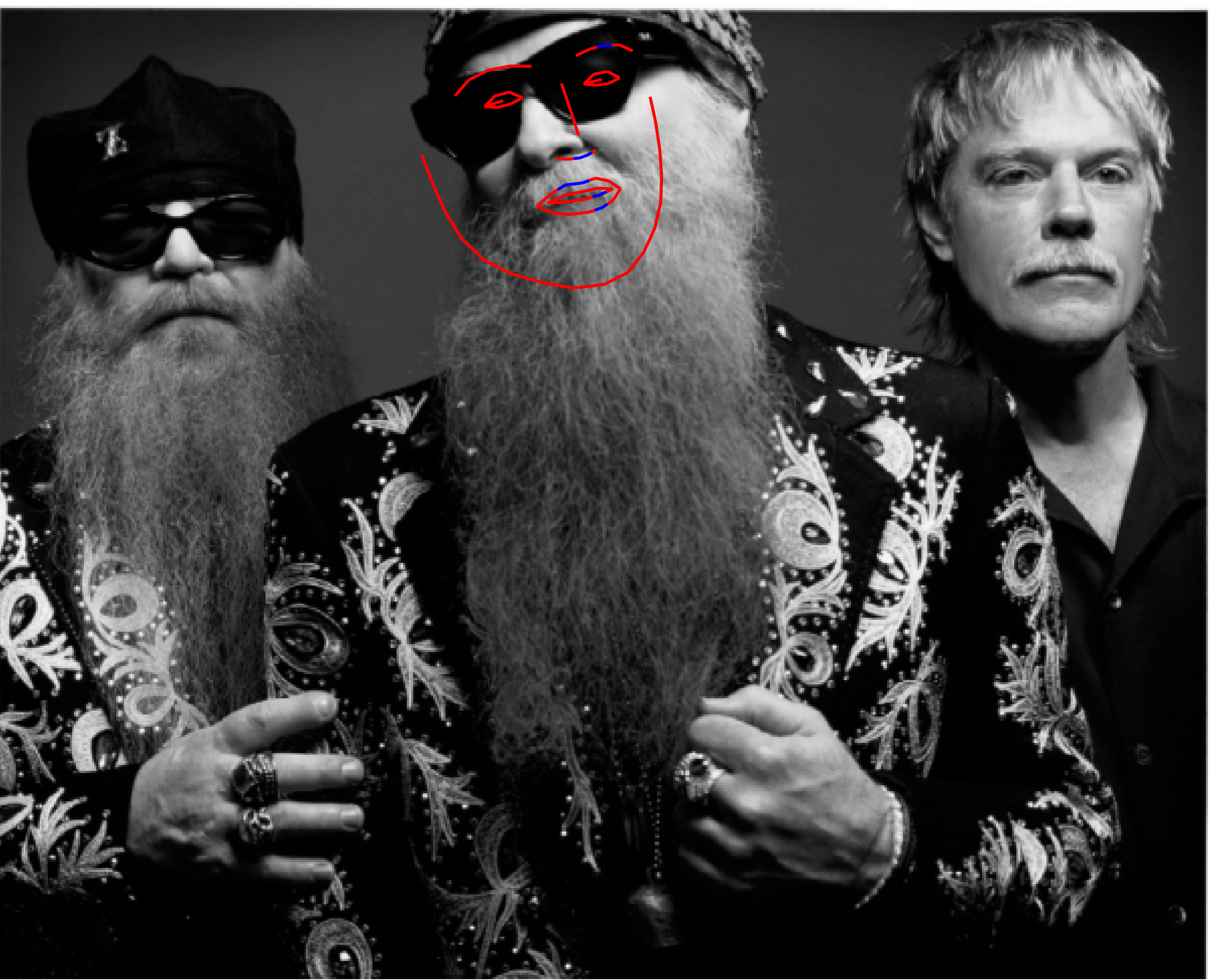}
	\includegraphics[scale=0.078]{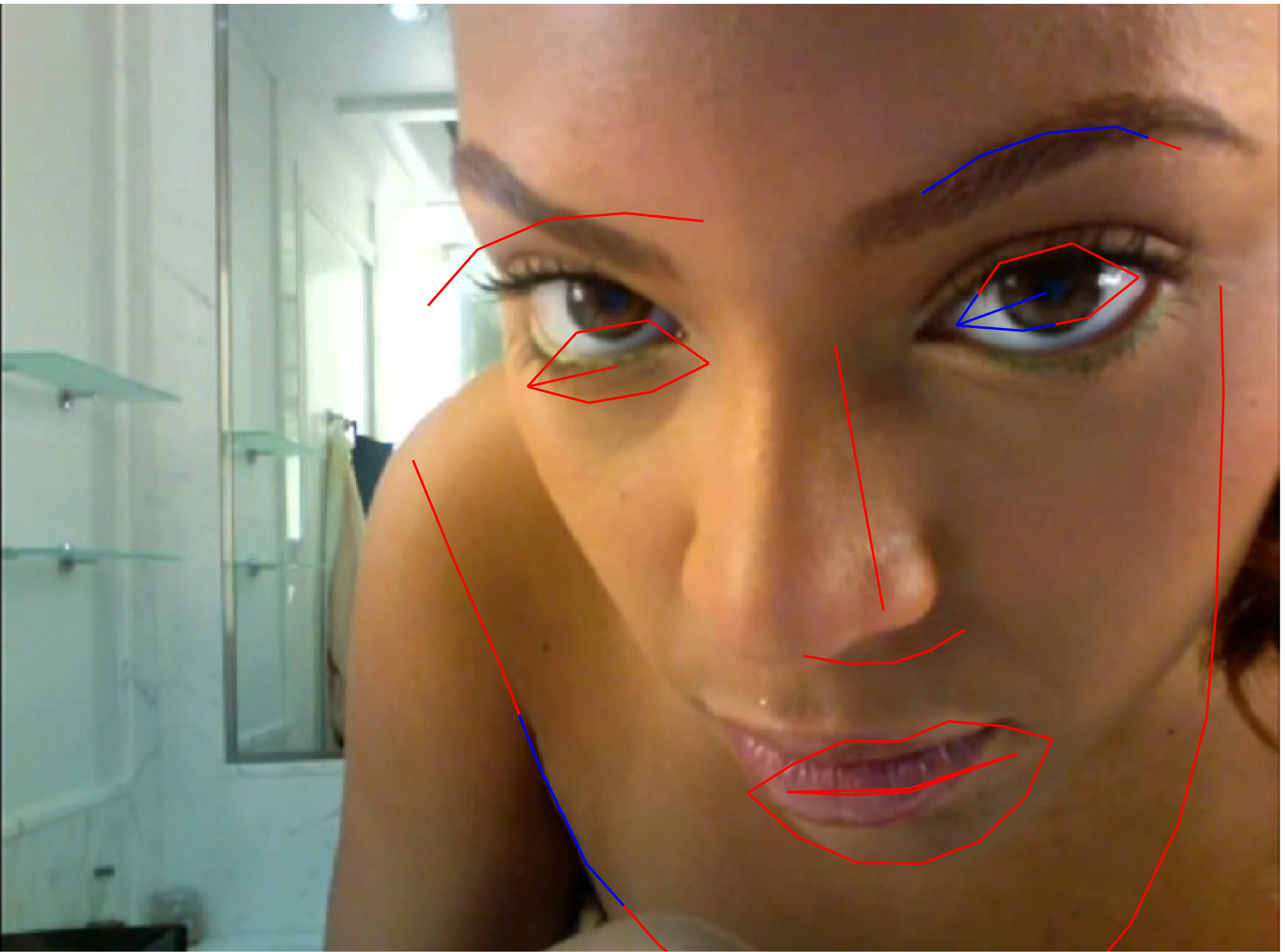}
	\includegraphics[scale=0.095]{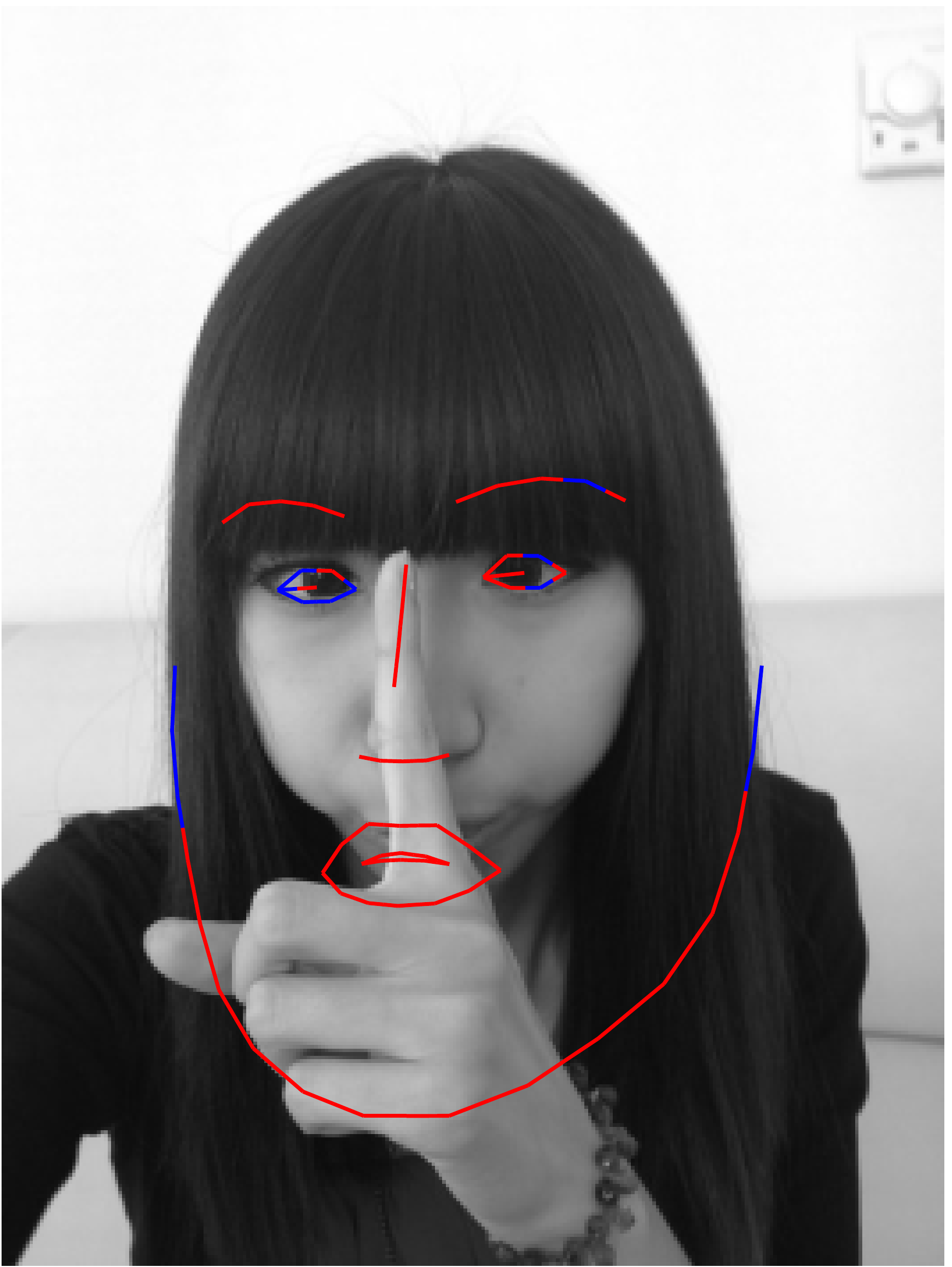}
	\includegraphics[scale=0.075]{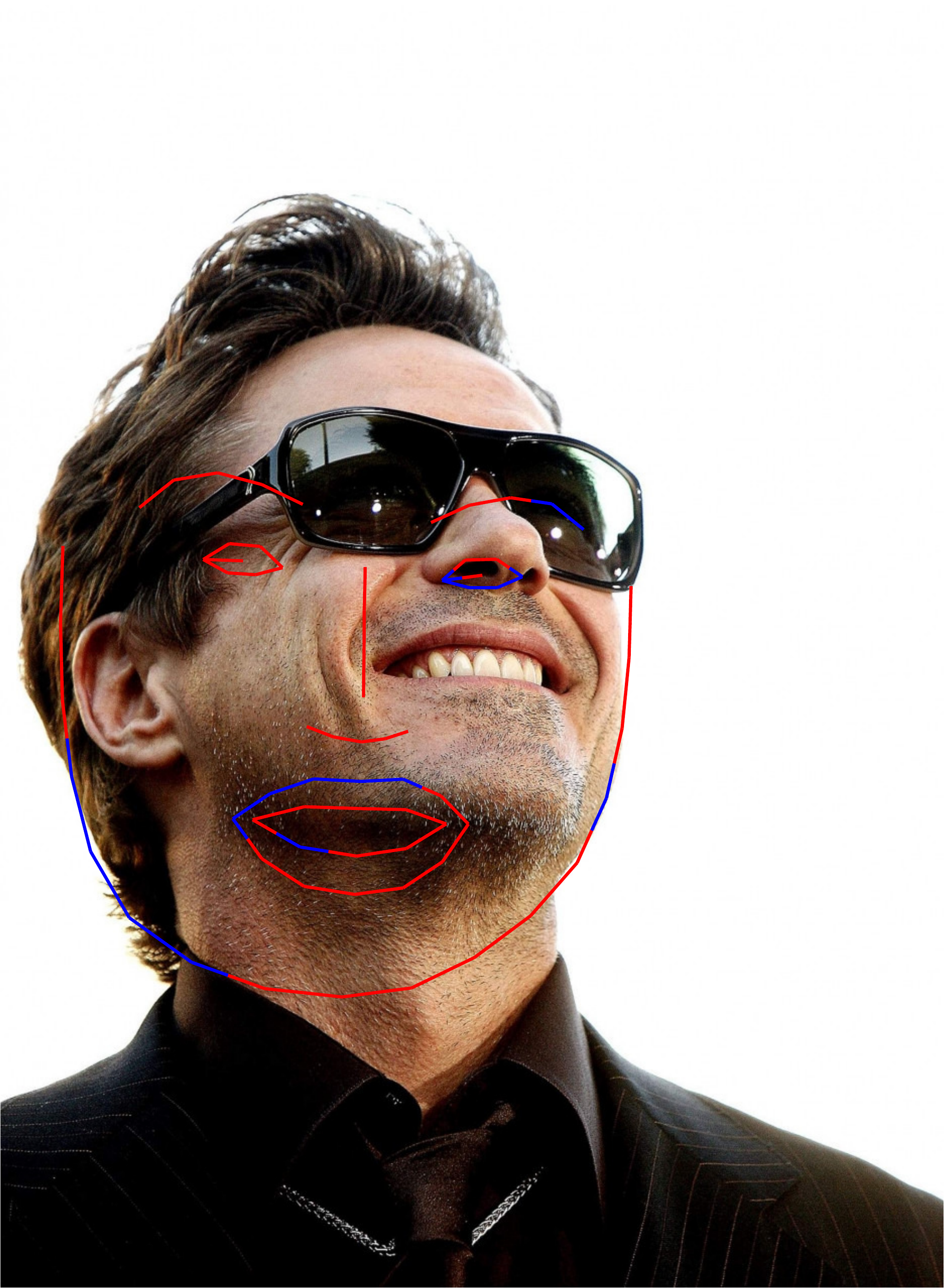}
	\includegraphics[scale=0.09]{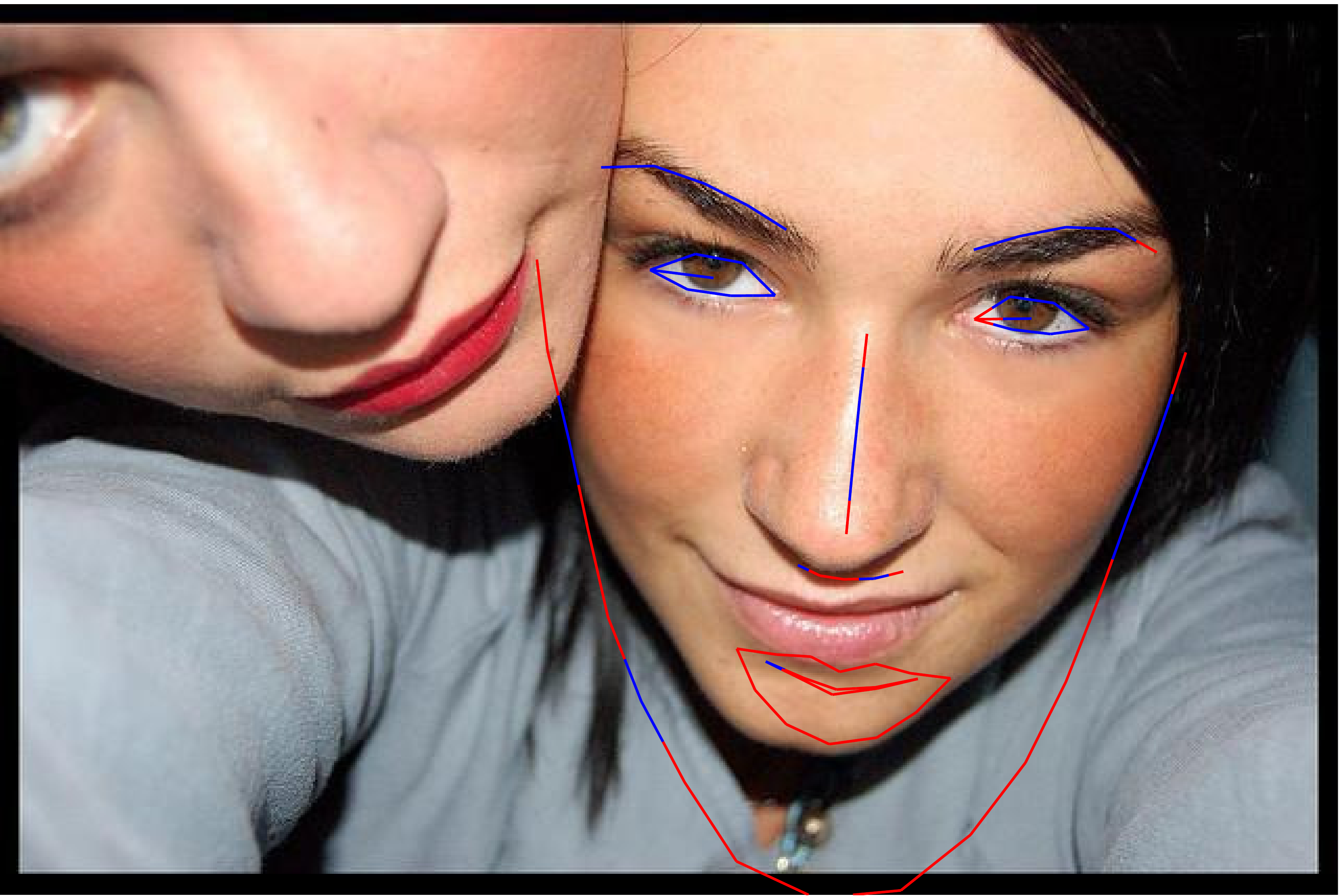}

	\caption{Qualitative Face Alignment Results: blue and red curves represent regions which are determined as visible and occluded, respectively. The top three rows show examples of successful face alignment, while the last row shows examples where ERCLM failed. We note that most of the failures are either due to extreme amounts of facial occlusions or due to pitch variation which is not present in the our training set.}
	\label{fig:examples}
\end{figure*}

\noindent \textbf{Qualitative Results:} Qualitative examples of successful and failed alignment results are shown in Fig. \ref{fig:examples}. Most of these results are from AFW, IBUG and COFW due to the challenging nature of these datasets (large shape variations and variety of occlusions). Despite the presence of significant facial occlusions our proposed method successfully aligns the face across pose and expressions while also predicting the landmark occlusion labels. We note that some visible landmarks are determined as occluded since some regions like the lower jawline are very difficult to detect using the local landmark detectors and hence are not hypothesized to be visible. However, our method is able to accurately hallucinate the facial shape even on the occluded parts of the face from the visible set of landmarks. Most of the face alignment failures of our method are either due to extreme amounts of facial occlusions or due to pitch variation not present in the our training set. Including facial pitch variation in our models can help mitigate such failures.

\section{Ablation Study \label{sec:ablation}}
In this section we provide quantitative evaluation of the various components of ERCLM, namely, discrete multi-modal appearance and shape priors spanning pose and expressions, dense point distribution model and different hypotheses generating sampling strategies for occlusion reasoning. Table \ref{table:ablative} presents quantitative results of the ablative analysis on the AFW, HELEN, LFPW, IBUG, 300W-INDOOR and 300W-OUTDOOR datasets.

\begin{table}[!ht]
\captionsetup{font=footnotesize}
\centering
\scalebox{0.7}{
\begin{tabular}{cccccccccccc}
Dataset & Metric & & \multicolumn{3}{c}{Multi-Modal} & & SPDM & & \multicolumn{3}{c}{Sampling} \\
& & & & & & & & & & & \\
& & & (a) & (b) & (c) & & (d) & & (e) & (f) & (g) \\
\cline{4-6} \cline{8-8} \cline{10-12}
& & & & & & & & & & & \\
\multirow{2}{*}{AFW} & MNLE & & 6.31 & 6.18 & 5.83 & & 5.80 & & 5.77 & 5.73 & 5.79 \\
& FR & & 8.01 & 7.72 & 5.34 & & 4.75 & & 5.34 & 4.74 & 6.23 \\
& & & & & & & & & & & \\
\multirow{2}{*}{HELEN} & MNLE & & 5.25 & 5.14 & 4.94 & & 5.05 & & 4.89 & 4.91 & 4.91 \\
& FR & & 3.09 & 2.62 & 1.93 & & 1.59 & & 1.50 & 1.59 & 1.46 \\
& & & & & & & & & & & \\
\multirow{2}{*}{LFPW} & MNLE & & 5.08 & 4.98 & 4.85 & & 4.91 & & 4.80 & 4.81 & 4.80 \\
& FR & & 3.00 & 2.71 & 2.22 & & 1.55 & & 1.74 & 1.74 & 1.16 \\
& & & & & & & & & & & \\
\multirow{2}{*}{IBUG} & MNLE & & 10.34 & 9.98 & 9.28 & & 9.32 & & 8.89 & 8.94 & 8.84 \\
& FR & & 41.48 & 37.78 & 30.37 & & 25.19 & & 24.44 & 26.67 & 26.67 \\
& & & & & & & & & & & \\
\multirow{2}{*}{300W-IN} & MNLE & & 7.61 & 7.44 & 6.67 & & 6.93 & & 6.58 & 6.51 & 6.72 \\
& FR & & 18.33 & 16.67 & 11.67 & & 11.00 & & 9.33 & 9.33 & 9.67 \\
& & & & & & & & & & & \\
\multirow{2}{*}{300W-OUT} & MNLE & & 8.29 & 8.04 & 7.35 & & 7.37 & & 7.13 & 7.15 & 7.21 \\
& FR & & 21.33 & 18.67 & 14.67 & & 17.00 & & 13.00 & 12.67 & 14.33 \\
& & & & & & & & & & & \\
\hline
\end{tabular}}
\caption{Ablative analysis of the components of ERCLM on datasets with varying difficulty, \textbf{AFW}, \textbf{LFPW}, \textbf{HELEN}, \textbf{IBUG}, \textbf{300W-OUTDOOR} and \textbf{300W-INDOOR}, evaluated over 68 (includes jawline). We report both the Mean Normalized Landmark Error (MNLE) and the alignment Failure Rate (FR). Multiple-Modes: (a) one model spanning pose and expressions, (b) two models, one for each expression spanning pose, (c) five models, one for each pose spanning expression, SDPM: (d) sparse PDM with 68 points instead of our proposed dense PDM and Sampling: (e) random sampling, (f) sampling from detector confidence, (g) greedy selection.}
\label{table:ablative}
\end{table}

\noindent\textbf{Multi-Modal Models:} We compare the performance of our system with varying number of appearance and shape models to span the entire range of pose and expression variations. We consider three models, (a) a single mode spanning the whole range of pose and expression variations, (b) two modes, one for each expression, spanning the full range of pose and (c) five modes, one for each pose, spanning the range of expressions. Each of these models is evaluated using our dense PDM and confidence sampled hypotheses. Unsurprisingly increasing the number of appearance and shape modes improves the performance of our system.

\noindent\textbf{Dense Point Distribution Model:} We evaluate the benefit of modeling the jawline landmarks as contour-like landmarks instead of point-like landmarks as is the common practice. As shown in Table \ref{table:ablative} modeling the contour like nature of the landmarks on the jawline of the face results in lower MNLE. The flexibility afforded to the jawline landmarks by explicitly allowing them to move along its contour results in more accurate localization of these landmarks.

\noindent\textbf{Hypothesis Generation Strategies:} Here we describe the implications of using different sampling based hypotheses generation strategies described in Fig.\ref{fig:sampling}, namely, random sampling, detector confidence sampling and greedy selection. For random and detector confidence based sampling we first sample the landmark indices followed by the true positives from the associated candidate landmarks. For greedy selection, we exhaustively select all combinatorial pairs of landmark indices and then greedily select the top detection for the associated candidate landmarks. The three sampling strategies offer different trade-offs between performance and computational complexity and differ in the prior assumptions on the efficacy of the local landmark detectors. The random sampling strategy makes no assumptions on the detector's ability and instead treats all candidate detections as equally likely, and is thus more robust to erroneous detections (see Table \ref{table:ablative}). Greedy selection on the other hand is highly dependent on the landmark detector's confidence and is thus severely affected by outlier detections. The detector confidence based sampling strategy seeks to tread a middle ground between random sampling and greedy selection, evaluating most of the high confidence detections along with some low confidence detections. Computationally, in our experiments, the number of hypotheses evaluated for greedy selection is about 3x lower than random and detector confidence based sampling is 2x lower than random. 

\section{Discussion \label{sec:discussion}}
\begin{figure}[!ht]
	\captionsetup{font=small}
	\centering
	\begin{subfigure}[b]{0.48\linewidth}
		\centering
		\includegraphics[width=1\linewidth]{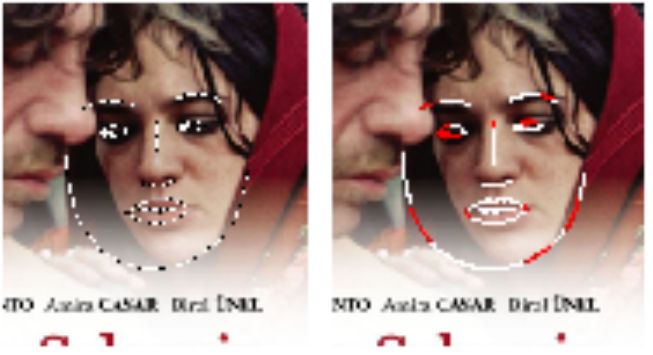}
		\caption{}
	\end{subfigure}
	\begin{subfigure}[b]{0.48\linewidth}
		\centering
		\includegraphics[width=1\linewidth]{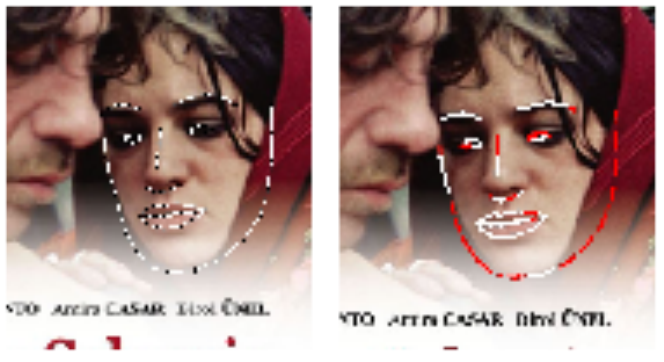}
		\caption{}
	\end{subfigure}
	\caption{Failure case where the top ranked shape is incorrect while a lower ranked shape fits correctly (a) Top ranked hallucinated shape (left) and its refinement(right), (b) Second ranked hallucinated shape (left) and its refinement (right).}
	\label{fig:failure}
\end{figure}

\noindent\textbf{Multiple Hypotheses:} Since face alignment is usually part of a larger system, it is often beneficial to output multiple results and delay the final selection. This allows subsequent steps in the system select the best alignment result using additional top level information, such as human body detection and pose estimation, thereby improving overall system performance. This is one of the main advantages of the proposed approach over existing face alignment methods. Moreover, in most real world images due to the inherent ambiguity in the ground truth face alignment (e.g., occluded parts of the face) it is fallacious to demand one and only one correct face alignment result. In Fig. \ref{fig:failure} we show an example with two hypothesized face alignment results where the top ranked shape is incorrect while the second ranked shape fits correctly. We empirically observed that the correct alignment result is within the top three ranked hypotheses.

\noindent\textbf{Computational Complexity:} We provide a comparative analysis of our method from a computational perspective. Since our method is CLM based it is comparatively slower than regression based face alignment approaches. Our model takes $\sim$10s to align each face while serially searching over all pose and expression modes. Our approach, however, lends itself to heavy parallelization both at the level of pose/expression model as well as at the level of hypotheses evaluation within each model. However, as observed in \cite{yan2013learn} and in our own experiments, regression based methods are highly sensitive to their initializations while CLM based approaches by virtue of searching over locations and scale are highly tolerant to facial bounding box initializations. To improve the tolerance of regression based models to initializations, \cite{yan2013learn} proposes to combine multiple results from randomly shifting and scaling the initial bounding boxes considerably slowing down regression based approaches, taking up to 120 secs for alignment as reported in \cite{yan2013learn}.

\section{Conclusions \label{sec:conclusions}}
Fitting a shape to unconstrained faces ``in-the-wild" with unknown pose and expressions is a very challenging problem, especially in the presence of severe occlusions. In this paper, we proposed ERCLM, a CLM based face alignment method which is robust to partial occlusions across facial pose and expressions. Our approach poses face alignment as a combinatorial search over a discretized representation of facial pose, expression and occlusions. We span over the entire range of facial pose and expressions through an ensemble of independent deformable shape and appearance models. We proposed an efficient hypothesize-and-evaluate routine to jointly infer the geometric transformation and shape representation parameters along with the occlusion labels. Experimental evaluation on multiple face datasets demonstrates accurate and stable performance over a wide range of pose variations and varying degrees of occlusions.

Despite the rapid progress in the recent past on the problem of face alignment, a major challenge remains to be addressed. The current dominant scheme, including ours, that relies on face detection as a pre-requisite for alignment is incorrect. Detection and alignment of faces of unknown pose, expressions and occlusions presents a deeper and more challenging ``chicken-and-egg" problem. Addressing this problem is an exciting direction of future research.

\bibliographystyle{IEEEtran}
\bibliography{IEEEabrv,mybib}
\end{document}